%% file: main.tex
\documentclass{article}

\usepackage{arxiv}
\usepackage{float}
\usepackage[utf8]{inputenc} 
\usepackage[T1]{fontenc}    
\usepackage{hyperref}       
\usepackage{url}            
\usepackage{booktabs}       
\usepackage{amsfonts}       
\usepackage{nicefrac}       
\usepackage{microtype}      
\usepackage{lipsum}
\usepackage{hyperref}
\usepackage{subfig}
\usepackage{natbib}
\usepackage{graphicx}
\usepackage{caption}
\usepackage{float}
\usepackage{subfloat}
\usepackage{xcolor}
\usepackage{booktabs}
\usepackage{amssymb}
\usepackage{amsthm}
\usepackage{verbatim}
\usepackage{amsmath}

\usepackage{algorithm2e}
\usepackage[normalem]{ulem}
\usepackage{xcolor,cancel}
\usepackage[section]{placeins}
\usepackage{multirow}
\usepackage{enumitem}
\usepackage{arydshln}

\title{Multi-way Spectral Clustering\\ of Augmented Multi-view Data through\\ Deep Collective Matrix Tri-factorization}

\author{
  Ragunathan Mariappan, Siva Rajesh Kasa, Vaibhav Rajan\\
Department of Information Systems \& Analytics\\ 
School of Computing\\ 
National University of Singapore
}

\begin{document}
\maketitle

\begin{abstract}
We present the first deep learning based architecture for collective matrix tri-factorization (DCMTF) of arbitrary collections of matrices, also known as augmented multi-view data.
DCMTF can be used for multi-way spectral clustering of heterogeneous collections of relational data matrices to discover latent clusters in each input matrix, across both dimensions, as well as the strengths of association across clusters.
The source code for DCMTF is available on our public repository: \url{https://bitbucket.org/cdal/dcmtf\_generic}.
\end{abstract}



\section{Problem Statement}
Let $E$ be a set of $N$  entities, where the $e^{\rm th}$ entity is denoted by $E^{[e]}$.
Data about the entities is available to us through $d_e$ entity instances (observations along row or column dimensions) in a set of $M$ matrices.
The $m^{\rm th}$ matrix $X^{(m)}_{r_m,c_m}$ 
describes the relationships between the entities across the row and column dimensions denoted by $r_m, c_m$ respectively ($r_m, c_m \in \{1,\ldots,N\}$). 
To show the dimensions of the matrix, we use the notation $X^{(m)}_{d_{r_m} \times d_{c_m}}$.
    
We aim to collectively obtain the matrices: 
    \begin{enumerate}
        \item  $U^{[e]}_{d_e \times b}$ - Latent representations of each entity, where $b$ is the common representation dimension.
        \item $I^{[e]}_{d_e \times k_e}$ - Entity cluster indicators for each entity, where $k_e$ is the number of disjoint clusters of entity $e$.
        \item $A^{(m)}_{k_{r_m} \times k_{c_m}}$ - Association matrix corresponding to each of the matrices $X^{(m)}$, containing the strength of association between clusters of the row and column entities.
    \end{enumerate}
We use the indices $i,j$ for row and column indices in a matrix and the indices $u,v$ 
for clusters.
With $A^{(m)}$, we can find the pair of clusters
that are strongly associated between the entities $r_m$ and $c_m$.
The $u_{r_m}^{\rm th}$ cluster, $u_{r_m} = 1,\ldots, k_{r_m}$ refers to the $u^{\rm th}$ cluster in the row entity instances of the $m^{\rm th}$ matrix. 
The strength of association of the $u_{r_m}^{\rm th}$ cluster of the row entities, with the  $v_{c_m}^{\rm th}$ cluster of the column entities is given by the element $A^{(m)}_{u_{r_m},v_{c_m}}$.

Associations can be found across multiple matrices in the input collection in the form of a \textbf{cluster chain} $\{\mathcal{B}^{(1)}_{\{u_{r_1},v_{c_1}\}},
\mathcal{B}^{(2)}_{\{u_{r_2},v_{c_2}\}},
\ldots, 
\mathcal{B}^{(m)}_{\{u_{r_m}, v_{c_m} \}},
\ldots\}$ where
$\mathcal{B}^{(m)}_{\{u_{r_m}, v_{c_m} \}}$ 
refers to the sub-matrix block formed by row entity instances of the $u_{r_m}^{\rm th}$ cluster and the column entity instances of the 
$u_{c_m}^{\rm th}$ cluster in $X^{(m)}$.
The sub-matrix can be found by considering the row and column entity clusters most strongly associated (given by $A^{(m)}_{u_{r_m}, {v_{c_m}}}$).
The same entity instances (and therefore the same cluster) in a different matrix allows us to ``follow the chain'' to find another entity cluster that is closely associated.

\subsection{Illustration}

\begin{figure*}[!h] 
\centering
    \subfloat[]{
    \centering
            \def\svgwidth{0.35\textwidth} 
            \fontsize{8pt}{8pt}\selectfont
            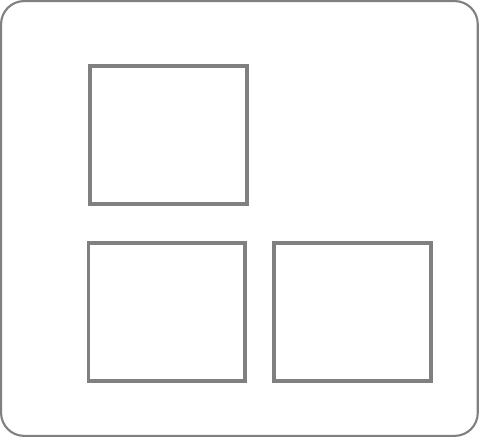
    }
    \subfloat[]{
    \centering
            \def\svgwidth{0.35\textwidth} 
            \fontsize{8pt}{8pt}\selectfont
            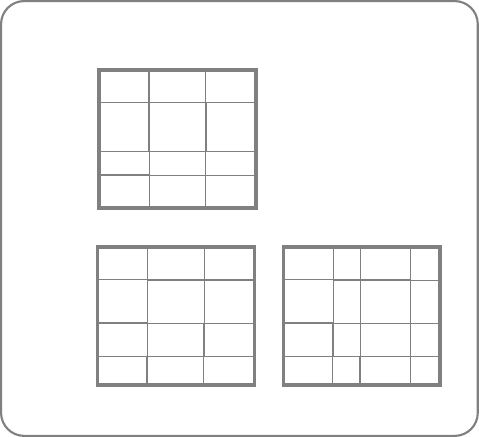
    }
    \caption{
    Left: 4 entities $E^{[1]}, E^{[2]}, E^{[3]}, E^{[4]}$ and 3 relations between the entities, matrices $X^{(1)}_{E^{[1]},E^{[2]}}$, $X^{(2)}_{E^{[1]},E^{[3]}}$ and $X^{(3)}_{E^{[4]},E^{[2]}}$
    Right: An example entity cluster chain $\{
\mathcal{B}^{(3)}_{\{4,2 \}},
\mathcal{B}^{(1)}_{\{1,2\}},
\mathcal{B}^{(2)}_{\{1,3\}}
\}
$. Here instances of entities $E^{[1]}, E^{[2]}, E^{[3]}, E^{[4]}$ are clustered into 4, 3, 4 and 4 groups respectively. 
    }
\label{fig:adr:setting}
\end{figure*}

Figure \ref{fig:adr:setting} (left) shows an example of 3 input matrices $X^{(1)}_{E^{[1]},E^{[2]}}$, $X^{(2)}_{E^{[1]},E^{[3]}}$ and $X^{(3)}_{E^{[4]},E^{[2]}}$ containing dyadic relationships between 4 entities: 
$E^{[1]}, E^{[2]}, E^{[3]}, E^{[4]}$

Figure \ref{fig:adr:setting} (Right) shows multi-way clusters in each matrix obtained by clustering each entity. The shaded blocks correspond to one cluster chain
$\{
\mathcal{B}^{(3)}_{\{4,2 \}},
\mathcal{B}^{(1)}_{\{1,2\}},
\mathcal{B}^{(2)}_{\{1,3\}}
\}
$ 
found by associating the 4th $E^{[4]}$ cluster -- the 2nd $E^{[2]}$ cluster -- the 1st $E^{[1]}$ cluster -- the 3rd $E^{[3]}$ cluster.
Similarly other cluster chains exist (not shown) that allow us to associate clusters across entity types for an arbitrary collection of input matrices.

\section{Background}

\subsection*{Spectral Clustering}
\label{spectralclust}

Consider 
$n$ instances of $p$--dimensional observations in a single data matrix $X \in \mathbb{R}^{n \times p}$, and a similarity matrix $S \in \mathbb{R}^{n \times n}$ that gives us pairwise similarities between the observations.
We can view this data as a graph where each node is an observation and edges denote similarity, their weights quantifying the similarity.
A natural notion of clustering is given by partitioning the graph into subgraphs such that edges {\it within} each subgraph have high similarity and edges {\it across} subgraphs have low similarity. 
Such partitions can be obtained by solving the mincut problem.

The mincut problem aims to find $k$ partitions that minimizes the sum of edge weights across all pairs of partitions, i.e., $\frac12 \sum_{u=1}^k W (B_u, \bar{B}_u)$, where $B_u$ denotes the $u^{\rm th}$ partition and $\bar{B}_u$ represents the complement (the set of all nodes not in $B_u$) and $W$ is the sum of all edge weights across the partitions, $W(B_u,B_v) = \sum_{i \in B_u, j \in B_v} S_{ij}$.
In practice, this approach does not work well as it often yields imbalanced clusters, e.g., a cluster with one or a few nodes.
So, formulations like RatioCut
\citep{hagen1992new} add balancing conditions to penalize small-sized clusters as given below, where $|B_u|$ denotes the number of nodes in the partition $B_u$.

\begin{equation}
\text{RatioCut}(B_1,\ldots,B_k) = \frac12 \sum_{u=1}^k \frac{W (B_u, \bar{B}_u)}{|B_u|}
\end{equation}

However, the use of such balancing conditions makes the problem NP-hard \citep{wagner1993between}.
Spectral clustering provides a polynomial-time relaxation that works very well in practice. We describe the key idea of the relaxation.

Given $k$ partitions, define vigorous cluster indicator vectors as follows $J_{iu}$, for $i = 1,\ldots,n$ and $u = 1, \dots, k$ as follows:
\begin{equation}
J_{iu} =    	
	    	\begin{cases}
				\frac{1}{\sqrt{|B_u|}} & \text{if row } X_i \in B_u \\ 
				 0 & \text{Otherwise}
			\end{cases}
\label{eqn:constraint:vigorous} 
\end{equation}
Note that although $J \in \mathbb{R}^{n \times k}$, it still retains interpretability as cluster indicators by capturing  the disjoint cluster memberships, 
and the columns of $J$ are orthonormal, i.e., $JJ^T = \mathbf{I}$, where $\mathbf{I}$ is the identity matrix. 

Consider the Graph Laplacian $L$ constructed from the Similarity matrix: $L = D - S$, where $D$ is the $n \times n$ diagonal matrix where
the  $i^{\rm th}$ row's diagonal element contains
the degree $\sum_{j=1}^n S_{ij}$.
It can be shown that minimizing RatioCut is equivalent to solving the following trace minimization problem \citep{}:
\begin{equation}
\min_{B_1,\ldots,B_k} Tr(J^T L J) \text{ subject to }  J^T = \mathbf{I}
\end{equation}
This problem is relaxed by allowing the matrix $J$ to take {\it arbitrary} real values and denote this relaxed matrix by $C$:
\begin{equation}
\label{eq:spectralloss}
\min_{C \in \mathbb{R}^{n \times k}} Tr(C^T L C) \text{ subject to }  CC^T = \mathbf{I}
\end{equation}

The solution to this problem can be obtained by choosing $C$ as the matrix containing the $k$ eigenvectors with the smallest eigenvalues (follows from the Rayleigh-Ritz theorem, section 5 in \citep{lutkepohl1996handbook}).
To transform this real valued matrix to a binary matrix indicating the partitions, $k$-Means is used on the rows of $C$.
This is essentially the normalized spectral clustering algorithm of \cite{shi2000normalized}:
\begin{enumerate}
    \item Construct the Similarity matrix 
    	  $S \in \mathbb{R}_{q \times q}$
    \item Construct the Degree matrix 
    	  $D$ = $\textit{diag}(\sum_{j=1}^n S_{ij})$ 
    	  and the Laplacian matrix 
    	  $L$ = $D$ - $S$
    \item Construct $C \in \mathbb{R}_{q \times k}$, 
    	  where the column space ${(c_u)}_{(u = 1, \ldots, k)} \in \mathbb{R}^{q}$ are the eigenvectors corresponding to the {\lq \textit{smallest} \rq} $k$ eigenvalues ${(\lambda)}_{(u = 1, \ldots, k)}$ of $L$
    \item Obtain the cluster indicators by clustering $C$ using k-means
\end{enumerate}

More details and other interpretations of spectral clustering are discussed in \citep{von2007tutorial}.

\subsection*{SpectralNet}
\label{spectralnet}

SpectralNet \citep{shaham2018spectralnet} is a neural architecture to obtain approximate Laplacian embeddings and perform clustering.
Gradient-based stochastic optimization on the neural network enables SpectralNet to scale to large datasets where computing eigenvectors can be expensive.
Further, it enables out-of-sample extension that allows us to obtain spectral embeddings of {\it unseen} datapoints, i.e., data not used during network training.

SpectralNet obtains the embeddings $C$ through a neural network $F_{\theta}(X)$ where $\theta$ denotes the weights of the network.
The network is trained to minimize the spectral loss given by equation \ref{eq:spectralloss}.
However, incorporating the orthogonality constraint within the neural network is not straightforward.

The orthogonality constraint is implemented by adding an additional layer at the end whose weights are set to orthogonalize the input to the layer in the following manner.
It can be shown that for any matrix $Y$ such that $YY^T$ is full rank, an orthogonal matrix $Y(Z^{-1})^T$ can be obtained where $Z$ is obtained by the Cholesky decomposition $Y^TY = ZZ^T$ and $Z$ is lower triangular \citep{shaham2018spectralnet}.
Thus if  $\tilde{C}$ is the input to the last layer of the neural network, an orthogonal output can be obtained by constructing a linear layer with weights $(\tilde{H}^{-1})^T$ such that $\tilde{C}^T\tilde{C} = \tilde{H}\tilde{H}^T$.
This ensures that the output of the network $C = \widetilde{C} (\widetilde{H}^{-1})^{T}$ is orthogonal. 

However, this necessitates a two-step process of training SpectralNet, that alternates between the orthogonalization step and gradient descent.
In the orthogonalization step, Cholesky decomposition of $\tilde{C}^T\tilde{C}$ is performed to obtain $(\tilde{H}^{-1})^T$ and the weights of the last layer are frozen.
In the gradient descent step,
forward propagation is done using the weights $\theta$ (for the rest of the network) and the frozen weight in the last layer, and the gradient of $\mathcal{L}_{\theta} = Tr(C^T L C)$ is used during backpropagation to tune the weights $\theta$.

Other elements of SpectralNet include the use of a Siamese Network \citep{chopra2005learning} to obtain the similarities and minibatch training of the entire network. Details are in \citep{shaham2018spectralnet}.
At a high-level, their algorithm has the following steps. 
\begin{enumerate}
    \item Construct the Similarity matrix
          $S \in \mathbb{R}^{n \times n}$
    	  using Siamese network 
    \item Construct the Degree matrix $D$ = $\textit{diag}(
    \sum_{j=1}^n S_{ij})$ 
    	  and the Laplacian matrix $L$ = $D$ - $S$
    \item Construct $C \in \mathbb{R}^{q \times k}$,
    	  by minimizing \\
    	  \begin{eqnarray}\label{eqn:spectralnet:obj}
            \mathcal{L}_{\theta} &=& Tr(C^T L C)
          \end{eqnarray}
    	  where \\
    	  \begin{eqnarray}
    	    C &=& \widetilde{C} (\widetilde{H}^{-1})^{T} \label{eqn:spectralnet:c} \\
    	    \widetilde{H}\widetilde{H}^T &=& \text{cholskey}(\widetilde{C}\widetilde{C}^T) \label{eqn:spectralnet:cholskey} \\
    	    \widetilde{C} &=& F_{\theta}(X) \label{eqn:spectralnet:nn}
    	  \end{eqnarray}
    \item Obtain the cluster indicators by clustering $C$ using k-means
\end{enumerate}

Once the network is trained, out-of-sample extension is straightforward.
The given datapoint is forwardpropagated through the network to obtain its embedding and the cluster assignment is done through the nearest centroid obtained by k-Means.

Note that SpectralNet cannot be used for multi-way clustering or for inputs that are collections of matrices.
A generalization is non-trivial since it requires handling multiple datatypes in the input matrices and defining a suitable similarity metric of entities that may be part of multiple matrices, all within a formulation for multi-way clustering of each matrix.

\subsection*{Spectral Relational Clustering}
\label{cfrm}
Multi-way clustering
, also called co-clustering, for a single matrix $X \in \mathbb{R}^{d_1 \times d_2}$ 
attempts to find $k_1$ clusters of the row vectors and $k_2$ clusters of the column vectors.
A standard approach to multi-way clustering is through tri-factorization $X \approx I^{[1]} A {I^{[2]}}^T$.
In the binary cluster indicator matrices $I^{[e]} \in \{0,1\}^{d_e \times k_e}$, 
$I_{iu}^{[e]} = 1$ indicates that the $i^{\rm th}$ object along the $e^{\rm th}$ dimension of $X$ (i.e., the row or column entity), for $e \in \{ 1,2 \}$, is associated with the $u^{\rm th}$ cluster. 
Further,  to ensure that each object belongs to only one cluster, the constraint $\sum_{u=1}^{k_i} I_{iu}^{[e]} = 1$ is imposed.
$A \in \mathbb{R}^{k_1 \times k_2}$ is a cluster association matrix that quantifies the pairwise strength of association across $k_1$ clusters along the row dimension and $k_2$ clusters along the column dimension.

Collective Factorization of Related Matrices (CFRM) \citep{cfrm} is a general model for multi-way clustering of arbitrary collections of matrices,
where cluster indicator matrices $I^{[e]}$  for each entity and association matrices $A^{(m)}$ for each matrix are obtained. 
CFRM  is defined by the optimization model below, where $||.||$ denotes the Frobenius norm:
\begin{equation}
\label{eq:cfrmmodel}
\min \sum_{m} \Big(\Big|\Big|X^{(m)} - I^{[r_m]} A^{(m)} I^{[c_m]^T}\Big|\Big|^2\Big) \text{ subject to } \sum_{u=1}^{k_e} I_{iu}^{[e]} = 1 \quad 1 \le i \le d_e 
\end{equation}

It can be shown this minimization problem is equivalent to solving a trace minimization problem defined in the following manner.
Analogous to the cluster indicator matrix in equation \ref{eqn:constraint:vigorous} we define a cluster indicator matrix for each entity:
\begin{equation}
J^{[e]}_{iu} = \begin{cases}
				\frac{1}{\big|{\pi^{[e]}_u}\big|^{\frac{1}{2}}} & \text{if } 
				I^{[e]}_{iu} = 1, \text{ i.e., }
				i^{\rm th} \text{ entity instance } \in \pi^{[e]}_u \\ 
				 0 & \text{Otherwise}
\end{cases}
\label{eqn:cfrm:constraint:vigorous} 
\end{equation}
    where ${\pi^{[e]}_u}$ is the  $u^{\rm th}$ cluster of $e^{\rm th}$ entity instances
    and ${\big|\pi^{[e]}_u\big|}$ denotes its cardinality.
For each entity $E^{[e]}$, a symmetric matrix $\mathcal{M}^{[e]}$ is constructed using all the matrices that contain the entity:
\begin{equation}
\mathcal{M}^{[e]} = \sum_{m} \mathcal{I}[r_m == e] X^{(m)} J^{[c_m]} J^{[c_m]^T} X^{(m)^T} + \ 
    					\mathcal{I}[c_m == e] {X^{(m)}}^T J^{[r_m]} J^{[r_m]^T} X^{(m)} 
\end{equation}
where     
$$\mathcal{I}[c] = \begin{cases}
1 \text{ if condition } c \text{ is true}\\
0 \text{ otherwise.}
\end{cases}
$$
Note that the definition uses the cluster indicators of column entities in the matrices where row entity $e$ is present and vice versa.

The minimization problem (\ref{eq:cfrmmodel}) is shown to be equivalent to the following maximization problem, where $\mathbf{I}_{k_e}$ is the identity matrix of dimension $k_e \times k_e$:
\begin{equation}
\label{eq:cfrmtrace}
\max_{J^{[e]}} \sum_{e} Tr \big(J^{[e]^T} \mathcal{M}^{[e]} J^{[e]}\big) \text{ subject to } J^{[e]^T}J^{[e]} = \mathbf{I}_{k_e}
\end{equation}

Similar to the relaxation for RatioCut, if we allow the matrix $J^{[e]}$ to take arbitrary real valued orthonormal matrix, then each individual trace term above can be maximized by computing the leading $k_e$ eigenvectors of $\mathcal{M}^{[e]}$.
This forms the basis of the Spectral Relational Clustering algorithm which iteratively solves the eigendecomposition problem for each $\mathcal{M}^{[e]}$ while fixing the cluster indicator matrices for all other entities, until convergence. Finally, the real valued matrices are used to obtain discrete partitions using k-means.

Note that the matrix $\mathcal{M}^{[e]}$ can be viewed as a  weighted \textit{distance} matrix formed by the inner product of $X^{(m)}$ weighted by the cluster indicators $J^{[r_m]}$ or $J^{[c_m]}$:
\begin{eqnarray}\label{eqn:cfrm:M}
    		\mathcal{M}^{[e]} &=& \sum_{m} \mathcal{I}[r_m == e] X^{(m)} J^{[c_m]} J^{[c_m]^T} X^{(m)^T} + \ 
    					\mathcal{I}[c_m == e] {X^{(m)}}^T J^{[r_m]} J^{[r_m]^T} X^{(m)} \\
    				&=& \sum_{m} \mathcal{I}[r_m == e] (X^{(m)} J^{[c_m]}) (X^{(m)} J^{[c_m]})^T + 
    					\mathcal{I}[c_m == e] ({X^{(m)}}^T J^{[r_m]}) (X^{(m)^T} J^{[r_m]})^T
        \end{eqnarray}

CFRM provides a unified view of spectral clustering and several variants of spectral clustering are shown to be special cases \citep{cfrm}.
They also show that if the association matrices and cluster indicator matrices are optimal solutions to problem  (\ref{eq:cfrmmodel}), then
\begin{equation}
\label{eqn:cfrm:asso}
A^{(m)} = J^{[r_m]^T} X^{(m)} J^{[c_m]}
\end{equation}
which allows us to obtain approximate association matrices from the learnt cluster indicator matrices and input matrices.

\subsection*{Deep Collective Matrix Factorization}

Deep Collective Matrix Factorization (DCMF) \citep{mariappan2018deep} jointly obtains latent representations of each entity $U^{[e]}$ and low-rank factorizations of each matrix $X^{(m)} \approx U^{[r_m]} \cdot U^{{[c_m]}^T}$,
such that $U^{[e]} = f^{[e]}_{\theta}([X]_{e})$
where $f_{\theta}$ is an entity-specific non-linear transformation, obtained through a neural network based encoder with weights $\theta$ and $[X]_{e}$ denotes all matrices in the collection that contains a relationship of entity $E^{[e]}$.

The DCMF network is dynamically constructed using $E$ autoencoders.
The entity instances from all the matrices containing entity $E^{[e]}$ are concatenated and provided as input to the $e^{\rm th}$ autoencoder.
The bottleneck or encoding of each autoencoder, after training, forms the latent factor $U^{[e]}$.
The latent factors are learnt by training all the autoencoders together by minimizing the sum of all the
entity-specific
autoencoder reconstruction losses
and the matrix-specific matrix reconstruction losses.

Collective training of all autoencoders 
induces dependencies between the autoencoder networks that may result in simultaneous under-fitting in some networks and over-fitting in other networks.
This makes collective learning of all latent representations challenging and, to scale to arbitrary collections of matrices, necessitates automatic hyperparameter selection.
(details in \citep{mariappan2018deep}).

\section{Our model: Deep Collective Matrix Tri-factorization (DCMTF)}
\label{dcmtf}

Our model is designed to simultaneously fulfil two objectives.
The first objective is to learn real-valued latent representations of all entities in an input collection of matrices.
The second objective is to learn the cluster structure and the cluster associations based on the learnt latent representations. 

The input to our model consists of a set of $M$ matrices describing relations among $N$ entities.
We denote by $Y^{(m)}_{[e]}$ the $e^{th}$ entity instances in the matrix $X^{(m)}$: when $e = r_m$, $Y^{(m)}_{[e]} = X^{(m)}$ consists of $d_{r_m}$ row vectors and when $e =  c_m$, $Y^{(m)}_{[e]} =  X^{{(m)}^T}$ consists of $d_{c_m}$ column vectors.

\begin{figure*}[!h]
     \centering 
        \subfloat[]{
                \centering
                \def\svgwidth{0.45\textwidth}
                \fontsize{8pt}{8pt}\selectfont
                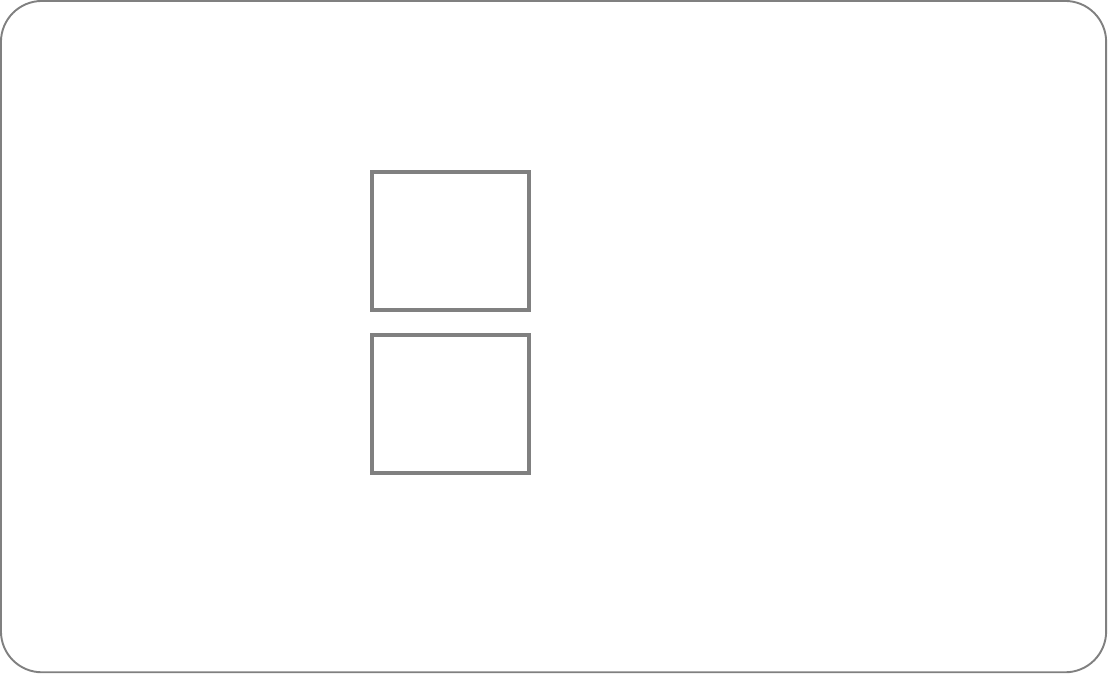
                \label{fig:dcmf_clustering_overview:a}
        }
        \subfloat[]{
                \centering
                \def\svgwidth{0.45\textwidth}
                \fontsize{8pt}{8pt}\selectfont
                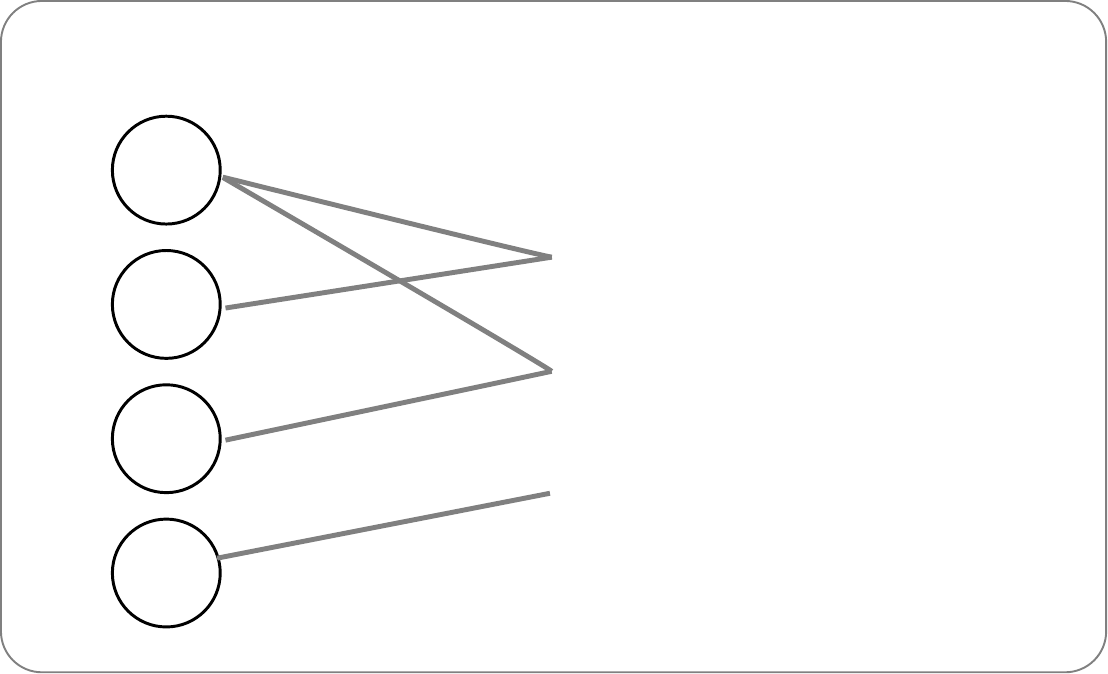
                \label{fig:dcmf_clustering_overview:b}
        } 
        \caption{(a) A collection of views (b) Entity-matrix relationship graph for the setup in fig (a) [square nodes: matrices, circular nodes: entities]}  
        \label{fig:dcmtf:sample} 
\end{figure*}

We assume another input that associates the $N$ entities with the $M$ matrices through a bipartite entity-matrix relationship graph $G(V_E, V_M, Q)$, 
where vertex sets $V_E, V_M$ represent entities and matrices respectively.
We use the same notation -- $X^{(m)}, E^{[e]}$ -- to denote the vertices corresponding to the $m^{\rm th}$ matrix and $e^{\rm th}$ entity, that should be clear from the context.
Edges $(E^{[r_m]},X^{(m)})$, $(E^{[c_m]},X^{(m)}) \in Q$ associate each matrix with its row and column entities (see fig.
\ref{fig:dcmtf:sample} (a) and (b)). 
The datatype (real or binary) of the matrix is provided as a node attribute. 
Note that the 
edge between the $e^{\rm th}$ entity with the $m^{\rm th}$ matrix is associated with entity instances in the matrix $Y^{(m)}_{[e]}$.
Let $\mathcal{N}[e]$ be the list of matrix indices containing either the row or column entity as $E^{[e]}$, i.e. indices of the neighbors of $E^{[e]}$ in $G$.

We first describe our model for entity representation.
Then, we describe how these representations are used in our clustering model. 
Then we explain our network architecture and finally 
how the network is trained to simultaneously learn the representations and clustering.

\subsection{Representation Learning}

We first obtain matrix-specific entity representations using variational autoencoders that can model different data distributions in the matrices.
These representations are then fused to obtain a single real-valued representation for each entity.

A variational autoencoder (VAE) consists of two parameterized functions, $f_{\epsilon}, f_{\delta}$ corresponding to the encoder and decoder respectively.
The encoder outputs the parameters of the learnt distribution: mean and covariance for Gaussian distribution (for real-valued data), and probabilities for Bernoulli distribution (for binary data). Let $\mu_{\epsilon}$ denote the mean of the distribution learnt through the encoder that is used as the VAE representation.

For each $Y_{[e]}^{(m)}$, we obtain, through separate VAEs, the representations:
$$\mu_{\epsilon}^{(e,m)} = f_{\epsilon^{(e,m)}}(Y_{[e]}^{(m)}).$$
If the same entity is present in more than a single matrix, i.e., if the $e^{\rm th}$ entity vertex in $G$
has degree $\mathcal{d}(E^{[e]}) > 1$, then the final entity-specific representation is obtained by 
$$U^{[e]} = f_{\eta^{[e]}} ( \Gamma_{\ell \in \mathcal{N}[e]}
[ \mu_{\epsilon}^{(e,\ell)} ]  ).$$
where $\Gamma_{\ell}[.]$ represents the concatenation operation and the index $\ell$ iterates through the indices of all the matrices containing the $e^{\rm th}$ entity (i.e., neighbors of $E^{[e]}$ in $G$), for which VAE representations are obtained.
If the entity is present in a single matrix, no fusion is required and we set $U^{[e]} = \mu_{\epsilon}^{(e,m)}$.

The functions are realized through neural networks (described in the following sections) and $\epsilon, \delta, \eta$ represent the parameters of the networks. 

\subsection{Multi-way Spectral Clustering}

First we define a similarity metric and prove a technical lemma that shows the equivalence between collective matrix tri-factorization and a trace minimization formulation using the similarity metric.

\subsubsection*{A Similarity Metric}

We define 
$$
P^{(e,m)} = \begin{cases}
X^{(m)} I^{[c_m]} \quad \text{ if } \quad  e = r_m\\
X^{(m)^T} I^{[r_m]} \quad \text{ if } \quad e = c_m\\
\end{cases}
$$

For the $i^{\rm th}$ and $j^{\rm th}$ instances of entity $E^{[e]}$, we define the similarity metric using a Gaussian kernel:
        	        \begin{eqnarray} \label{eqn:dcmtf:s}
                        \mathcal{S}^{[e]}_{ij} = \text{exp}\Bigg(\frac{-||P^{(e,m)}_i - P^{(e,m)}_j||^2}{2\sigma^2}\Bigg) 
                	\end{eqnarray}
            	    where $\sigma > 0$ is the scale hyperparameter. 
                    
\subsubsection*{Clustering through Matrix Tri-factorization}

Consider the trace minimization problem
\begin{equation}
\label{eq:cfrmtrace}
\min \sum_{e=1}^{e=N} Tr\big(J^{[e]^T} L^{[e])} J^{([e])}\big) \text{ subject to } J^{[e])^T}J^{[e])} = \mathbf{I}_{k_e}
\end{equation}

where $J^{[e]}$ is a vigorous cluster indicator (equation \ref{eqn:cfrm:constraint:vigorous}).
$L^{[e]}$ is a Laplacian defined using the similarity metric $\mathcal{S}^{[e]}$:
$$
L^{[e]} = D^{[e]} - S^{[e]}; \quad
D^{[e]} = \text{diag}\big(\sum_{\textit{row}}\mathcal{S}^{[e]} \big)
$$

Following the usual relaxation adopted in spectral clustering, we let $C^{[e]}$ be any real-valued orthonormal matrix and use this trace minimization formulation:

\begin{equation}
\label{eq:cfrmtracedcmtf}
\min \sum_{i=1}^{i=E} Tr\big(C^{[e]^T} L^{[e]} C^{[e]}\big) \text{ subject to } C^{[e]^T}C^{[e]} = \mathbf{I}_{k_e}
\end{equation}

Once learnt, the final binary cluster indicator $I^{[e]} \in \{0,1\}^{d_{e} \times k_e}$ by clustering $C^{[e]}$ using k-means. 
Also, as shown in \citep{cfrm}, the cluster associations for the $m^{\rm th}$ matrix $X^{(m)}$ can be obtained from the learnt $C^{[e]}$: $A^{(m)}$ = $C^{[r_m]^T} \cdot X^{(m)} \cdot C^{[c_m]}$.

We learn $C^{[e]}$ through a neural network (representing the function $g$ with parameters $\gamma$) that ensures orthogonality (described in the following section):
$$C^{[e]} = g_{\gamma^{[e]}} (U^{[e]} ) \text{ such that } C^{[e]^T}C^{[e]} = \mathbf{I}_{k_e}$$ 

The entity-specific representations are also used in defining the similarity metric.
In the definition of 
$S^{[e]}$, we do not use the input matrices $X^{(m)}$.
Instead we use $X^{(m)^{\prime\prime}}$ = $U^{[r_m]}U^{[c_m]^T}$.
This allows us to use our latent representations,  collectively learnt from all the input matrices, to compute the similarity.

\subsection{Dynamic Network Construction}

The functions $f,g$ are realized using multiple neural networks.
The entire network architecture is dynamically constructed since the number of networks depends on the number of input matrices and entities.
The input bipartite entity-matrix relationship graph $G$ is used to determine the relations.

    \begin{itemize}
        \item 
        A VAE is constructed for each edge in $G$, realizing the encoder and decoder functions $f_{\epsilon^{(e,m)}}, f_{\delta^{(e,m)}}$, to obtain 
$\mu_{\epsilon}^{(e,m)}$.   
We denote this network by $\mathcal{A}^{(e,m)}$     
		\item
		For each entity $E^{[e]}$ with degree greater than 1, a feedforward network is constructed to realize $f_{\eta^{[e]}}$. The input to this network is $\Gamma_{\ell}[ \mu_{\epsilon}^{(e,\ell)} ] $, the  concatenation of all the mean vectors from the VAEs corresponding to the $e^{\rm th}$ entity. 
We denote this network by $\mathcal{F}^{[e]}$.
        \item 
        For each entity $E^{[e]}$ a feedforward network is created to realize $g_{\gamma^{[e]}}$. The last layer of this network is designed to orthogonalize the outputs, as described below.
We denote this network by $\mathcal{C}^{[e]}$.
    \end{itemize}

The hyperparameters activation functions, scaling function, mini batch size are data-specific i.e. selected based on the data's range, type and size of the matrices. 
Other hyperparameters, such as learning rate, weight decay of the SGD algorithm, convergence threshold, number of hidden layers are set through automated hyperparameter tuning.
The weights of the network are trained using gradient descent as described below.
Algorithm \ref{algo:dcmtf} summarizes the entire process of model construction.

\subsection{Network Training}

\subsubsection*{Loss function}

The networks  $\mathcal{A}^{(e,m)}, \mathcal{F}^{[e]},\mathcal{C}^{[e]}$ are trained collectively to minimize 
$$\min_{\epsilon, \delta,\gamma, \eta}
\sum\limits_{{e,m}} \mathcal{L}_\mathcal{A}^{(e,m)}
+
\sum\limits_{{e = 1}}^{N} \mathcal{L}_\mathcal{C}^{[e]} + 
\sum\limits_{m = 1}^{M} \mathcal{L}_\mathcal{R}^{(m)} 
$$

The summation in the VAE reconstruction loss 
$\sum\limits_{{e,m}} \mathcal{L}_\mathcal{A}^{(e,m)}$
iterates over all the edges in $G$ (thus, over all pairs $(e,m)$). 
The loss 
$
\mathcal{L}_\mathcal{A}^{(e,m)}
$
is set to be binary cross entropy for binary inputs and mean square error for real input matrices,
as described in \citep{kingma2013auto}.
The datatype is obtained from the node attribute of the matrix vertices in $G$. 

The clustering loss $ \sum\limits_{{e = 1}}^{N} \mathcal{L}_\mathcal{C}^{[e]}$ 
effects tri-factorization through trace minimization:
$$\mathcal{L}_\mathcal{C}^{[e]} =  Tr\big(C^{([e])^T} L^{[e]} C^{[e]}\big)$$

The matrix-specific reconstruction loss
$\sum\limits_{m = 1}^{M} \mathcal{L}_\mathcal{R}^{(m)} $ 
enforces a factor interpretation on the latent representations used within the similarity metric:
$$ \mathcal{L}_\mathcal{R}^{(m)} = \ell_\mathcal{R}^{[m]} (X^{(m)}, U^{[r_m]}{U^{[c_m]}}^T)$$
where $\ell_\mathcal{R}^{(m)}$ 
is set to binary cross entropy for binary inputs and Frobenius norm of the error for real input matrices.

\subsubsection*{Optimization}

Optimization is done in a coordinate descent manner, using stochastic gradient descent (SGD), where we iteratively train $\mathcal{A}^{(e,m)}, \mathcal{F}^{[e]}$ (phase 1) followed by
$\mathcal{A}^{(e,m)}, \mathcal{C}^{[e]}$ (phase 2).

In phase 1, there are two steps. In the first step, we update the weights of the VAE: $\epsilon^{(e, m)}, \delta^{(e, m)}$ by backpropagating $\mathcal{L}_{\mathcal{A}}^{(e,m)}$.
In the second step, we update the weights of the fusion network and encoder: $\eta^{[e]}, \epsilon^{(e,m)}$ by backpropagating $\mathcal{L}_{\mathcal{R}}^{(m)}$.

\begin{figure*}[h]
     \centering 
        \subfloat[]{
                \centering
                \def\svgwidth{1.0\textwidth}
                \fontsize{8pt}{8pt}\selectfont
                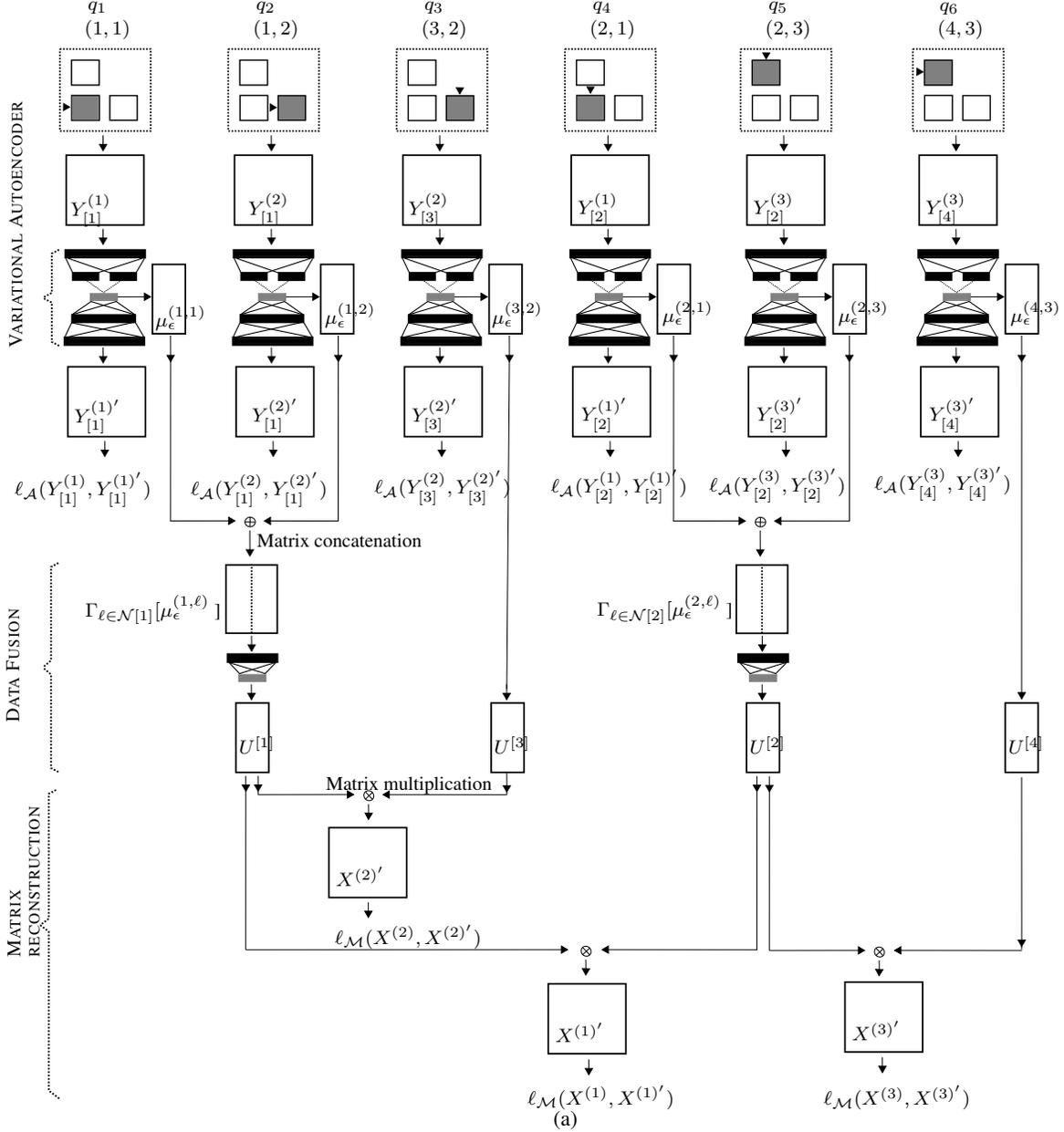 
        }
        \caption{(a) Pass 1: DCMTF Training}
        \label{fig:dcmtf:pass1}
\end{figure*}

\begin{figure*}[h]
     \centering 
        \subfloat[]{
                \centering
                \def\svgwidth{1.0\textwidth}
                \fontsize{6pt}{6pt}\selectfont
                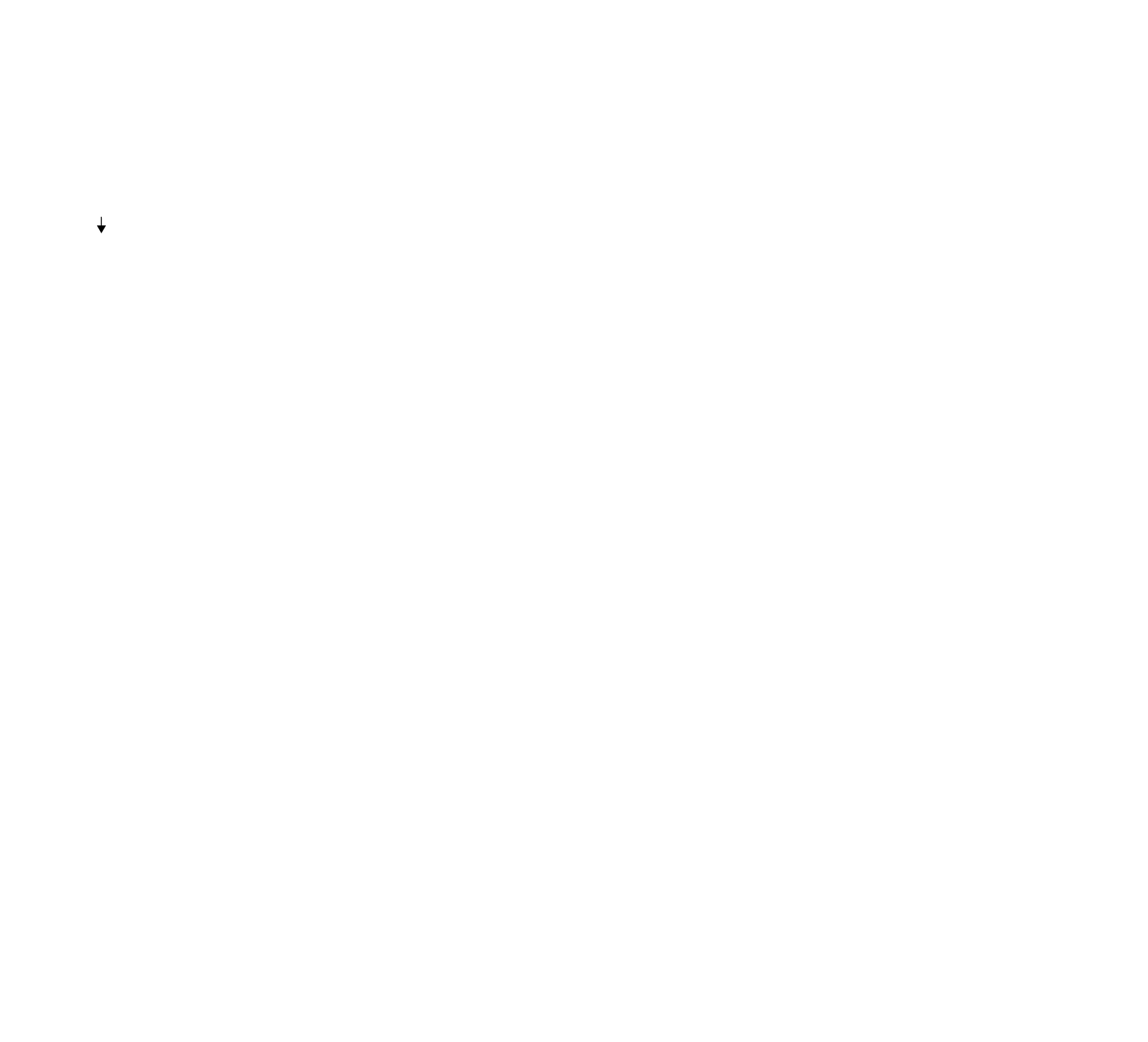 
        } 
        \caption{(a) Pass 2: DCMTF Training}
        \label{fig:dcmtf:pass2}
\end{figure*}

In phase 2, there are three steps.
In the first step, we update the weights of the VAE: $\epsilon^{(e, m)}, \delta^{(e, m)}$ by backpropagating $\mathcal{L}_{\mathcal{A}}^{(e,m)}$.
To ensure orthogonal outputs $C^{[e]}$ in the clustering network, we follow the procedure described in \citep{shaham2018spectralnet}.
During forward propagation through $\mathcal{C}^{[e]}$, the input to the last layer $\widetilde{C}^{[e]}$ is used to perform Cholesky decomposition:
$$\widetilde{H}^{[e]}
\widetilde{H}^{{[e]}^T}
= \text{Cholesky}(
\widetilde{C}^{[e]}
\widetilde{C}^{{[e]}^T}
) $$
The weight of the final linear layer is fixed to 
$\widetilde{H}^{{[e]}^{{-1}^T}}$, which yields the output:
$$
C^{[e]} = \widetilde{C}^{[e]} \widetilde{H}^{{[e]}^{{-1}^T}}$$ 
In the final step of phase 2, we update the weights of the clustering network and encoder: $\gamma^{(i)}, \epsilon^{(e,m)}$ by backpropagating $\mathcal{L}_{\mathcal{C}}^{[e]}$.
The complete procedure is summarized in algorithms \ref{algo:dcmtf},
\ref{algo:train},
\ref{algo:train:pass1} and \ref{algo:train:pass2}. Figures (\ref{fig:dcmtf:pass1}) and (\ref{fig:dcmtf:pass2}) illustrate the passes 1 and 2 respectively for the sample setting with 3 matrices in Figure (\ref{fig:dcmtf:sample}).

\begin{algorithm}[!h]
    \caption{Deep Collective Matrix Trifactorization}
    \label{algo:dcmtf}
    \SetKwInOut{Input}{inputs}
    \SetKwInOut{Output}{outputs}
    \SetKwProg{DCMTF}{DCMTF}{}{}
    \DCMTF{$(G,\mathcal{X})$}{

        \Input{
        Entity-matrix relationship graph $G(V_E,V_M,Q)$,  
        Input matrices $\mathcal{X} = {X^{(1)},...,X^{(M)}}$
        }

        \Output{
        Entity representations $\mathcal{U}=U^{(1)},...,U^{(E)}$,  
        Entity cluster indicators $\mathcal{I}=I^{(1)},...,I^{(E)}$, \\
        Cluster associations $\mathcal{T} = {A^{{(1)}^{\prime}},...,A^{{(M)}^{\prime}}}$
        }
        
        $\vec{p}_{t}$ $\gets$ Initialize DCMTF Hyperparameters \\
        
        \ForEach{BO iteration \lq{t}\rq}{
            \tcp{Input Transformation}
            \ForEach{\textit{edge} $(e,m) \in Q$}{
                \eIf{$e$ == $r_m$}{
                    $Y^{(m)}_{[e]}$ = $X^{(m)^T}$
                }{
                    $Y^{(m)}_{[e]}$ = $X^{(m)}$
                }
            }
            
            \tcp{Network Construction with hyperparams $\vec{p}_{t}$}
            \ForEach{\textit{edge} $(e,m) \in Q$ }
            {
                 Construct \textit{vae-network} $\mathcal{A}^{(e,m)}$
            }
        
            \ForEach{\textit{entity} $e \in V_E$ }
            {
                \If{\textit{degree}($G$,$e$) $>$ 1}{
                    Construct \textit{fusion-network} $\mathcal{F}^{[e]}$
                 }
                 Construct \textit{clustering-network} $\mathcal{C}^{[e]}$
            }
        
           Construct network $\mathcal{NN}$ with $\mathcal{A}^{(e,m)}$,  $\mathcal{F}^{[e]}$, $\mathcal{C}^{[e]}$ 
            
            \tcp{Train the network $\mathcal{NN}$ with Algorithm \ref{algo:train} and obtain the loss $\mathcal{L}$}
            
            $\mathcal{L}$, $\gamma^{[e]}, \eta^{[e]}, \epsilon^{(e,m)}$ = \textbf{TRAIN}($\mathcal{NN}$, $\vec{p}_{t}$) \\
            \tcp{Obtain the hyperparameters for next iteration using Bayesian Optimization}
            $\vec{p}_{t+1}$ $\gets$ \textbf{BO}($\mathcal{L}$, $\vec{p}_{t}$)
        }
        
        \tcp{Obtain network parameters corresponding to the best hyperparameters $\vec{p}_{\#}$ }
        $\gamma^{[e]}_{^\#}, \eta^{[e]}_{^\#}, \epsilon^{(e,m)}_{^\#}$ $\gets$ \textbf{BO}($\vec{p}_{\#}$) \\

        \tcp{Entity representation generation using the trained network}
        \ForEach{\textit{edge} $(e,m) \in Q$ }{
            $\mu_{\epsilon}^{(e,m)}$ = $f_{\epsilon^{(e,m)}_{\#}}(Y_{[e]}^{(m)})$
        }
        \tcp{Data Fusion using the trained network}
        \ForEach{\textit{entity} $e$ in $V_E$}
        {
            \eIf{\textit{degree}(G,$e$) $>$ 1}{
                $U^{[e]} = f_{\eta^{[e]}_{\#}} ( \Gamma_{\ell \in \mathcal{N}[e]}[ \mu_{\epsilon}^{(e,\ell)} ]  )$
            }{
                $U^{[e]} = \mu_{\epsilon}^{(e,m)}$
            }
        }
        
        \tcp{Clustering using the trained network}
        \ForEach{\textit{entity} $e$ in $V_E$}{%
             $C^{[e]} = g_{\gamma^{[e]}_{\#}} (U^{[e]}); $\quad $I^{[e]}$ = K-means($C^{[e]}$) \\
             $J^{[e]}$ = vigorous($I^{[e]}$) \tcp{Convert $I^{(e)}$ to vigorous cluster indicator matrix using  (\ref{eqn:cfrm:constraint:vigorous})} 
        }
        \ForEach{\textit{entity} $e_i$ in $V_E$}{%
             $\widetilde{C^{(i)}}$ = $\mathcal{S}^{(i)}_{{\textbf{p}^{\#},\theta^{\#}}}$($U^{(i)}$) \\
             $C^{(i)}$ = Orthogonalize($\widetilde{C^{(i)}}$)
        }
        
        \tcp{Cluster associations}
        \ForEach{matrix $m$ in $V_M$ }{%
            $A^{(m)} = J^{[r_m]^T} X^{(m)} J^{[c_m]}$
        }

        }
        \KwRet{$\mathcal{U}$, $\mathcal{I}$, $\mathcal{T}$}
\end{algorithm}

 \begin{algorithm}[!h]
    \SetKwInOut{Input}{inputs}
    \SetKwInOut{Output}{outputs}
    \SetKwProg{train}{TRAIN}{}{}
    \train{($\mathcal{NN}$, $\vec{p}$)}{
        \Input{\text{DCMTF Network} $\mathcal{NN}$, \text{Hyperparameters} $\vec{p}$}
        \Output{\text{Trained Network Parameters} $\gamma^{[e]}, \eta^{[e]}, \epsilon^{(e,m)}$ }
                \tcp{Pass 1 with Algorithm \ref{algo:train:pass1}}
                $\mathcal{L}_1$ $\gets$ \textbf{PASS1}($\mathcal{NN}$, $\vec{p}_{n}$) \\
                Backpropagate $\mathcal{L}_1$ and update weights  $\epsilon^{(e,m)}$, $\eta^{[e]}$ of $\mathcal{A}^{(e,m)}$ and $\mathcal{F}^{[e]}$ respectively\\
                \tcp{Pass 2 with Algorithm \ref{algo:train:pass2}}
                $\mathcal{L}_2$ $\gets$ \textbf{PASS2}($\mathcal{NN}$, $\vec{p}_{n}$) \\
                Backpropagate $\mathcal{L}_2$ and update weights $\epsilon^{(e,m)}$, $\gamma^{[e]}$ of $\mathcal{A}^{(e,m)}$ and $\mathcal{C}^{[e]}$ respectively\\
        }
        \KwRet{$\gamma^{[e]}, \eta^{[e]}, \epsilon^{(e,m)}$}
    \caption{DCMTF Training }
    \label{algo:train}
\end{algorithm}

\begin{algorithm}[!h]
    \SetKwInOut{Input}{inputs}
    \SetKwInOut{Output}{outputs}
    \SetKwProg{pass}{PASS1}{}{}
    \pass{($\mathcal{NN}$, $\vec{p}$)}
    {
        \Input{\text{DCMTF Network} $\mathcal{NN}$, \text{Hyperparameters} $\vec{p}$}
        \Output{loss $\mathcal{L}_1$}
        \tcp{Step 1: Entity representation generation}
        \ForEach{\textit{edge} $(e,m) \in Q$ }{
            $\mu_{\epsilon}^{(e,m)}$, $Y_{[e]}^{(m)^\prime}$ = $\mathcal{A}_{\epsilon}^{(e,m)}(Y_{[e]}^{(m)})$ \\
            $\mathcal{L}_\mathcal{A}^{(e,m)}$ =  $\ell_{\mathcal{A}}(Y_{[e]}^{(m)^\prime}$,$Y_{[e]}^{(m)})$
        }
        \tcp{Data Fusion}
        \ForEach{\textit{entity} $e$ in $V_E$}{%
            \eIf{\textit{degree}(G,$e$) $>$ 1}{
                $U^{[e]} = f_{\eta^{[e]}} ( \Gamma_{\ell \in \mathcal{N}[e]}[ \mu_{\epsilon}^{(e,\ell)} ]  )$
            }{
                $U^{[e]} = \mu_{\epsilon}^{(e,m)}$
            }
        }
        
        \tcp{Step 2: Matrix Reconstruction}
        \ForEach{\textit{matrix} $m$ in $V_M$ }{%
            $X^{(m)\prime}$ = $U^{[r_m]}{U^{[c_m]}}^T$ \\
            $\mathcal{L}_\mathcal{R}^{(m)} = \ell_\mathcal{R}^{[m]} (X^{(m)}, X^{(m)\prime})$ 
        }
        
        \tcp{Pass 1 Loss $\mathcal{L}_1$}    
        $\mathcal{L}_\mathcal{A}$ = $\sum\limits_{(e,m)}  \mathcal{L}_\mathcal{A}^{(e,m)}$ 
        
        $\mathcal{L}_\mathcal{R}$ = $\sum\limits_{m}\mathcal{L}_\mathcal{R}^{(m)}$
    
        $\mathcal{L}_1 = \mathcal{L}_\mathcal{A} + \mathcal{L}_\mathcal{R}$ \\

        \KwRet{$\mathcal{L}_1$}
    }
    \caption{Pass 1 of DCMTF Training}
    \label{algo:train:pass1}
\end{algorithm}

\begin{algorithm}[!h]
    \SetKwInOut{Input}{inputs}
    \SetKwInOut{Output}{outputs}
    \SetKwProg{pass}{PASS2}{}{}
    \pass{($\mathcal{NN}$, $\vec{p}$)}
    {
        \Input{\text{DCMTF Network} $\mathcal{NN}$, \text{Hyperparameters} $\vec{p}$}
        \Output{loss $\mathcal{L}_2$}
        \tcp{Step 1: Entity representation generation}
        \ForEach{\textit{edge} $(e,m) \in Q$ }{
            $\mu_{\epsilon}^{(e,m)}$, $Y_{[e]}^{(m)^\prime}$ = $\mathcal{A}_{\epsilon}^{(e,m)}(Y_{[e]}^{(m)})$ \\
            $\mathcal{L}_\mathcal{A}^{(e,m)}$ =  $\ell_{\mathcal{A}}(Y_{[e]}^{(m)^\prime}$,$Y_{[e]}^{(m)})$
        }

        \tcp{Step 2: Data Fusion}
        \ForEach{\textit{entity} $e$ in $V_E$}{%
            \eIf{\textit{degree}(G,$e$) $>$ 1}{
                $U^{[e]} = f_{\eta^{[e]}_{\#}} ( \Gamma_{\ell \in \mathcal{N}[e]}[ \mu_{\epsilon}^{(e,\ell)} ]  )$
            }{
                $U^{[e]} = \mu_{\epsilon}^{(e,m)}$
            }
        }
        
        \tcp{Step 3: Clustering}
        \ForEach{\textit{entity} $e$ in $V_E$}{%
            $C^{[e]} = g_{\gamma^{[e]}_{\#}} (U^{[e]})$ \\
            $\mathcal{L}_\mathcal{C}^{(e)}$ = Tr($C^{{[e]}^{T}} \cdot L^{[e]]} \cdot C^{[e]}$)
        }

        \tcp{Pass 2 Loss $\mathcal{L}_2$}    
        $\mathcal{L}_\mathcal{A}$ = $\sum\limits_{(e,m)}  \mathcal{L}_\mathcal{A}^{(e,m)}$ 
        

        $\mathcal{L}_\mathcal{C}$ = $\sum\limits_{e}\mathcal{L}_\mathcal{C}^{[e]}$
    
        $\mathcal{L}_2 = \mathcal{L}_\mathcal{A} + \mathcal{L}_\mathcal{C} $ \\

        \KwRet{$\mathcal{L}_2$}
    }
    \caption{Pass 2 of DCMTF Training}
    \label{algo:train:pass2}
\end{algorithm}

Our training procedure iteratively minimizes the objective $\sum_{e} Tr(C^{[e]^T} L^{[e]} C^{[e]})$ with respect to each $J^{[e]}$ (relaxed to $C^{[e]}$).
This individual trace minimization is equivalent to spectral clustering, as shown in \citep{cfrm,shaham2018spectralnet}. 
While both CFRM and DCMTF use underlying principles of spectral clustering, 
DCMTF has enhanced modeling capabilities, by design, due to 
(i) simultaneous representation learning and clustering using neural networks, where cluster-aware entity representations ($U^{[e]}$) are learnt from all matrices collectively (ii) a similarity metric ($\mathcal{S}_{P}^{[e]}$) based on neural representations and computed during network training ($P^{(e,m)} = X^{{(m)\prime}^T} J^{[r_m]}$, where $X^{(m)^{\prime}} = U^{[r_m]}U^{[c_m]^T}$)
and (iii) ability to handle mixed datatypes (through VAEs).

\subsection{Simulation Study}

\begin{figure*}[h] 
\centering
    \subfloat[]{
    \centering
            \def\svgwidth{0.25\textwidth} 
            \fontsize{8pt}{8pt}\selectfont
            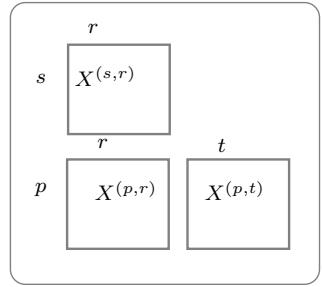
    }
    \caption{
    Schematic of synthetic dataset
    }
\label{fig:syn_data_1:setting}
\end{figure*}

\noindent
\textbf{Data.} We generated a synthetic dataset with 3 matrices and 4 entities as shown in Figure \ref{fig:syn_data_1:setting}. We synthetically generated 400 instances of entity $p$, 200 instances of entity $r$ and 240 instances of entities $d$
in the following manner. 

\begin{figure*}[h] 
\centering
	\subfloat[Entity-specific Cluster Indicators: $I^{(e)}$]{
	\centering
	\includegraphics[width=0.75\linewidth]{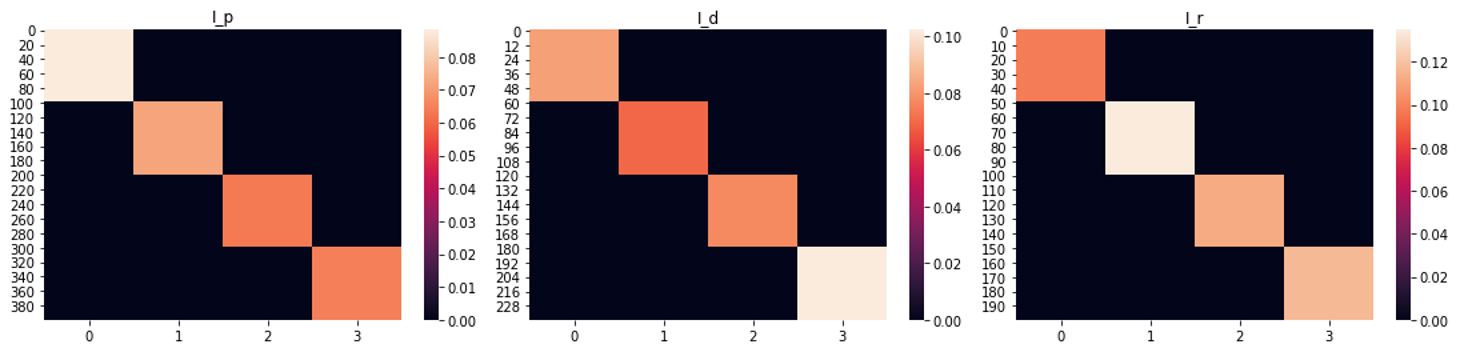}
	\label{}
	}
	
	\subfloat[Cluster Associations across Entity Pairs: $A^{(m)}$]{
	\centering
	\includegraphics[width=0.75\linewidth]{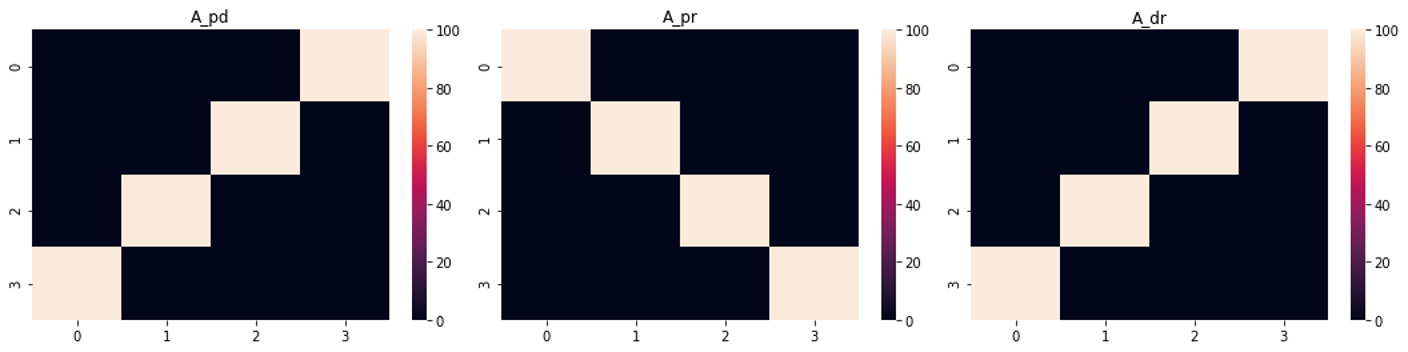}
	\label{}
	}

	\subfloat[Data matrices $X^{(m)}$]{
	\centering
	\includegraphics[width=0.5\linewidth]{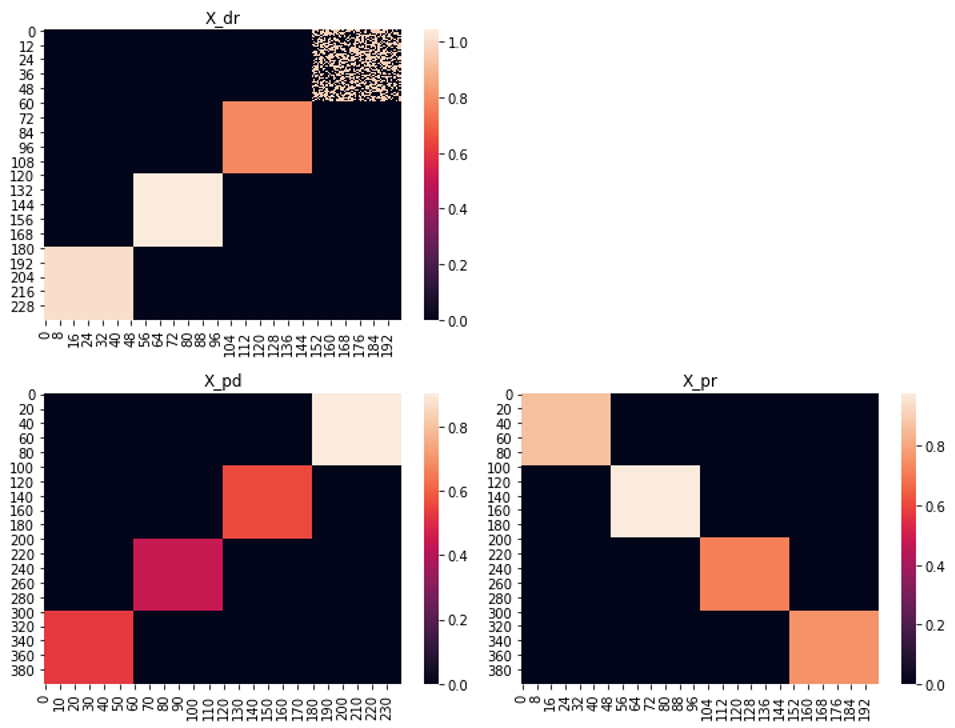}
	\label{}
	}
	\subfloat[Data matrices $X^{(m)}$ - rows and columns permuted]{
	\centering
	\includegraphics[width=0.5\linewidth]{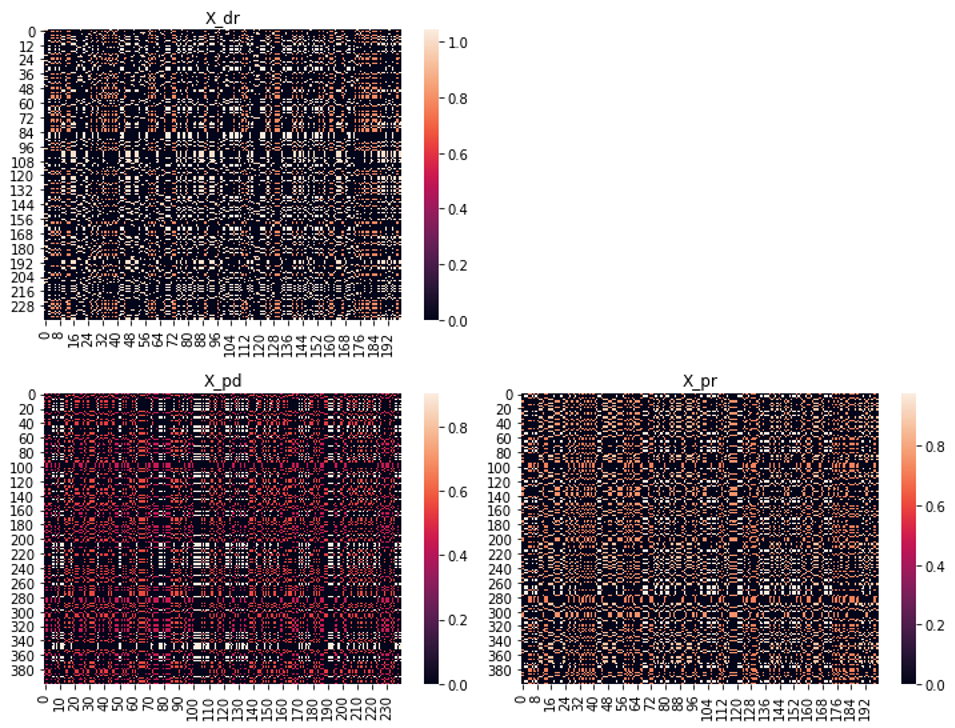}
	\label{}
	}
	\caption{Synthetic Dataset Generation}
\label{fig:syn_data_1}
\end{figure*}

\begin{itemize}
    \item We generate entity-specific cluster indicator matrices $I^{(d)}, I^{(r)}, I^{(p)}$ with the number of clusters $k_e$ = 4 as show in figure \ref{fig:syn_data_1}(a), where $I \in \{0,1\}^{d_e \times k_e}$. The $I$'s are then transformed to a vigorous cluster indicator matrix i.e. an orthogonal matrix with the columns normalized. 

    \item We also generate entity cluster association matrices $A^{(p,r)},A^{(p,t)},A^{(s,r)}$ as show in figure \ref{fig:syn_data_1}(b), where $A \in \{0,100\}^{k_{r_m} \times k_{c_m}}$. The entries indicating strong associations were filled with 100 and the rest of the entries are set to 0. 

    \item We generate data matrices $X^{(r_m,c_m)}$ by computing the product of the factors $I^{(r_m)} . A^{(r_m,c_m)} . I^{(c_m)}$. Thus we generate the matrices $X^{(p,r)}$, $X^{(p,t)}$ and $X^{(s,r)}$ as show in figure \ref{fig:syn_data_1}(c). 
    \item
    The input to DCMTF are the matrices in Figure \ref{fig:syn_data_1}(d) which are obtained by permuting the rows and columns of each matrix in Figure \ref{fig:syn_data_1}(c).
\end{itemize}

\begin{figure*}[h] 
\centering
	\subfloat[Entity Cluster Indicator $I^{(e)}$, obtained by Clustering the corresponding Entity Spectral Representation $C^{(e)}$]{
	\centering
	\includegraphics[width=0.75\linewidth]{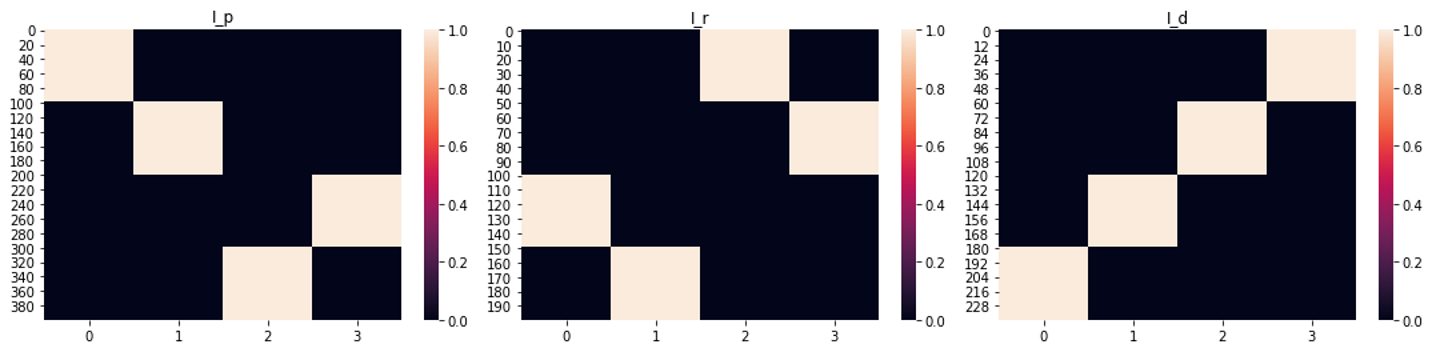}
	\label{}
	}
	
	\subfloat[Learnt Entity Cluster Associations: $A^{(m)}$]{
	\centering
	\includegraphics[width=0.75\linewidth]{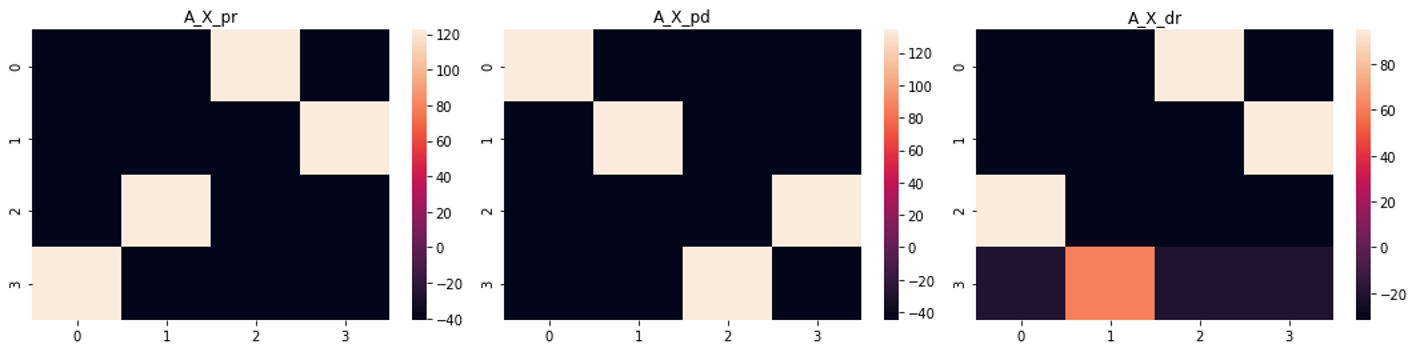}
	\label{}
	}

	\subfloat[Predicted Data matrices $X^{(m)}$]{
	\centering
	\includegraphics[width=0.5\linewidth]{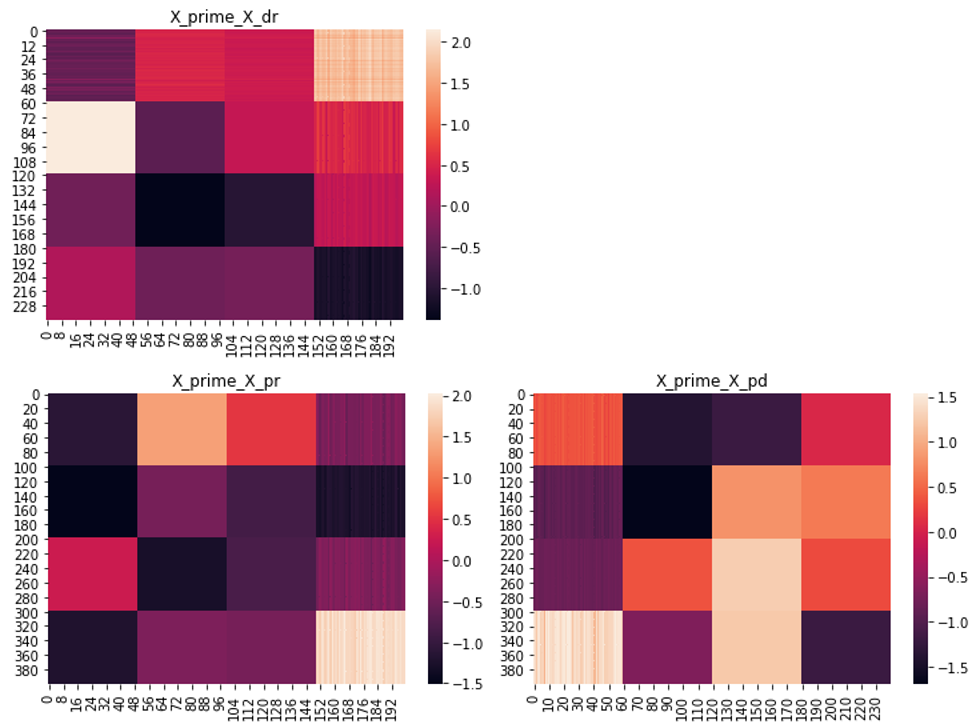}
	\label{}
	}
	\subfloat[Predicted Data matrices $X^{(m)}$ - rows and columns permuted]{
	\centering
	\includegraphics[width=0.5\linewidth]{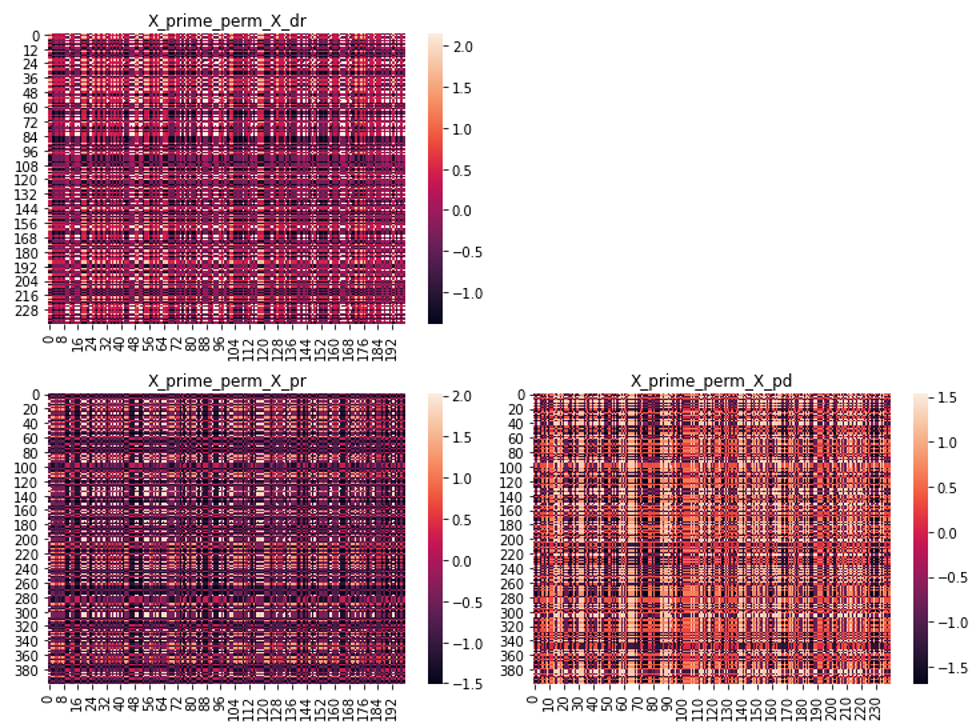}
	\label{}
	}

	\subfloat[Learnt Entity Spectral Representations $C^{(e)}$]{
	\centering
	\includegraphics[width=0.75\linewidth]{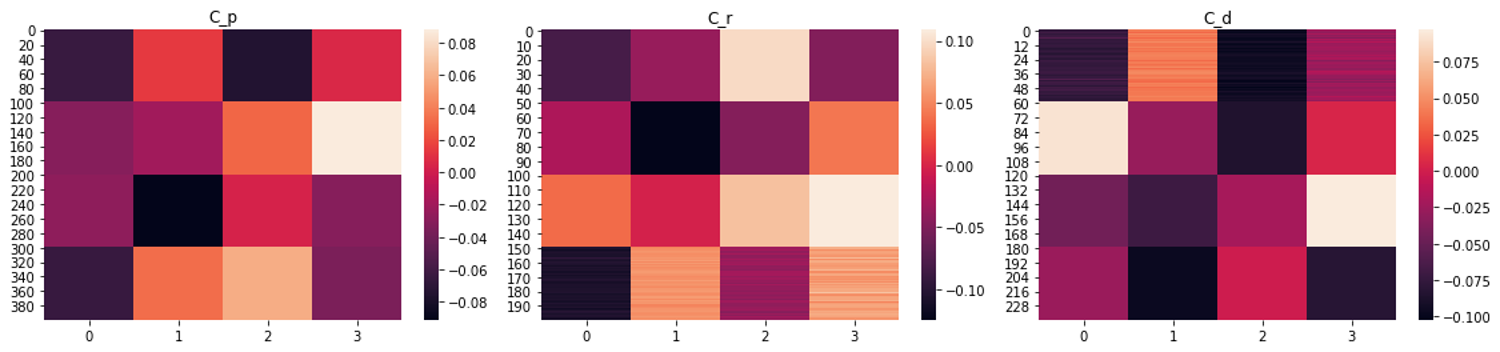}
	\label{}
	}	
	\caption{DCMTF Results on Synthetic dataset - Predicted $I^{(e)}$, $A^{(m)}$, $C^{(e)}$ and $X^{(m)^\prime}$}
\label{fig:syn_data_1_res}
\end{figure*}

\noindent
\textbf{Results.} 
The DCMTF outputs viz. the entity clusterings $I^{(e)}$, the strength of association $A^{(m)}$ across pairs of entities, the Spectral entity representations $C^{(e)}$ and the reconstructed matrices $X^{(m)^\prime}$ are shown in Figure \ref{fig:syn_data_1_res}(a), \ref{fig:syn_data_1_res}(b), \ref{fig:syn_data_1_res}(d) and 
\ref{fig:syn_data_1_res}(e) respectively. Note that the matrices in Figure \ref{fig:syn_data_1_res}(c) are the predicted matrices in \ref{fig:syn_data_1_res}(d) after reverting the permutations of the rows and columns. It can be seen that the Clusters and Cluster associations are learnt correctly by DCMTF. 

\section{Experiments on Real Data}

\subsection{Data and Evaluation Metrics}
\begin{figure}[h]
    \centering
    \includegraphics[width=0.4\textwidth]{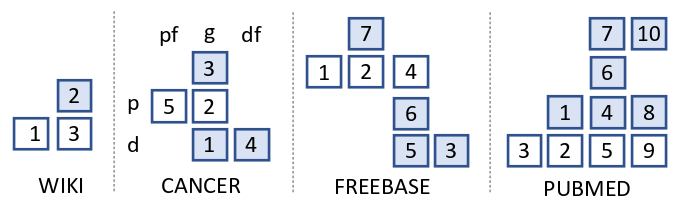}
    \caption{Schematic of datasets used (details in Appendix \ref{app:data_details})}
    \label{fig:data_schematic}
\end{figure}

We use four real datasets that have 3,5,7 and 10 matrices, a schematic is shown in fig. \ref{fig:data_schematic}.
In each case, the task is to cluster the row entity shared among the unshaded matrices, which form a multi-view setup. 
The shaded matrices are auxiliary data matrices.
Consider, e.g., the Cancer dataset:
The row entity patients (p) has patient features (pf) in matrix 5 and genetic data (g) in matrix 2.
Matrices 1, 3 and 4 contain gene(g)-disease(d) associations, gene-gene interactions and disease features (df) respectively. 
The ground truth labels and other dataset details such as dimensions, sparsity level and sources are in Appendix \ref{app:data_details}.
Most matrices are more than 99\% sparse, with more than 1000 instances in most entities. 
Standard clustering metrics are reported: Rand Index (RI), Normalized Mutual Information (NMI) along with chance-adjusted metrics Adjusted Rand Index (ARI) and Adjusted Mutual Information (AMI).

\subsection{Clustering Performance}

In the augmented multi-view setting, {\it all} the matrices are clustered and the clustering on the chosen entity is evaluated. 
As baselines we use \textbf{CFRM} \citep{cfrm} and \textbf{DFMF} \citep{zitnik2015data}.
The latter learns representations which are clustered using K-Means.
In the multi-view setting, not all matrices can be used and we consider only the unshaded matrices as inputs.
Baselines in this setting include Multi-view Spectral Clustering Network (\textbf{MvSCN}) \citep{huang2019multi}, and methods which concatenate the input matrices and cluster -- SpectralNet (\textbf{SNet}) \citep{shaham2018spectralnet} and K-Means.
DCMTF is run on both settings.
We also run K-Means (\textbf{KM}), Spectral Clustering (\textbf{Spectr}) and SpectralNet on each (unshaded) matrix within the multi-view setup individually as baselines.
E.g., to cluster patient instances, in the Cancer dataset, matrices 5 and 2 are individually used in the single-view setting, and both matrices are used together in the multi-view setting, whereas all matrices, 1--5, are used in the augmented multi-view setting.
For each method, we report the result where the best loss is achieved among the random runs tested.


\begin{table*}[!h]
    \centering
    \captionsetup{font=footnotesize}
    \footnotesize
    \begin{tabular}{c|c|p{2.25em}p{1.85em}p{2.75em}|p{2.25em}p{1.85em}p{2.75em}|p{2.25em}p{1.85em}p{2.75em}|p{2.25em}p{1.85em}p{2.75em}}
    & & \multicolumn{3}{c}{\textbf{Wiki}} & \multicolumn{3}{|c}{\textbf{Cancer}} & \multicolumn{3}{|c}{\textbf{Freebase}} & \multicolumn{3}{|c}{\textbf{PubMed}} \\ 
     Setting & Metric & \multicolumn{3}{c}{$M=3, k=3$} & \multicolumn{3}{|c}{$M=5, k=4$} & \multicolumn{3}{|c}{$M=7, k=8$} & \multicolumn{3}{|c}{$M=10, k=8$} \\ 
     \hline
     \multirow{5}{*}{\rotatebox{90}{\parbox[c]{1.5cm}{\centering Augmented Multi-view}}} 
     && DFMF & CFRM & DCMTF & DFMF & CFRM & DCMTF & DFMF & CFRM & DCMTF & DFMF & CFRM & DCMTF \\ 
    &
    ARI &   0.0533  &   0.0103  &   \underline{\textbf{0.4189}}  &   0.0045  &   0.0094  &   \textbf{0.0211}  &   0.0420  &   0.0419  &   \textbf{0.0660}  &   0.0060  &   0.0014  &   \underline{\textbf{0.0200}}  \\
    & AMI &   0.1147  &   0.0156  &   \underline{\textbf{0.3789}}  &   0.0051  &   0.0081  &   \textbf{0.0181}  &   0.0694  &   0.0509  &   \underline{\textbf{0.1473}}  &   0.0055 & $\text{-}$0.0033 &   \underline{\textbf{0.0183}}  \\
    & NMI &   0.1284  &   0.0247  &   \underline{\textbf{0.3796}}  &   0.0105  &   0.0139  &   \textbf{0.0228}  &   0.0830  &   0.0805  &   \underline{\textbf{0.1595}}  &   0.0481  &   0.0451  &   \textbf{0.0516}  \\ 
    & RI  &   0.5290  &   0.4347  &   \underline{\textbf{0.7354}}  &   0.5410  &   0.5422  &  \textbf{0.5478}  &   \underline{\textbf{0.7723}}  &   0.6234  &   0.6608  &   0.4976  &   0.3680  &   \underline{\textbf{0.6334}}  \\
     \bottomrule
       \multirow{5}{*}{\rotatebox{90}{Multi-view}} 
     & & MvSCN & SNet & DCMTF & MvSCN & SNet & DCMTF & MvSCN & SNet & DCMTF & MvSCN & SNet & DCMTF \\ 
   &
    ARI &   0.0072  &   0.0029  &   \textbf{0.2643}  & 
    \textbf{0.0336}  &   0.0044  &   0.0207  &   
    0.0  &   0.0395  &   \underline{\textbf{0.0681}} &  
    0.0006  &   \textbf{0.0073}  &   0.0065  \\
    & AMI &   0.0016  &   0.0023  &   \textbf{0.2913}  &   
    0.0329  &   0.0019  &   \underline{\textbf{0.0421}}  &   
    0.0  &   0.0535  &   \textbf{0.1425}  &  
    \textbf{0.0016}  &  \text{-}.0006 &   -.0102  \\
    & NMI &   0.0045  &   0.0034  &   \textbf{0.2921} &  
    0.0430  &   0.0070  &   \underline{\textbf{0.0467}}  &  
    0.0  &   0.0696  &   \textbf{0.1546}  &   
    \textbf{0.0480}  &   0.0335  &   0.0250  \\ 
    & RI  &  0.4042  &   0.5375  &   \textbf{0.6647}  & 
    \underline{\textbf{0.5479}}  &   0.5413  &   0.5477  &  
    0.1752  &   \textbf{0.7233}  &   0.4189  & 
    0.4015  &   \textbf{0.6276}  &   0.6147  \\
    \bottomrule
    \multirow{5}{*}{\rotatebox{90}{Single-view}} 
     && SNet & Spectr & KM & SNet & Spectr & KM & SNet & Spectr & KM & SNet & Spectr & KM \\
&ARI &   \textbf{0.0336}  &   \text{-}0.043  &  \text{-}0.0011  & \underline{\textbf{0.0499}}  &   0.0317  &  0.0143  & \text{-}0.0009  &   \textbf{0.0016}  & 0.0569 & \textbf{0.0088}  &  0.0041  &  0.0086   \\
&AMI &   \textbf{0.0360}  &    0.0258  &   0.0003 & \textbf{0.0277}  &   0.0238  &  0.0180 & 0.0345  &   \textbf{0.0569}  & 0.1116 & 0.0040  &  0.0012  &  \textbf{0.0084} \\
&NMI &   \textbf{0.0384}  &   0.0272  &   0.0024  &  \textbf{0.0351}  &   0.0306  &  0.0231  &  0.1021 &     0.0735 &  \textbf{0.1280} & \underline{\textbf{0.0560}}  &  0.0481  &  0.0497  \\
&RI  &   \textbf{0.5461}  &   0.4305  &   0.3505  &  \textbf{0.5209}  &   0.4651  &  0.5200  &   0.2491  &    \textbf{0.5562}  &  0.5269 & 0.3888  &  0.2120  & \textbf{0.4382} \\
\bottomrule
    \end{tabular}
    \caption{\centering Clustering Performance. $M$: Number of matrices in collection, $k$: \#clusters for a chosen entity (input); For each dataset and metric: best result in each setting (separately) shown in bold, best result across all settings underlined.}
    \label{tab:clust_perf_all}
\end{table*}

Table \ref{tab:clust_perf_all} shows
the performance of the algorithms in multiple settings.
In the single-view setting, for each baseline, we show the performance on the matrix where the best NMI is achieved (all other results are in Appendix \ref{app:expt_results}).
We observe that SpectralNet outperforms other methods in the single-view setting, in most cases.
When additional data is used in the multi-view setting, performance does not always improve for all methods, e.g., in most cases, MvSCN results are worse than the best single-view results.
Results for K-Means and Spectral Clustering in multi-view setting are in Appendix \ref{app:expt_results}-- their performance is comparable, and worse than that of SpectralNet in most cases.
Although SpectralNet is not designed for multi-view clustering, with concatenated inputs, it outperforms MvSCN on 2 datasets on some metrics.
With DCMTF, performance improves in the multi-view setting compared to the best results in the single-view setting in 3 out of 4 datasets, on most metrics.
Among the multi-view methods, DCMTF outperforms both MvSCN and SpectralNet on most metrics, except in PubMed data.
The use of auxiliary data, in the augmented multi-view setting, through DCMTF, improves the performance over multi-view and single view settings, in 3 out of 4 datasets, on most metrics.
Among the methods for the augmented setting, DCMTF outperforms other methods, CFRM and DFMF, in all 4 datasets.

\subsection{Model Analysis}

\subsubsection{Ablation Studies}
To empirically evaluate the importance of the subnetworks and training strategy, we evaluated the performance of DCMTF by making the following changes (separately):
\begin{itemize}[noitemsep,topsep=0pt,labelindent=0em,leftmargin=*]
\item {\bf AE $\rightarrow$ FFN.} The autoencoder network is replaced by a feedforward network. 
\item {\bf $\mathcal{C}\rightarrow$ KM.} The clustering network is removed and K-Means is performed on the representations learnt.
\item {\bf 2$\rightarrow$1-phase.} Training is done in a single-phase instead of 2 phases where in phase 1 we perform matrix reconstruction and in phase 2 we use reconstructed matrices $X^\prime$ to compute the similarity metric to perform clustering.
\end{itemize}

Results  on the Wiki dataset are in Table \ref{tab:abl:wiki}. Performance deteriorates with these changes indicating the importance of both these subnetworks. Note that the fusion network cannot be removed as it is required to get a single representation of entities in multiple input matrices. Similar results are seen in other datasets. 
Single-phase training leads to maximum decrease in performance in all the datasets, due to the dependency for similarity computation discussed earlier.

\begin{table}[h]
\centering
\footnotesize
\begin{tabular}{ccccc}
\hline
& DCMTF & AE $\rightarrow$ FFN & $\mathcal{C}\rightarrow$ KM & 2$\rightarrow$1-phase\\
\hline
ARI           & \textbf{0.4189}       & 0.3024 & 0.1274 & 0.0068  \\
AMI           & \textbf{0.3789}       & 0.3441 & 0.1854 & 0.0158  \\
NMI           & \textbf{0.3796}       & 0.3448 & 0.1863  & 0.0169  \\
RI            & \textbf{0.7354 }      & 0.6721 & 0.5923  & 0.5111  \\
\hline
\end{tabular}
\caption{Ablation Studies: Performance of DCMTF and modified variants on Wiki dataset.}
\label{tab:abl:wiki}
\end{table}

\subsubsection{Sensitivity Analysis}

We conduct a sensitivity analysis with respect to the representation dimension $l_e$, for values $\{20, 50, 100, 200, 300\}$.
Table \ref{tab:sensitivity:reprn_dim} shows the results on Wiki dataset.
We see that the performance is sensitive to $l_e$. For the same $l_e$, runs with different random initializations do not change the results much. 




\begin{table}[h]
\centering
\footnotesize
\begin{tabular}{cccccc}
 \hline
$l_e$ & 20 & 50 & 100 & 200 & 300 \\
 \hline
ARI & 0.2749 & 0.2435 & \textbf{0.4189} & 0.2288 & 0.2651\\
AMI & 0.2955 & 0.3023 & \textbf{0.3789} & 0.2406 & 0.2676\\
NMI & 0.2963 & 0.3030 & \textbf{0.3796} & 0.2414 & 0.2684\\
RI & 0.6668 & 0.6456 & \textbf{0.7354} & 0.6437 & 0.6552\\
\hline
\end{tabular}
\caption{
Sensitivity analysis: DCMTF performance with varying entity representation dimension $l_e$ on Wiki Dataset}
\label{tab:sensitivity:reprn_dim}
\end{table}




\subsection{Additional Results}
We report the Silhouette scores for all other entities, where ground truth labels are unavailable, in all four datasets. Overall, they appear to be well clustered by DCMTF.
We investigate clustering performance when the input number of clusters deviates from the true number.
Clustering performance on another dataset 20-Newsgroups, that was originally used to evaluate CFRM, is also discussed.
When CFRM is initialized with K-Means, its performance improves but remains inferior to that of DCMTF.
These results are in Appendix \ref{app:expt_results}.
Time and model complexity are discussed in Appendix \ref{app:complexity}, where we also list the running time on all four datasets.
DCMTF is considerably slower than both CFRM and DFMF.
\section{Discussion}

The general design of our network architecture follows that of DCMF. 
However, the representation learning in DCMTF is improved in two ways.
First, DCMTF obtains entity representations even when the data for the same entity (e.g., gene) has different datatypes in different matrices.
To obtain consistent real-valued latent representations for all entities, DCMTF uses VAEs and adds an additional data fusion network layer. 
This also aids in similarity computations that are required for clustering.
Second the entity representations are affected by simultaneous learning of clusters.

If DCMF representations are learnt independently, any clustering algorithm can be used on the representations, in a two-step process.
However, this would require a separate method to learn the cluster associations.
Further, simultaneous learning of both representations and clusters has been found to improve both tasks in many other settings, e.g., \citep{hsu2017cnn,zhuge2019simultaneous}.
Following the same principle, DCMTF simultaneously learns entity representations, entity-specific cluster structure as well as the cluster associations from an arbitrary collection of matrices.

An alternative approach to the above DCMTF architecture and optimization is to {\it replace} the factorization step of  DCMF with matrix tri-factorization and to obtain interpretable cluster indicator matrices in the factors various architectural constraints can be imposed, such as using a sigmoid activation layer to ensure that the row sum equals 1 (equation \ref{eq:cfrmmodel}).
However, we found that such constraints deteriorated the quality of either the clustering or the representations; or they led to difficulties in training the network.
Further, learning the association matrix in this setting is challenging.
Similar problems arise if a tri-factorization loss term is {\it added} as a third term to the DCMF loss.

\bibliographystyle{unsrtnat}
\bibliography{main}

\appendix

\section{Notation}

\begin{tabular}{|p{1cm}||p{12cm}|  }
 \hline
 \multicolumn{2}{|c|}{Notations List} \\
 \hline
 Symbol     &  Meaning\\
 \hline
 \hline
    $a$ & 
     \\
    \hline
    $b$ &  Dimension of the entity representation \\
    \hline 
    $c$ &   \\
    \hline
    $c_m$ & index of column entity of the matrix $X^{(m)}$ ($c_m \in  \{1,\ldots,N\}$)\\
    \hline    
    $d_e$ & number of instances of entity $E^{[e]}$  \\
    \hline
    $d_{r_m}$ & number of instances of entity $E^{[r_m]}$, i.e., the row entity of $X^{(m)}$ \\ \hline
    $e$ &  index of entities in input collection ($e \in  \{1,\ldots,N$\} \\
    \hline
    $[e]$ & superscript to indicate entity-specific object\\
    \hline
    $f$ &  parametric function (realized through a neural network)\\
    \hline
    $g$ &  parametric function (realized through a neural network)\\
    \hline
    $h$ &  \\
    \hline
    $i$ &  index \\
    \hline
    $j$ &  index \\
    \hline
    $k$ & number of clusters (in a single input matrix)\\
    \hline
    $k_e$ & number of clusters of instances of entity $E^{[e]}$ \\
    \hline
    $\ell$ & index \\
    \hline
    $m$ &  index of matrices in input collection ($m \in \{1,\ldots,M\}$) \\
    \hline
    $(m)$ & superscript to indicate matrix-specific object\\
    \hline
    $n$ &  \\
    \hline
    $o$ &  \\
    \hline
    $p$ &  number of dimensions in single input matrix \\
    \hline
    $q$ &  number of observations in single input matrix\\
    \hline
    $r$ &   \\
    \hline
    $r_m$ & row entity of the matrix $X^{(m)}$ ($r_m \in  \{1,\ldots,N\}$) \\
    \hline
    $s$ &  \\
    \hline
    $t$ &  BO iteration index\\
    \hline
    $u$ &  cluster index\\
    \hline
    $u_{r_m}$ &  $u^{\rm th}$ cluster in the row entity instances of $X^{(m)}$\\
    \hline
    $v$ &  cluster index \\
    \hline
    $w$ &  \\
    \hline
    $x$ &  \\
    \hline
    $y$ &  \\
    \hline
    $z$ &  \\
 \hline
\end{tabular}

\begin{tabular}{ |p{1cm}||p{12cm}|  }
 \hline
 \multicolumn{2}{|c|}{} \\
 \hline
 Symbol     &  Meaning\\
 \hline
 \hline
     $A_{(m)}$ &  Association matrix containing the strength of association between clusters of row and column entities $X^{(m)}$ \\
    \hline
    \hline
     B &  Graph Partition (set of nodes)\\
    \hline
    $\bar{B}$ &  Complement of $B$ (set of nodes)\\
    \hline
     $C$ &  Relaxation to Vigorous Cluster Indicator (Spectral representation) \\
     \hline
     $\widetilde{C}$ &  Input to the last layer in network multiplied by Cholesky factor to obtain Spectral representation \\
     \hline
     $CC$ &  Cluster chain\\
    \hline
     $D$ &  Degree matrix \\
    \hline
     $E$ &  Set of entities \\
    \hline
     $F$ &  Parametric function (realized via Neural network) \\
    \hline
     $G$ & Input bipartite entity-matrix relationship graph \\
    \hline
     H &  \\
    \hline
    $\widetilde{H}^{[e]}$ & Cholskey factor for entity $E^{[e]}$ \\
    \hline
     $I^{[e]}$ &  Cluster Indicator of entity $E^{[e]}$ \\
    \hline

     $J^{[e]}$ &  Vigorous Cluster Indicator for the $e^{\rm th}$ entity instances \\
    \hline
     K &  \\
    \hline
     $L$ &  Laplacian matrix \\
    \hline
     $M$ &  number of input data matrices\\
    \hline
     N &  number of entities in input collection of matrices\\
    \hline
     O &  \\
    \hline
     $P$ &   Matrix used to define Similarity Metric 
     \\
    \hline
     Q & Set of edges in the input bipartite entity-matrix relationship graph $G$ \\
    \hline
     R &  \\
     \hline
     $S$ & Similarity matrix\\
    \hline
     $T$ &  Matrix transpose operation \\
    \hline
     U &  Entity Representation \\
    \hline
     $V_E$ & Vertices of the bipartite entity-matrix relationship graph $G$ representing Entities \\
    \hline
    $V_M$ & Vertices of the bipartite entity-matrix relationship graph $G$ representing Matrices \\
    \hline
     W &  $W(B_u, B_v)$ Sum of edge weights across partitions $B_u,B_v$\\
    \hline
     $X$ & Data matrix \\
    \hline
     $X^{(m)}$ & $m$-th matrix in a set of Data matrices \\
    \hline
    $X^{(m)^{\prime}}$ & Autoencoder reconstruction of the $m$-th matrix \\
    \hline
    $X^{(m)^{\prime\prime}}$ & Matrix reconstruction of the $m$-th matrix by multiplying latent row and column factors\\
    \hline
    $Y^{(m)}_{[e]}$ & $e^{th}$ entity instances in the matrix $X^{(m)}$ \\ \hline
     Y & matrix (used in spectralnet explanation)\\
    \hline
     Z & lower triangular matrix (used in spectralnet explanation)\\
 \hline
\end{tabular}

\begin{tabular}{ |p{2cm}||p{12cm}|  }
 \hline
 \multicolumn{2}{|c|}{} \\
 \hline
 Symbol     &  Meaning\\
 \hline
 \hline
     $\mathcal{A}$ &  Variational Autoencoder \\
    \hline
    \hline
     $\mathcal{B}$ & \\
    \hline
    $\mathcal{B}^{(m)}_{\{u_{r_m}, u_{c_m} \}}$ &
 sub-matrix block formed by row entity instances of the $u_{r_m}^{\rm th}$ cluster and the column entity instances of the $u_{c_m}^{\rm th}$ cluster in $X^{(m)}$. \\ \hline
     $\mathcal{C}$ & Clustering network \\
    \hline
     $\mathcal{D}$ & \\
    \hline
     $\mathcal{E}$ & \\
    \hline
     $\mathcal{F}$ & Fusion network\\
    \hline
     $\mathcal{G}$ & \\
    \hline
     $\mathcal{H}$ & \\
    \hline
     $\mathcal{I}$ & Entity cluster indicators $\mathcal{I}=I^{(1)},...,I^{(E)}$ \\
    \hline
     $\mathcal{J}$ & Vigorous Cluster Indicator matrix (for single matrix)\\
    \hline
     $\mathcal{K}$ & \\
    \hline
     $\mathcal{L}$ & Loss \\
    \hline
    $\mathcal{L}_\mathcal{A}$ & Autoencoder Reconstruction Loss \\
    \hline
    $\mathcal{L}_\mathcal{C}$ & Spectral Clustering Loss \\
    \hline
    $\mathcal{L}_\mathcal{R}$ & Matrix Reconstruction Loss \\
    \hline
     $\mathcal{M}$ & Weighted \textit{distance} matrix \\
    \hline
     $\mathcal{N}[e]$ & Neighbors of entity $E^{[e]}$ in $G$\\
    \hline
    $\mathcal{N}\mathcal{N}$ & Neural Network \\ \hline
     $\mathcal{O}$ & \\
    \hline
     $\mathcal{P}$ & \\
    \hline
     $\mathcal{Q}$ & \\
    \hline
     $\mathcal{R}$ & \\
    \hline
     $\mathcal{S}$ & Similarity Matrix\\
    \hline
     $\mathcal{T}$ & Cluster associations $\mathcal{T} = {A^{{(1)}^{\prime}},...,A^{{(M)}^{\prime}}}$\\
    \hline
     $\mathcal{U}$ & Entity representations $\mathcal{U}=U^{(1)},...,U^{(E)}$,\\
    \hline
     $\mathcal{V}$ & \\
    \hline
     $\mathcal{W}$ & \\
    \hline
     $\mathcal{X}$ & Input matrices $\mathcal{X} = {X^{(1)},...,X^{(M)}}$ \\
    \hline
     $\mathcal{Y}$ & \\
    \hline
     $\mathcal{Z}$ & \\
 \hline
\end{tabular}

 \begin{tabular}{ |p{1cm}||p{12cm}|  }
 \hline
 \multicolumn{2}{|c|}{} \\
 \hline
 Symbol     &  Meaning\\    
 \hline
 \hline
 $\lambda$ & Eigenvalue \\
 \hline
 $\theta, \epsilon, \delta, \eta$ & Neural network parameters \\
 \hline
 $\pi^{[e]}_u$ & the $u$-th cluster of entity instances $E^{[e]}$ \\
 \hline
 $\Gamma[.]$ & Concatenation operator \\
 \hline
  $\mu_{\epsilon}^{e,m}$ & Mean of the VAE encoder for the $e^{\rm th}$ entity in the $m^{\rm th}$ matrix  \\
 \hline
 $\mathbf{I}_{k_e}$ & Identity Matrix with dimensions ($k_e \times k_e$) \\ \hline
     $\mathbf{I}$ & Identity Matrix \\
    \hline
    $\mathbb{R}$ & Real numbers \\ \hline
\end{tabular}
 
     

\section{Details of Datasets Used}
\label{app:data_details}

\subsection*{Wiki}

Wikidata is a secondary database that collects structured data to support Wikipedia.
It extracts and transforms different parts of Wikipedia articles to create a relational knowledge base, that can be programmatically accessed, to support a wide range of applications. 
Wikidata is part of the non-profit, free-content \href{https://wikimediafoundation.org/}{\textit{Wikimedia foundation}}. The content of Wikidata is available under \href{https://creativecommons.org/publicdomain/zero/1.0/}{\textit{a free license}}.

\begin{figure}[H] 
\centering
    \subfloat[]{
    \centering
            \def\svgwidth{0.35\textwidth} 
            \fontsize{8pt}{8pt}\selectfont
            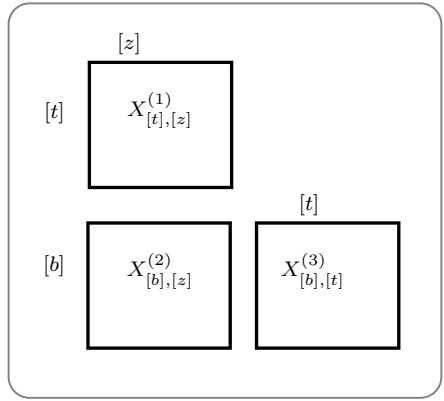
    }
    \caption{
    Case study 2:
    3 entities $[t]$: \textit{Items}, $[z]$: \textit{Subject}, $[b]$: \textit{Bag-of-Words} vocabulary and 3 relations between the entities, matrices $X_{[t],[z]}^{(1)}$, $X_{[b],[z]}^{(2)}$, $X_{[b],[t]}^{(3)}$,
   }
\label{fig:wiki:setting}
\end{figure}

    We created a dataset with 3 entities namely \textit{Items} $[t]$, \textit{Subjects} $[z]$, \textit{Bag-of-Words} $[b]$ and 3 matrices $X_{[t],[z]}^{(1)}$, $X_{[b],[z]}^{(2)}$, $X_{[b],[t]}^{(3)}$	as shown in the Figure \ref{fig:wiki:setting} and matrix dimensions as listed in the table \ref{tab:dataset:wiki}. The sparsity level of the matrices can be seen in the table \ref{tab:dataset:wiki}. The entity $t$ refers to \textit{Items} belonging to one of the 3 categories: movies, games and books, the entity $z$ refers to the \textit{Subject(s)} that an \textit{Item}'s instance can belong to and the entity $b$ refers to the vocabulary of the \textit{Bag-of-Words} model generated using the text in abstract section of the Wikipedia articles corresponding to \textit{Item} and \textit{Subject} instances. 

From Wikidata, we selected 1800 \textit{Item} instances belonging to the 3 categories: movies, games and books. We also selected 1116 \textit{Subjects} that the \textit{Items} belong to and created the matrix $X_{[t],[z]}^{(1)}$, where $X_{[t],[z]}^{(1)}(i,j)$ is 1 if $i$-th \textit{Item} belongs to the $j$-th \textit{Subject}. We then used the Wikidata webservices to obtain the abstract section of the Wikipedia articles corresponding to the \textit{Item} and \textit{Subject}  instances. We then obtained $b$, the vocabulary of the \textit{Bag-of-Words} model built with all the text of abstracts of $t$ and $z$ instances - using which we created the matrices $X_{[b],[z]}^{(2)}$ and 
$X_{[b],[t]}^{(3)}$. 
We evaluate the clustering performance over the entity \textit{Items} with 1800 instances whose ground truth labels are one of the 3 categories.
 
 \begin{table}[H]
    \centering
    \small
    \begin{tabular}{|c|c|c|c|c|c|}
         \hline
         Matrix & Row Entity & Col Entity & Row Dim & Col Dim  & Sparsity\%\\ \hline
         $X_{[t],[z]}^{(1)}$ & Items & Subject & 1800 & 1116  & 99.86 \\ \hline
         $X_{[b],[z]}^{(2)}$ & Bag-of-Words & Subject & 3883 & 1116 & 97.75 \\ \hline
         $X_{[b],[t]}^{(3)}$ & Bag-of-Words  & Items & 3883 & 1800 & 98.86 \\ \hline
    \end{tabular}
    \caption{\small Wiki: Dataset Statistics (Col: Column, Dim: Dimension)}
    \label{tab:dataset:wiki}
\end{table} 
 
 \begin{table}[H]
    \centering
    \small
    \begin{tabular}{|c|c|}
         \hline
         $[e]$ & Entity name\\ \hline \hline
         $[t]$ & Items \\ \hline
         $[z]$ & Subjects \\ \hline
         $[b]$ & Bag-Of-Words \\ \hline
    \end{tabular}
    \caption{\small Wiki Dataset: Entity names}
    \label{tab:dataset:wiki:entities}
\end{table} 
 
\newpage
\subsection*{Cancer}
\begin{figure}[H] 
\centering
    \subfloat[]{
    \centering
            \def\svgwidth{0.35\textwidth} 
            \fontsize{6pt}{6pt}\selectfont
            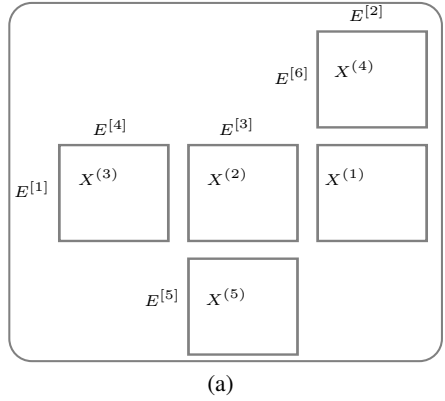
    }
        \caption{Cancer dataset setup: 6 entities and 5 matrices.}
        \label{fig:aug_multiview}
\end{figure}

We construct a dataset that forms the setup shown in fig. \ref{fig:aug_multiview}. Table \ref{tab:dataset} shows sparsity level, entity type for each matrix and the matrix dimensions. We use four publicly available biomedical data sources to construct the dataset: 

\begin{enumerate}
\item 
\textit{DisGeNET} (\cite{pinero2016disgenet}) is a database of known gene-disease associations, collected from expert curated repositories, GWAS catalogues, animal models and scientific literature. The DisGeNET database is made available under the \href{https://creativecommons.org/licenses/by-nc-sa/4.0/}{\textit{Attribution-NonCommercial-ShareAlike 4.0 International License}}.
\item 
\textit{The Cancer Genome Atlas (TCGA)} (\cite{weinstein2013cancer}) 
contains genomic and clinical data of 33 different cancers and over 10,000 patients. The TCGA is currently available at the Genomic Data Commons (GDC), which requires the user to adhere to \href{https://osp.od.nih.gov/scientific-sharing/policies/}{\textit{NIH Genomic Data Sharing Policy}}
\item 
\textit{Humannet} (\cite{lee2011prioritizing}) is a functional gene network of human genes obtained by integration of 21 types of `omics' data sources. Each edge in HumanNet is associated with the probability of a true functional linkage between two genes. Humannet is licensed under a \href{http://creativecommons.org/licenses/by-sa/4.0/}{\textit{Creative Commons Attribution-ShareAlike 4.0 International License}}.
\item 
\textit{UMLS Metathesaurus} (\cite{schuyler1993umls}) is a 
large database of biomedical concepts and their relationships. Users need \href{https://uts.nlm.nih.gov/uts/assets/LicenseAgreement.pdf}{\textit{License Agreement for Use of the UMLS Metathesaurus}}
\end{enumerate}

We only consider the expert curated gene-disease associations from DisGeNET for our dataset construction, since these are the most reliable.
We also restrict our data to a single cancer (Breast Cancer) in TCGA. With these restrictions, there were 11939 genes that were present in all three databases: DisGeNET, TCGA and HumanNet, with 1093 and 11809 associated patients and diseases respectively. 
We chose a random subset of 2000 genes and associated diseases (968) and all the patients (1093).
For these genes and diseases, the gene-disease association matrix $X^{(1)}$ was constructed by using all known associations from DisGeNET: there were 69850 associations, resulting in sparsity level of 96.5\%.
To construct $X^{(2)}$ we used RNA-Seq Expresssion data from TCGA, where a single sample per patient was chosen.
TCGA also contains 115 demographic and clinical features for these patients. We chose a subset of 8 numeric and 21 categorical features as listed in Table \ref{table:patient_features} 
We then transformed the categorical features to their one-hot encodings, normalized the numeric features and obtained a total of 94 patient features.
Gene-gene and disease-disease graphs for the selected genes and diseases were obtained from HumanNet and UMLS respectively.
Similar to preprocessing done by \cite{natarajan2014inductive}, we use principal components of the adjacency matrices of these graphs as features to obtain matrices $X^{(3)}, X^{(4)}$.

We evaluate the clustering performance over the entity \textit{Patient} with 1093 instances whose ground truth labels are one of the 4 cancer stages.

\begin{table}[H]
    \centering
    \small
    \begin{tabular}{|c|c|c|c|c|c|}
         \hline
         Matrix & Row Entity & Col Entity & Row Dim & Col Dim & Sparsity\% \\ \hline
         $X^{(1)}$ & Gene & Disease & 2000 & 968 & 96.49\\ \hline
         $X^{(2)}$ & Gene & Patient & 2000 & 1093 & 69.92\\ \hline
         $X^{(3)}$ & Gene & Gene Features & 2000 & 1000 & 53.18\\ \hline
         $X^{(4)}$ & Disease Features & Disease & 500 & 968 & 49.39\\ \hline
         $X^{(5)}$ & Patient Features & Patient & 94 & 1093 & 73.61\\ \hline
    \end{tabular}
    \caption{\small Gene-Disease Association: Dataset Statistics (Col: Column, Dim: Dimension) }
    \label{tab:dataset}
\end{table} 

\begin{table}[H]
    \centering
    \small
    \begin{tabular}{|c|c|}
         \hline
         $e$ & Entity name \\ \hline \hline
         $1$ & Gene \\ \hline
         $2$ & Disease \\ \hline
         $3$ & Patient \\ \hline
         $4$ & Gene features \\ \hline
         $5$ & Patient features \\ \hline
    \end{tabular}
    \caption{\small Cancer Dataset: Entity names}
    \label{tab:dataset:cancer:entities}
\end{table} 

\begin{table}[H]
\resizebox{\textwidth}{!}{%
\begin{tabular}{@{}ll@{}}
\toprule
CATEGORICAL                                                               & NUMERIC                                                                     \\ \midrule
\hline
American Joint Committee on Cancer Tumor Stage Code                       & Diagnosis Age                                                               \\
Neoplasm Disease Lymph Node Stage American Joint Committee on Cancer Code & Death from Initial Pathologic Diagnosis Date                                \\
American Joint Committee on Cancer Metastasis Stage Code                  & Positive Finding Lymph Node Hematoxylin and Eosin Staining Microscopy Count \\
Neoplasm Disease Stage American Joint Committee on Cancer Code            & Disease Free (Months)                                                       \\
New Neoplasm Event Post Initial Therapy Indicator                         & Lymph Node(s) Examined Number                                               \\
Metastatic tumor indicator                                                & Last Alive Less Initial Pathologic Diagnosis Date Calculated Day Value      \\
Overall Survival Status                                                   & HER2 ihc score                                                              \\
Disease Free Status                                                       & Overall Survival (Months)                                                   \\
Patient's Vital Status                                                    &                                                                             \\
ER Status By IHC                                                          &                                                                             \\
Prior Cancer Diagnosis Occurence                                          &                                                                             \\
Micromet detection by ihc                                                 &                                                                             \\
PR status by ihc                                                          &                                                                             \\
Person Neoplasm Status                                                    &                                                                             \\
Ethnicity Category                                                        &                                                                             \\
Tissue Retrospective Collection Indicator                                 &                                                                             \\
Disease Surgical Margin Status                                            &                                                                             \\
Sex                                                                       &                                                                             \\
Primary Lymph Node Presentation Assessment Ind-3                          &                                                                             \\
Neoadjuvant Therapy Type Administered Prior To Resection Text             &                                                                             \\
Tissue Prospective Collection Indicator                                   &                                                                             \\ \bottomrule
\end{tabular}%
}
\caption{List of patient features.}
\label{table:patient_features}
\end{table}

\newpage
\subsection*{Freebase}
Freebase was a collaborative knowledge base consisting of data curated by its community members from various data sources including Wikipedia, Notable Names Database (NNDB), etc. Freebase data was available for use under \href{https://creativecommons.org/}{\textit{Creative Commons Attribution License}} through open API. The Freebase data service was shut down and the data was moved to Wikidata. 

\begin{figure}[H] 
\centering
    \subfloat[]{
    \centering
            \def\svgwidth{0.4\textwidth} 
            \fontsize{6pt}{6pt}\selectfont
            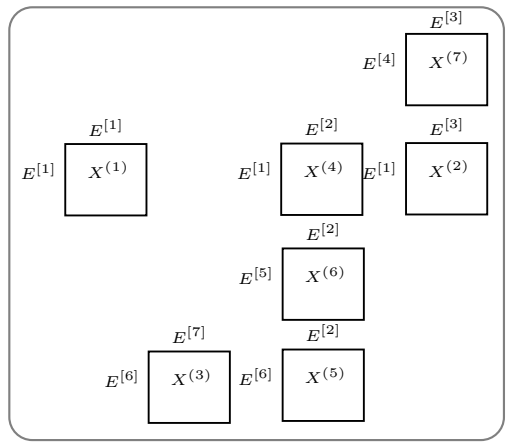
    }
    \caption{
    Freebase dataset setup: 7 entities and 7 matrices
    }
\label{fig:freebase:setting}
\end{figure}

We used a subset of the \textit{Freebase} dataset provided as one of the benchmark datasets in \citep{yang2020heterogeneous} for evaluating Heterogeneous Network Representation Learning techniques. The original dataset consists of 36 matrices and 8 entities. We selected a set of 7 entities and 7 matrices with the dimensions as listed in the table \ref{tab:dataset:freebase} and the matrices setup as shown in figure \ref{fig:freebase:setting}. The matrices, entities and entity instances were selected such that we get minimal sparsity in the matrices, the values of which are show in the table \ref{tab:dataset:freebase}. We evaluate the clustering performance over the entity \textit{Books} with 985 instances whose ground truth labels are one of the 8 categories given in table \ref{tab:dataset:freebase:labels}.  

\begin{table}[H]
    \centering
    \small
    \begin{tabular}{|c|c|c|c|c|c|}
         \hline
         Matrix & Row Entity & Col Entity & Row Dim & Col Dim & Sparsity\%\\ \hline \hline
         $X^{(1)}$ & Book & Book & 985 & 985 & 99.87\\ \hline
         $X^{(2)}$ & Book & Music & 1000 & 985 & 99.84\\ \hline
         $X^{(3)}$ & Location & Organization & 1000 & 1000 & 98.58\\ \hline
         $X^{(4)}$ & Book & Film & 1000 & 985 & 99.64\\ \hline
         $X^{(5)}$ & Location & Film & 1000 & 1000 & 99.64\\ \hline
         $X^{(6)}$ & People & Film & 1000 & 422 & 99.57 \\ \hline
         $X^{(7)}$ & Sports & Music & 1000 & 1000 & 98.79\\ \hline
    \end{tabular}
    \caption{\small Freebase: Dataset Statistics (Col: Column, Dim: Dimension)}
    \label{tab:dataset:freebase}
\end{table} 

\begin{minipage}{0.5\textwidth}
\begin{table}[H]
    \centering
    \small
    \begin{tabular}{|c|c|}
         \hline
         $e$ & Entity name \\ \hline \hline
         1 &  Book \\ \hline
         2 &  Film \\ \hline
         3 &  Music \\ \hline
         4 &  Sports \\ \hline
         5 &  People \\ \hline
         6 &  Location \\ \hline
         7 &  Organization \\ \hline
    \end{tabular}
    \caption{\small Freebase Dataset: Entity names}
    \label{tab:dataset:freebase:entities}
\end{table} 
\end{minipage}
\begin{minipage}{0.5\textwidth}
\begin{table}[H]
    \centering
    \small
    \begin{tabular}{|c|c|}
         \hline
         Class & $E^{[1]}$ Book's Category\\ \hline \hline
         0 & Scholarly work\\ \hline
         1 & Book character/Book subject\\ \hline
         2 & Publication/Published work\\ \hline
         3 & Short story\\ \hline
         4 & Magazine/Magazine issue/Magazine genre \\ \hline
         5 & Newspaper\\ \hline
         6 & Journal article/Journal/Journal publication\\ \hline
         7 & Poem/Poem character\\ \hline
    \end{tabular}
    \caption{\small Freebase Dataset: Categories of entity \textit{Books} }
    \label{tab:dataset:freebase:labels}
\end{table} 
\end{minipage}

\newpage
\subsection*{PubMed}
\href{https://pubmed.ncbi.nlm.nih.gov/}{\textit{PubMed}} is a free search engine for accessing biomedical and life sciences literature. The PubMed database contains more than 32 million citations and abstracts of journal articles. PubMed was developed and is maintained by the National Center for Biotechnology Information (NCBI). Once a year, NLM releases a complete (baseline) set of PubMed citation records, with daily Incremental update files including new, revised, and deleted citations. Downloading them from their FTP serves indicate acceptance of the \href{https://ftp.ncbi.nlm.nih.gov/pubmed/baseline/README.txt}{\textit{PubMed Specific Terms}}. NLM freely provides PubMed data.

\begin{figure}[H] 
\centering
    \subfloat[]{
    \centering
            \def\svgwidth{0.3\textwidth} 
            \fontsize{6pt}{6pt}\selectfont
            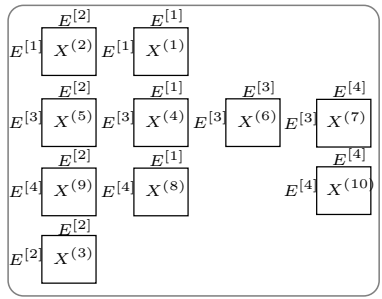
    }
    \caption{
    PubMed dataset setup: 4 entities and 10 matrices
    }
\label{fig:pubmed:setting}
\end{figure}

We used a subset of the \textit{PubMed} dataset provided as one of the benchmark datasets in \citep{yang2020heterogeneous} for evaluating Heterogeneous Network Representation Learning techniques. The original dataset consists of 10 matrices and 4 entities. We used the same set of 10 matrices and 4 entities but with a subset of entity instances as listed in the table \ref{tab:dataset:pubmed} and the matrices setup as shown in figure \ref{fig:pubmed:setting}. The entities instances were selected such that we get minimal sparsity in the matrices, the values of which are show in the table \ref{tab:dataset:pubmed}. We evaluate the clustering performance over the entity \textit{Diseases} with 440 instances whose ground truth labels are known and are one of the 8 categories given in table \ref{tab:dataset:pubmed:labels}.    

\begin{table}[H]
    \centering
    \small
    \begin{tabular}{|c|c|c|c|c|c|}
         \hline
         Matrix & Row Entity & Col Entity & Row Dim & Col Dim & Sparsity\% \\ \hline \hline
         $X^{(1/11)}$ & Gene & Gene & 563 & 563 & 98.56 \\ \hline
         $X^{(2/12)}$ & Gene & Disease & 563 & 440 & 99.90  \\ \hline
         $X^{(3/22)}$ & Disease & Disease & 440 & 440 & 99.99  \\ \hline
         $X^{(4/31)}$ & Chemical & Gene & 708 & 563 & 98.07  \\ \hline
         $X^{(5/32)}$ & Chemical & Disease & 708 & 440 & 99.85  \\ \hline
         $X^{(6/33)}$ & Chemical & Chemical & 708 & 708 & 97.35 \\ \hline
         $X^{(7/34)}$ & Chemical & Species & 708 & 657 & 99.51  \\ \hline         
         $X^{(8/41)}$ & Species & Gene & 657 & 563 & 99.64  \\ \hline
         $X^{(9/42)}$ & Species & Disease & 657 & 440 & 99.98  \\ \hline
         $X^{(10/44)}$ & Species & Species & 657 & 657 & 99.87  \\ \hline
    \end{tabular}
    \caption{\small PubMed: Dataset Statistics (Col: Column, Dim: Dimension)}
    \label{tab:dataset:pubmed}
\end{table} 

\begin{minipage}{0.5\textwidth}
\begin{table}[H]
    \centering
    \footnotesize
    \begin{tabular}{|c|c|}
         \hline
         $e$ & Entity name\\ \hline \hline
         1 & Gene \\ \hline
         2 & Disease \\ \hline
         3 & Chemical \\ \hline
         4 & Species \\ \hline
    \end{tabular}
    \caption{\small PubMed Dataset: Entity names}
    \label{tab:dataset:pubmed:entities}
\end{table} 
\end{minipage}
\begin{minipage}{0.5\textwidth}
\begin{table}[H]
    \centering
    \footnotesize
    \begin{tabular}{|c|c|}
         \hline
         Class & $E^{[2]}$ Disease Category\\ \hline \hline
         0 & Cardiovascular disease\\ \hline
         1 & Glandular disease\\ \hline
         2 & Nervous disorder\\ \hline
         3 & Communicable disease\\ \hline
         4 & Inflammatory disease \\ \hline
         5 & Pycnosis\\ \hline
         6 & Skin disease\\ \hline
         7 & Cancer\\ \hline
    \end{tabular}
    \caption{\small PubMed Dataset: Categories of entity \textit{Diseases} }
    \label{tab:dataset:pubmed:labels}
\end{table} 

\end{minipage}

\section{Computational Complexity}
\label{app:complexity}

\subsection{Time Complexity}
\label{app:time_complexity}

\begin{itemize}
    \item The training time complexity of DCMTF in each iteration is dominated by 
    \begin{itemize}
        \item the Autoencoder with largest input $Y^{(m)}_{[e^*]}$ of dimension, say, $d_{[e^*]} \times d_{[c_m]}$ (assuming that $[e^*]$ is the row entity of $X^{[m]}$) and is $\mathcal{O}(d_{[e^*]} \cdot d_{[c_m]} \cdot \lambda_1^{\mathcal{A}})$ where $\lambda_1^{\mathcal{A}}$ is the number of neurons in the first layer of the Autoencoder $\mathcal{A}^{(e^*,m)}$. 

        \item the Fusion network corresponding to the largest input
        $\Gamma_{m}[ \mu_{\epsilon}^{(e^*,m)} ]  ),$ where $\Gamma_{m}[.]$ represents the concatenation operation and the index $m$ iterates over all $M_{[e^*]}$ matrices containing the $e^*$-th entity, assuming that the $e^*$-th entity is associated with the maximum number of matrices. The complexity then is $\mathcal{O}(d_{[e^*]} \cdot (M_{[e^*]} \cdot l) \cdot \lambda_1^{\mathcal{F}})$ where $\lambda_1^{\mathcal{F}}$ is the number of neurons in the first layer of the fusion network $\mathcal{F}^{[e^*]}$.
        
        \item the matrix inversion operation performed in the clustering network corresponding to the entity with the largest number of instances and is $\mathcal{O}(d_{[e^*]}^3)$
        
        \item the matrix completion step, corresponding to the matrix with the largest row and column dimensions, say $d_{[e^*]} \times d_{[c_m]}$. The matrix will be reconstructed by multiplying the row and column latent factors $U^{[e^*]}$ and $U^{[c_m]}$ of dimension $d_{[e^*]} \times l$ and $d_{[c_m]} \times l$ respectively. The time complexity is $\mathcal{O}(d_{[e^*]} \cdot l \cdot d_{[c_m]} )$, where $l$ is the latent representation dimension.
    \end{itemize}
    
    \item The training time complexity of phase 1 is $t$ $\cdot$ $\mathcal{O}(d_{[e^*]} \cdot d_{[c_m]} \cdot \lambda_1^{\mathcal{A}})$ + $\mathcal{O}(d_{[e^*]} \cdot (M_{[e^*]} \cdot l) \cdot \lambda_1^{\mathcal{F}})$ + $\mathcal{O}(d_{[e^*]} \cdot l \cdot d_{[c_m]} )$ = $\mathcal{O}( t \cdot max(d_{[e^*]}  d_{[c_m]}  \lambda_1^{\mathcal{A}}, \  d_{[e^*]} M_{[e^*]} l , \  d_{[e^*]} l d_{[c_m]}))$
    
    \item Similarly the training time complexity of phase 2 is $\mathcal{O}( t \cdot max(d_{[e^*]}  d_{[c_m]}  \lambda_1^{\mathcal{A}}, \  d_{[e^*]} M_{[e^*]} l , \  d_{[e^*]}^3))$
    
    \item The testing time clustering complexity of DCMTF is same as that of phase 2 training and matrix completion complexity is same as that of phase 1 training.

\end{itemize}

\subsection{Running time}
\label{app:running_time}

The running time (in minutes) of DCMTF and the baselines, for clustering the datasets listed in Table \ref{tab:clust_perf_all}, are given in Table \ref{tab:running_time}. These are recorded in the same machine that is described above.

\begin{table}[h]
\centering
\begin{tabular}{ccccc}
\hline
 & Wiki & Cancer & Freebase & PubMed \\
\hline
DCMTF    & 13.511 & 27.370 & 35.975   & 14.204 \\
CFRM     & 0.101  & 0.050  & 0.038    & 0.022  \\
DFMF     & 0.194  & 0.095  & 0.083    & 0.049  \\
\hline
\end{tabular}
\caption{Running time in minutes}
\label{tab:running_time}
\end{table}

    
    
	
    
\subsection{Model Complexity}
\label{app:model_complexity}

\begin{itemize}
    \item The model complexity (number of free parameters),  $\text{p}_\textsubscript{dcmf} = \mathcal{O}(\text{p}_\textsubscript{u} + \text{p}_\textsubscript{a} + \text{p}_\textsubscript{f} + \text{p}_\textsubscript{c})$, where
        \begin{itemize}
            \item the latent entity representation of dimension $l$ (considered as parameters): $\text{p}_\textsubscript{u} = \sum_{[e] \in \mathcal{E}}(d_{[e]}*l)$
            
            \item the weights of the three subnetworks:
            the autoencoder networks: $\text{p}_\textsubscript{a} = \sum_{(e,m) \in Q}\{\text{\textit{par}}(\mathcal{A}^{(e,m)})\}$,
            the fusion networks: $\text{p}_\textsubscript{f} = \sum_{[e] \in \mathcal{E}}\{\text{\textit{par}}(\mathcal{F}^{[e]})\}$, and 
            the clustering networks: 
            $\text{p}_\textsubscript{c} = \sum_{[e] \in \mathcal{E}}\{\text{\textit{par}}(\mathcal{C}^{[e]})\}$, where $par$ denotes the network weights.
        \end{itemize}

\item The number of model parameters is a function of number of matrices $M$, number of entities $N$ and the entity dimensions $d_{[e]}$:

    \begin{itemize}

        \item If the number of matrices $M$ alone increases i.e more edges $(e,m)$ in $G$, then $\text{p}_a$ and $\text{p}_\text{f}$ increases: $\text{p}_a$ increases due to the increase in the number of Autoencoders $\mathcal{A}^{(e,m)}$ corresponding to the new edges and $\text{p}_\text{f}$ increases as the number of neurons in the input layer of the fusion network $\mathcal{F}^{[e]}$ is proportional to the number matrices the entity $[e]$ is associated with. 

        \item If the number of matrices $M$ increases along with the number of entities $N$, then $\text{p}_{\text{dcmtf}}$ increases as all the subnetworks grow in size.

        \item If the entities size $d_{[e]}$ increases, thereby increasing the associated matrix dimensions, then $\text{p}_u$ and $\text{p}_\text{a}$ increases: $\text{p}_u$ increases due to increase in number of latent representations to be learnt and $\text{p}_\text{a}$ increases as the number of neurons in the first layer of $\mathcal{A}^{(e,m)}$ is determined by the size of the entities associated with $[e]$ through $(m)$.

    \end{itemize}

\end{itemize}


\newpage
\section{Additional Experimental Results}
\label{app:expt_results}

\subsection{Clustering Performance}
\label{app:clust_perf}

Given below are the results for SpectralNet, Spectral Clustering and K-Means for:
\begin{itemize}
\item 
Multi-view setting: the input matrices (unshaded in fig. \ref{fig:data_schematic}) are concatenated and given as inputs.
\item
Single-view setting: each of the input matrices in the multi-view setup (unshaded in fig. \ref{fig:data_schematic}) is clustered separately.
\end{itemize}

\begin{table*}[!h]
    \centering
    \captionsetup{font=footnotesize}
    \small
    \begin{tabular}{c|p{1.65em}p{1.65em}p{2.75em}|p{1.65em}p{1.65em}p{2.75em}|p{1.65em}p{1.65em}p{2.75em}}
    \toprule
     \textbf{Dataset:} &  \multicolumn{3}{|c}{\textbf{Spectral Net}} & \multicolumn{3}{|c}{\textbf{Spectral Clustering}} & \multicolumn{3}{|c}{\textbf{KMeans}} \\ 
     \hline
     \textbf{Metric}  & $X_{ts}$ & $X_{tb}$ & Multi-View & $X_{ts}$ & $X_{tb}$ & Multi-View & $X_{ts}$ & $X_{tb}$ & Multi-View \\ \hline
    ARI &    0.0336  &   0.0079  &   0.0029  &   \text{-}.0430  &   0.0  &   0.0  &   \text{-}.0011  &   0.0002  &   0.0002  \\
    AMI &     0.036  &   0.0074  &   0.0023  &   0.0258  &   0.0  &   0.0  &   0.0003 & 0.0002 &  0.0002  \\
    NMI &     0.0384  &   0.0088  &   0.0034 &   0.0272  &   0.0  &   0.0  &   0.0024  &   0.0024  &   0.0024  \\ 
    RI  &    0.5461  &   0.5347  &  0.5375  &   0.4305  &   0.3448  &   0.3448  &   0.3505  &   0.3456  &   0.3456  \\
     \bottomrule
    \end{tabular}
    \caption{Clustering Performance of different benchmark algorithms on Wiki Dataset.}
    \label{tab:results_wiki1}
\end{table*}

\begin{table*}[!h]
    \centering
    \captionsetup{font=footnotesize}
    \small
    \begin{tabular}{c|p{1.65em}p{1.65em}p{2.75em}|p{1.65em}p{1.65em}p{2.75em}|p{1.65em}p{1.65em}p{2.75em}}
    \toprule
     \textbf{Dataset:} &  \multicolumn{3}{|c}{\textbf{Spectral Net}} & \multicolumn{3}{|c}{\textbf{Spectral Clustering}} & \multicolumn{3}{|c}{\textbf{KMeans}} \\ 
     \hline
     \textbf{Metric} &  $P-PF$ & $G-P$ & Multi-View & $P-PF$ & $G-P$ & Multi-View & $P-PF$ & $G-P$ & Multi-View \\ \hline
    ARI &     0.0499  &   0.0152  &   0.0044  &   0.0317  &   0.0011  &   0.0049  &   0.0143  &   0.0  &   0.0  \\
    AMI &     0.0277  &   0.0021  &   0.0019  &   0.0238  &   0.0023  &  0.0007  &   0.0180 & 0.0004 &  0.0002  \\
    NMI &    0.0351  &   0.0073  &   0.0070 &   0.0306  &   0.0070  &  0.0058  &   0.0231  &   0.0052  &   0.0055  \\ 
    RI  &   0.5209  &   0.5423  &  0.5413  &   0.4651  &   0.5419  &   0.5307  &   0.5200  &   0.5380  &   0.5377  \\
     \bottomrule
    \end{tabular}
    \caption{Clustering Performance of different benchmark algorithms on Cancer Dataset.}
    \label{tab:results_genephene}
\end{table*}

\begin{table*}[!h]
    \centering
    \captionsetup{font=footnotesize}
    \small
    \begin{tabular}{c|p{1.65em}p{1.65em}p{1.65em}p{2.75em}|p{1.65em}p{1.65em}p{1.65em}p{2.75em}|p{1.65em}p{1.65em}p{1.65em}p{2.75em}}
    \toprule
     \textbf{Dataset:} &  \multicolumn{4}{|c}{\textbf{Spectral Net}} & \multicolumn{4}{|c}{\textbf{Spectral Clustering}} & \multicolumn{4}{|c}{\textbf{KMeans}} \\ 
     \hline
     \textbf{Metric} &  $X1$ & $X4$ & $X2$ & Multi-View & $X1$ & $X4$ & $X2$ & Multi-View & $X1$ & $X4$ & $X2$ & Multi-View \\ \hline
    ARI &      0.0406  &   0.0251  &   \text{-}.0009  &   0.0395  &   0.0  &   0.0024  &   0.0016  &   0.0386  &   0.0100 & 0.0569 & 0.0023 & 0.0455  \\
    AMI &      0.0357  &   0.0629  &   0.0345  &   0.0535  &  0.0  &   0.0077  &   0.0569  &   0.0585 & 0.0594 & 0.1116 & 0.0788 & 0.1278  \\
    NMI &    0.0510  &   0.0948  &   0.1021 &   0.0696  &   0.0  &   0.0236  &   0.0735 &   0.0693  &   0.0742 & 0.1280 & 0.0980 & 0.1398 \\
    RI  &    0.7175  &   0.5431  &  0.2491  &   0.7233  &   0.1269  &   0.5734  &  0.5562  &   0.6613  &  0.3650 & 0.5269 & 0.2867 & 0.5473  \\
     \bottomrule
    \end{tabular}
    \caption{Clustering Performance of different benchmark algorithms on Freebase Dataset.}
    \label{tab:results_freebase}
\end{table*}

\begin{table*}[!h]
    \centering
    \captionsetup{font=footnotesize}
    \footnotesize
    \small
    \begin{tabular}{c|p{1.65em}p{1.65em}p{1.65em}p{1.65em}p{2.75em}|p{1.65em}p{1.65em}p{1.65em}p{1.65em}p{2.75em}|p{1.65em}p{1.65em}p{1.65em}p{1.65em}p{2.75em}}
    \toprule
     \textbf{Dataset:} &  \multicolumn{5}{|c}{\textbf{Spectral Net}} & \multicolumn{5}{|c}{\textbf{Spectral Clustering}} & \multicolumn{5}{|c}{\textbf{KMeans}} \\ 
     \hline
     \textbf{Metric} &   $X2$ & $X5$ & $X9$ & $X3$ & Multi-View & $X2$ & $X5$ & $X9$ & $X3$ & Multi-View & $X2$ & $X5$ & $X9$ & $X3$ & Multi-View \\ \hline
    ARI &      
    0.0088  &   0.0124  &   -  &   -  &   0.0073 &
    0.0016  &   0.0070  &   \text{-}.0006  &   0.0041  &   0.0058  
    &   0.0086  &   0.0081 & \text{-}.0008 & 0.0041 & 0.0755  \\
    AMI &          
    0.0040  &   \text{-}.0024  &   -  &   -  &   \text{-}.0006 &
    \text{-}.0058  &   \text{-}.0032  &  \text{-}.0013  &   0.0012  &   0.0048  
    &   0.0084 & \text{-}.0052 & \text{-}.0025 & 0.0127 & 0.0053  \\
    NMI &      
     0.0560  &   0.0355  &   -  &   -  &   0.0335 &
    0.0329 &   0.0351  &   0.0367  &   0.0481  &   0.0032 & 
    0.0497  &   0.0318 & 0.0406 & 0.0481 & 0.0435 \\
    RI  &    
     0.3888  &   0.5572  &   -  &   -  &   0.6276 &
    0.3516  &   0.4266  &   0.2782  &   0.2120  &  0.5418  
    &   0.4382  &  0.5571 & 0.2775 & 0.2120 & 0.5265  \\
     \bottomrule
    \end{tabular}
    \caption{Clustering Performance of different benchmark algorithms on Pubmed Dataset.}
    \label{tab:results_pubmed}
\end{table*}

\subsection{Ablation Studies}
\begin{minipage}{0.5\textwidth}
\begin{table}[H]
\footnotesize
\centering
\begin{tabular}{lllll|}
\hline
& DCMTF & AE $\rightarrow$ FFN & $\mathcal{C}\rightarrow$ KM \\
\hline
ARI                    & 0.0212 & 0.0154 & 0.0015 \\
AMI                    & 0.0182 & 0.0166 & 0.0005 \\
NMI                    & 0.0229 & 0.0214 & 0.0053 \\
RI                     & 0.5479 & 0.5458 & 0.5400 \\
\hdashline
Silhouette:            &        &        &        \\
Disease                & 0.3931 & 0.3691 & 0.2119 \\
Disease Features       & 0.3622 & 0.3702 & 0.1170 \\
Gene                   & 0.3038 & 0.3117 & 0.1372 \\
Gene Features          & 0.3535 & 0.8720 & 0.0929 \\
Patient*  & 0.2248 & 0.2730 & 0.1201 \\
Patient Features       & 0.2664 & 0.2880 & 0.2628 \\
\hline
\end{tabular}
\caption{Ablation Studies: Cancer dataset}
\label{tab:abl:cancer2}
\end{table}
\end{minipage}
\begin{minipage}{0.5\textwidth}
\begin{table}[H]
\footnotesize
\centering
\begin{tabular}{lllll|}
\hline
& DCMTF & AE $\rightarrow$ FFN & $\mathcal{C}\rightarrow$ KM \\
\hline

ARI           & 0.4189       & 0.3024 & 0.1274 \\
AMI           & 0.3789       & 0.3441 & 0.1854 \\
NMI           & 0.3796       & 0.3448 & 0.1863 \\
RI            & 0.7354       & 0.6721 & 0.5923 \\
Silhouette:   &              &        &        \\
BOW features  & 0.3892       & 0.4115 & 0.1321 \\
Subjects      & 0.3544       & 0.5929 & 0.1985 \\
Items*         & 0.4395       & 0.5231 & 0.3084 \\
\hline
\end{tabular}
\caption{Ablation Studies: Wiki dataset}
\label{tab:abl:wiki2}
\end{table}
\end{minipage}
\subsection{Sensitivity Analysis}



\begin{table}[H]
\footnotesize
\centering
\begin{tabular}{llllll}
 \hline
$l_e$ & 20 & 50 & 100 & 200 & 300 \\
 \hline
ARI & 0.2749 & 0.2435 & 0.4189 & 0.2288 & 0.2651\\
AMI & 0.2955 & 0.3023 & 0.3789 & 0.2406 & 0.2676\\
NMI & 0.2963 & 0.3030 & 0.3796 & 0.2414 & 0.2684\\
RI & 0.6668 & 0.6456 & 0.7354 & 0.6437 & 0.6552\\
\hdashline
Silhouette: & & & & &\\
BOW features & 0.3667 & 0.3717 & 0.3892 & 0.3563 & 0.3561\\
Subjects & 0.5312 & 0.3194 & 0.3544 & 0.3661 & 0.4328\\
Items* & 0.3641 & 0.3965 & 0.4395 & 0.4000 & 0.4223\\
\hline
\end{tabular}
\caption{Sensitivity analysis on Wiki Dataset, $l_e$: representation dimension. *Labels available.}
\label{tab:sensitivity:reprn_dim2}
\end{table}




\subsection{CFRM with K-Means Initialization: 20NG and Wiki Data}

We have performed an experiment using the 20-Newsgroup dataset (``TM3" from section *6.2* in the paper by \citep{cfrm}). It consists of 2 matrices: a word-document matrix and a document-category matrix. In \citep{cfrm}, authors sampled 100 documents from each newsgroup category. Since those random samples are unavailable, we use all the available 7640 documents from the 8 groups in TM3.  In \citep{cfrm}, authors have used k-means to initialize their model. This improves performance over random initialization. We did not use k-means initialization because it is not possible to do so in all baselines. In Table \ref{tab:20ng} we show performance of CFRM (with both initializations), DFMF and DCMTF. DCMTF outperforms all three by a large margin. Also, in our expriments, CFRM performed poorly on wiki dataset. With k-means initialization, performance improves but not above that of DCMTF as shown in Table \ref{tab:20ng:wiki}. 

\begin{minipage}{0.5\textwidth}
\begin{table}[H]
\centering
\footnotesize
\begin{tabular}{lllll}
\toprule
 & DCMTF & CFRM & CFRM-k & DFMF \\
\hline
ARI: & \textbf{0.5150} & 0.1239 & 0.1938 & 0.1135 \\
AMI: & \textbf{0.6814} & 0.1796 & 0.2305 & 0.1788 \\
NMI: & \textbf{0.6816} & 0.1805 & 0.2390 & 0.1830 \\
RI:  & \textbf{0.7963} & 0.6687 & 0.6790 & 0.6553 \\
\hline
\end{tabular}
\caption{20NG dataset, CFRM-k: CFRM with K-Means initialization}
\label{tab:20ng}
\end{table}
\end{minipage}
\begin{minipage}{0.5\textwidth}
\begin{table}[H]
\centering
\footnotesize
\begin{tabular}{lllll}
\toprule
& CFRM & CFRM-k & DCMTF \\
\hline
ARI: & 0.0103 & 0.0146 & \textbf{0.4189} \\
AMI: & 0.0156 & 0.0370 & \textbf{0.3789} \\
NMI: & 0.0247 & 0.0505 & \textbf{0.3796} \\
RI:  & 0.4347 & 0.4533 & \textbf{0.7354} \\
\hline
\end{tabular}
\caption{Wiki dataset, CFRM-k: CFRM with K-Means initialization}
\label{tab:20ng:wiki}
\end{table}
\end{minipage}

\subsection{Performance with Varying Number of Clusters}

We evaluate the clustering performance for different input number of clusters on Freebase dataset. At 4 and 6 clusters, for entity ``books" [where ground truth labels are available], ARI is not very low. The performance trend remains the same: DCMTF (Table \ref{tab:small_ari_dcmtf}) surpasses CFRM (Table \ref{tab:small_ari_cfrm}).

\begin{minipage}{0.5\textwidth}
\begin{table}[H]
\centering
\footnotesize
\begin{tabular}{llll}
\toprule
 & 4 & 6 & 8\\
\hline
ARI & 0.1595 & 0.1215 & 0.0660\\
AMI & 0.1720 & 0.1586 & 0.1473\\
NMI & 0.1754 & 0.1657 & 0.1595\\
RI & 0.6015 & 0.6492 & 0.6608\\
\hdashline
Silhouette: & & &\\
Book* & 0.5547 & 0.4765 & 0.4825\\
Film & 0.5298 & 0.3424 & 0.2767\\
Music & 0.4178 & 0.2898 & 0.2398\\
Sports & 0.6191 & 0.4897 & 0.3725\\
People & 0.6006 & 0.6799 & 0.6293\\
Location & 0.3232 & 0.2746 & 0.1980\\
\hline
\end{tabular}
\caption{DCMTF on Freebase}
\label{tab:small_ari_dcmtf}
\end{table}
\end{minipage}
\begin{minipage}{0.5\textwidth}
\begin{table}[H]
\centering
\footnotesize
\begin{tabular}{llll}
\toprule
& 4 & 6 & 8\\
\hline
ARI                                  & 0.0992 & 0.0820 & 0.0419 \\
AMI                                  & 0.0635 & 0.0528 & 0.0509 \\
NMI                                  & 0.0888 & 0.0938 & 0.0805 \\
RI                                   & 0.5370 & 0.5085 & 0.6234 \\
\hdashline
Silhouette:                          &        &        &        \\
Book* & 0.6968 & 0.4982 & 0.3786 \\
Film                                 & 0.7129 & 0.7199 & 0.6855 \\
Music                                & 0.3575 & 0.2649 & 0.2283 \\
Sports                               & 0.4119 & 0.3554 & 0.3316 \\
People                               & 0.6188 & 0.6656 & 0.5850 \\
Location                             & 0.2564 & 0.1776 & 0.1530 \\
\hline
\end{tabular}
\caption{CFRM on Freebase}
\label{tab:small_ari_cfrm}
\end{table}
\end{minipage}
\subsection{DCMTF: More entity statistics}

Clustering performance on a single entity where true labels are available are shown in Table \ref{tab:clust_perf_all}. For other entities we can compare the Silhouette Coefficient given in Table \ref{tab:clust_perf_others}. Overall, all input entities appear to be well clustered.

\begin{table}[H]
\centering
\begin{tabular}{cccccccc}
\hline
Cancer  & Silhouette & Freebase & Silhouette & Pubmed & Silhouette & Wiki & Silhouette \\
\hline
Disease          & 0.3931     & Book              & 0.4825     & Gene            & 0.0948     & BOW features  & 0.3892     \\
Disease Features & 0.3622     & Film              & 0.2767     & Disease         & 0.4706     & Subjects      & 0.3544     \\
Gene             & 0.3038     & Music             & 0.2398     & Chemical        & 0.1257     & Items         & 0.4395     \\
Gene Features    & 0.3535     & Sports            & 0.3725     & Species         & 0.2031     &               &            \\
Patient          & 0.2248     & People            & 0.6293     &                 &            &               &            \\
Patient Features & 0.2664     & Location          & 0.1982     &                 &            &               &			   \\
\hline
\end{tabular}
\caption{Clustering performance on other views}
\label{tab:clust_perf_others}
\end{table}

\end{document}

%% file: dcmtf_setup_v1.pdf_tex
\begingroup%
  \makeatletter%
  \providecommand\color[2][]{%
    \errmessage{(Inkscape) Color is used for the text in Inkscape, but the package 'color.sty' is not loaded}%
    \renewcommand\color[2][]{}%
  }%
  \providecommand\transparent[1]{%
    \errmessage{(Inkscape) Transparency is used (non-zero) for the text in Inkscape, but the package 'transparent.sty' is not loaded}%
    \renewcommand\transparent[1]{}%
  }%
  \providecommand\rotatebox[2]{#2}%
  \newcommand*\fsize{\dimexpr\f@size pt\relax}%
  \newcommand*\lineheight[1]{\fontsize{\fsize}{#1\fsize}\selectfont}%
  \ifx\svgwidth\undefined%
    \setlength{\unitlength}{137.90080153bp}%
    \ifx\svgscale\undefined%
      \relax%
    \else%
      \setlength{\unitlength}{\unitlength * \real{\svgscale}}%
    \fi%
  \else%
    \setlength{\unitlength}{\svgwidth}%
  \fi%
  \global\let\svgwidth\undefined%
  \global\let\svgscale\undefined%
  \makeatother%
  \begin{picture}(1,0.91254439)%
    \lineheight{1}%
    \setlength\tabcolsep{0pt}%
    \put(0,0){\includegraphics[width=\unitlength,page=1]{dcmtf_setup_v1.pdf}}%
    \put(0.2520713,0.84145656){\color[rgb]{0,0,0}\makebox(0,0)[lt]{\begin{minipage}{0.26322288\unitlength}\raggedright $E^{[2]}$\end{minipage}}}%
    \put(0.27432127,0.33514781){\color[rgb]{0,0,0}\makebox(0,0)[lt]{\begin{minipage}{0.28329762\unitlength}\raggedright $X^{(1)}$\end{minipage}}}%
    \put(0.6727263,0.33596975){\color[rgb]{0,0,0}\makebox(0,0)[lt]{\begin{minipage}{0.28329762\unitlength}\raggedright $X^{(2)}$\end{minipage}}}%
    \put(0.27842109,0.69686582){\color[rgb]{0,0,0}\makebox(0,0)[lt]{\begin{minipage}{0.28329762\unitlength}\raggedright $X^{(3)}$\end{minipage}}}%
    \put(0.01355645,0.69176643){\color[rgb]{0,0,0}\makebox(0,0)[lt]{\begin{minipage}{0.26322288\unitlength}\raggedright $E^{[4]}$\end{minipage}}}%
    \put(0.0057869,0.32853949){\color[rgb]{0,0,0}\makebox(0,0)[lt]{\begin{minipage}{0.26322288\unitlength}\raggedright $E^{[1]}$\end{minipage}}}%
    \put(0.27694459,0.4839307){\color[rgb]{0,0,0}\makebox(0,0)[lt]{\begin{minipage}{0.26322288\unitlength}\raggedright $E^{[2]}$\end{minipage}}}%
    \put(0.66231474,0.4839307){\color[rgb]{0,0,0}\makebox(0,0)[lt]{\begin{minipage}{0.26322288\unitlength}\raggedright $E^{[3]}$\end{minipage}}}%
  \end{picture}%
\endgroup%

%% file: dcmtf_chain_v1.pdf_tex
\begingroup%
  \makeatletter%
  \providecommand\color[2][]{%
    \errmessage{(Inkscape) Color is used for the text in Inkscape, but the package 'color.sty' is not loaded}%
    \renewcommand\color[2][]{}%
  }%
  \providecommand\transparent[1]{%
    \errmessage{(Inkscape) Transparency is used (non-zero) for the text in Inkscape, but the package 'transparent.sty' is not loaded}%
    \renewcommand\transparent[1]{}%
  }%
  \providecommand\rotatebox[2]{#2}%
  \newcommand*\fsize{\dimexpr\f@size pt\relax}%
  \newcommand*\lineheight[1]{\fontsize{\fsize}{#1\fsize}\selectfont}%
  \ifx\svgwidth\undefined%
    \setlength{\unitlength}{137.90079072bp}%
    \ifx\svgscale\undefined%
      \relax%
    \else%
      \setlength{\unitlength}{\unitlength * \real{\svgscale}}%
    \fi%
  \else%
    \setlength{\unitlength}{\svgwidth}%
  \fi%
  \global\let\svgwidth\undefined%
  \global\let\svgscale\undefined%
  \makeatother%
  \begin{picture}(1,0.91254446)%
    \lineheight{1}%
    \setlength\tabcolsep{0pt}%
    \put(0,0){\includegraphics[width=\unitlength,page=1]{dcmtf_chain_v1.pdf}}%
    \put(0.20045446,0.78225434){\color[rgb]{0,0,0}\makebox(0,0)[lt]{\lineheight{1.25}\smash{\begin{tabular}[t]{l}$1$\end{tabular}}}}%
    \put(0.31480061,0.78225434){\color[rgb]{0,0,0}\makebox(0,0)[lt]{\lineheight{1.25}\smash{\begin{tabular}[t]{l}$2$\end{tabular}}}}%
    \put(0.43134403,0.78225434){\color[rgb]{0,0,0}\makebox(0,0)[lt]{\lineheight{1.25}\smash{\begin{tabular}[t]{l}$3$\end{tabular}}}}%
    \put(0.1968953,0.41473367){\color[rgb]{0,0,0}\makebox(0,0)[lt]{\lineheight{1.25}\smash{\begin{tabular}[t]{l}$1$\end{tabular}}}}%
    \put(0.31124141,0.41473367){\color[rgb]{0,0,0}\makebox(0,0)[lt]{\lineheight{1.25}\smash{\begin{tabular}[t]{l}$2$\end{tabular}}}}%
    \put(0.42778483,0.41473367){\color[rgb]{0,0,0}\makebox(0,0)[lt]{\lineheight{1.25}\smash{\begin{tabular}[t]{l}$3$\end{tabular}}}}%
    \put(0.13839513,0.35794535){\color[rgb]{0,0,0}\makebox(0,0)[lt]{\lineheight{1.25}\smash{\begin{tabular}[t]{l}$1$\end{tabular}}}}%
    \put(0.14170932,0.2780416){\color[rgb]{0,0,0}\makebox(0,0)[lt]{\lineheight{1.25}\smash{\begin{tabular}[t]{l}$2$\end{tabular}}}}%
    \put(0.14153936,0.1988744){\color[rgb]{0,0,0}\makebox(0,0)[lt]{\lineheight{1.25}\smash{\begin{tabular}[t]{l}$3$\end{tabular}}}}%
    \put(0.14408875,0.12058527){\color[rgb]{0,0,0}\makebox(0,0)[lt]{\lineheight{1.25}\smash{\begin{tabular}[t]{l}$4$\end{tabular}}}}%
    \put(0.54257782,0.3581395){\color[rgb]{0,0,0}\makebox(0,0)[lt]{\lineheight{1.25}\smash{\begin{tabular}[t]{l}$1$\end{tabular}}}}%
    \put(0.54589198,0.27823576){\color[rgb]{0,0,0}\makebox(0,0)[lt]{\lineheight{1.25}\smash{\begin{tabular}[t]{l}$2$\end{tabular}}}}%
    \put(0.54572206,0.19906855){\color[rgb]{0,0,0}\makebox(0,0)[lt]{\lineheight{1.25}\smash{\begin{tabular}[t]{l}$3$\end{tabular}}}}%
    \put(0.54827145,0.12077942){\color[rgb]{0,0,0}\makebox(0,0)[lt]{\lineheight{1.25}\smash{\begin{tabular}[t]{l}$4$\end{tabular}}}}%
    \put(0.14361082,0.72253243){\color[rgb]{0,0,0}\makebox(0,0)[lt]{\lineheight{1.25}\smash{\begin{tabular}[t]{l}$1$\end{tabular}}}}%
    \put(0.14692505,0.64262882){\color[rgb]{0,0,0}\makebox(0,0)[lt]{\lineheight{1.25}\smash{\begin{tabular}[t]{l}$2$\end{tabular}}}}%
    \put(0.14675505,0.56346162){\color[rgb]{0,0,0}\makebox(0,0)[lt]{\lineheight{1.25}\smash{\begin{tabular}[t]{l}$3$\end{tabular}}}}%
    \put(0.14930444,0.48517249){\color[rgb]{0,0,0}\makebox(0,0)[lt]{\lineheight{1.25}\smash{\begin{tabular}[t]{l}$4$\end{tabular}}}}%
    \put(0,0){\includegraphics[width=\unitlength,page=2]{dcmtf_chain_v1.pdf}}%
    \put(0.58090622,0.40948944){\color[rgb]{0,0,0}\makebox(0,0)[lt]{\lineheight{1.25}\smash{\begin{tabular}[t]{l}$1$\end{tabular}}}}%
    \put(0.69525233,0.40948944){\color[rgb]{0,0,0}\makebox(0,0)[lt]{\lineheight{1.25}\smash{\begin{tabular}[t]{l}$2$\end{tabular}}}}%
    \put(0.76828617,0.40948944){\color[rgb]{0,0,0}\makebox(0,0)[lt]{\lineheight{1.25}\smash{\begin{tabular}[t]{l}$3$\end{tabular}}}}%
    \put(0.86876623,0.40346801){\color[rgb]{0,0,0}\makebox(0,0)[lt]{\lineheight{1.25}\smash{\begin{tabular}[t]{l}$4$\end{tabular}}}}%
    \put(0,0){\includegraphics[width=\unitlength,page=3]{dcmtf_chain_v1.pdf}}%
    \put(0.01666435,0.30212303){\color[rgb]{0,0,0}\makebox(0,0)[lt]{\begin{minipage}{0.2632229\unitlength}\raggedright $E^{[1]}$\end{minipage}}}%
    \put(0.0193837,0.68593933){\color[rgb]{0,0,0}\makebox(0,0)[lt]{\begin{minipage}{0.2632229\unitlength}\raggedright $E^{[4]}$\end{minipage}}}%
    \put(0.27694465,0.90154463){\color[rgb]{0,0,0}\makebox(0,0)[lt]{\begin{minipage}{0.2632229\unitlength}\raggedright $E^{[2]}$\end{minipage}}}%
    \put(0.66619958,0.53132505){\color[rgb]{0,0,0}\makebox(0,0)[lt]{\begin{minipage}{0.2632229\unitlength}\raggedright $E^{[3]}$\end{minipage}}}%
    \put(0.29248376,0.0954527){\color[rgb]{0,0,0}\makebox(0,0)[lt]{\begin{minipage}{0.2632229\unitlength}\raggedright $E^{[2]}$\end{minipage}}}%
  \end{picture}%
\endgroup%

%% file: dcmtf_model_v1_a.pdf_tex
\begingroup%
  \makeatletter%
  \providecommand\color[2][]{%
    \errmessage{(Inkscape) Color is used for the text in Inkscape, but the package 'color.sty' is not loaded}%
    \renewcommand\color[2][]{}%
  }%
  \providecommand\transparent[1]{%
    \errmessage{(Inkscape) Transparency is used (non-zero) for the text in Inkscape, but the package 'transparent.sty' is not loaded}%
    \renewcommand\transparent[1]{}%
  }%
  \providecommand\rotatebox[2]{#2}%
  \newcommand*\fsize{\dimexpr\f@size pt\relax}%
  \newcommand*\lineheight[1]{\fontsize{\fsize}{#1\fsize}\selectfont}%
  \ifx\svgwidth\undefined%
    \setlength{\unitlength}{318.85429803bp}%
    \ifx\svgscale\undefined%
      \relax%
    \else%
      \setlength{\unitlength}{\unitlength * \real{\svgscale}}%
    \fi%
  \else%
    \setlength{\unitlength}{\svgwidth}%
  \fi%
  \global\let\svgwidth\undefined%
  \global\let\svgscale\undefined%
  \makeatother%
  \begin{picture}(1,0.60810768)%
    \lineheight{1}%
    \setlength\tabcolsep{0pt}%
    \put(12.47038421,2.23825501){\color[rgb]{0,0,0}\makebox(0,0)[lt]{\begin{minipage}{1.70689864\unitlength}\raggedright \end{minipage}}}%
    \put(0,0){\includegraphics[width=\unitlength,page=1]{dcmtf_model_v1_a.pdf}}%
    \put(0.34529441,0.27525835){\color[rgb]{0,0,0}\makebox(0,0)[lt]{\begin{minipage}{0.13932415\unitlength}\raggedright $X^{(1)}$\end{minipage}}}%
    \put(0.35722954,0.41592316){\color[rgb]{0,0,0}\makebox(0,0)[lt]{\begin{minipage}{0.13932415\unitlength}\raggedright $X^{(3)}$\end{minipage}}}%
    \put(0,0){\includegraphics[width=\unitlength,page=2]{dcmtf_model_v1_a.pdf}}%
    \put(0.52625136,0.27525835){\color[rgb]{0,0,0}\makebox(0,0)[lt]{\begin{minipage}{0.13932415\unitlength}\raggedright $X^{(2)}$\end{minipage}}}%
    \put(0.26344593,0.26450426){\color[rgb]{0,0,0}\makebox(0,0)[lt]{\begin{minipage}{0.10956918\unitlength}\raggedright $E^{[1]}$\end{minipage}}}%
    \put(0.25917744,0.41276228){\color[rgb]{0,0,0}\makebox(0,0)[lt]{\begin{minipage}{0.11123242\unitlength}\raggedright $E^{[4]}$\end{minipage}}}%
    \put(0.36474423,0.48511309){\color[rgb]{0,0,0}\makebox(0,0)[lt]{\begin{minipage}{0.11304387\unitlength}\raggedright $E^{[2]}$\end{minipage}}}%
    \put(0.52885835,0.33625339){\color[rgb]{0,0,0}\makebox(0,0)[lt]{\begin{minipage}{0.11523381\unitlength}\raggedright $E^{[3]}$\end{minipage}}}%
  \end{picture}%
\endgroup%

%% file: dcmtf_model_v1_b.pdf_tex
\begingroup%
  \makeatletter%
  \providecommand\color[2][]{%
    \errmessage{(Inkscape) Color is used for the text in Inkscape, but the package 'color.sty' is not loaded}%
    \renewcommand\color[2][]{}%
  }%
  \providecommand\transparent[1]{%
    \errmessage{(Inkscape) Transparency is used (non-zero) for the text in Inkscape, but the package 'transparent.sty' is not loaded}%
    \renewcommand\transparent[1]{}%
  }%
  \providecommand\rotatebox[2]{#2}%
  \newcommand*\fsize{\dimexpr\f@size pt\relax}%
  \newcommand*\lineheight[1]{\fontsize{\fsize}{#1\fsize}\selectfont}%
  \ifx\svgwidth\undefined%
    \setlength{\unitlength}{318.85429803bp}%
    \ifx\svgscale\undefined%
      \relax%
    \else%
      \setlength{\unitlength}{\unitlength * \real{\svgscale}}%
    \fi%
  \else%
    \setlength{\unitlength}{\svgwidth}%
  \fi%
  \global\let\svgwidth\undefined%
  \global\let\svgscale\undefined%
  \makeatother%
  \begin{picture}(1,0.60810768)%
    \lineheight{1}%
    \setlength\tabcolsep{0pt}%
    \put(12.41006623,2.80134203){\color[rgb]{0,0,0}\makebox(0,0)[lt]{\begin{minipage}{1.70689864\unitlength}\raggedright \end{minipage}}}%
    \put(0,0){\includegraphics[width=\unitlength,page=1]{dcmtf_model_v1_b.pdf}}%
    \put(0.59904017,0.28456349){\color[rgb]{0,0,0}\makebox(0,0)[lt]{\begin{minipage}{0.39047669\unitlength}\raggedright $X^{(2)}$\end{minipage}}}%
    \put(0.59567989,0.41863736){\color[rgb]{0,0,0}\makebox(0,0)[lt]{\begin{minipage}{0.39047669\unitlength}\raggedright $X^{(1)}$\end{minipage}}}%
    \put(0.59904017,0.16124255){\color[rgb]{0,0,0}\makebox(0,0)[lt]{\begin{minipage}{0.39047669\unitlength}\raggedright $X^{(3)}$\end{minipage}}}%
    \put(0.34166284,0.53585564){\color[rgb]{0,0,0}\makebox(0,0)[lt]{\lineheight{0.25}\smash{\begin{tabular}[t]{l}$G$\end{tabular}}}}%
    \put(0,0){\includegraphics[width=\unitlength,page=2]{dcmtf_model_v1_b.pdf}}%
    \put(0.23323176,0.46185294){\color[rgb]{0,0,0}\makebox(0,0)[lt]{\lineheight{0.25}\smash{\begin{tabular}[t]{l}$q_{1}$\end{tabular}}}}%
    \put(0.22389673,0.27189697){\color[rgb]{0,0,0}\makebox(0,0)[lt]{\lineheight{0.25}\smash{\begin{tabular}[t]{l}$q_{5}$\end{tabular}}}}%
    \put(0.24245719,0.08607528){\color[rgb]{0,0,0}\makebox(0,0)[lt]{\lineheight{0.25}\smash{\begin{tabular}[t]{l}$q_{6}$\end{tabular}}}}%
    \put(0.22389618,0.38480091){\color[rgb]{0,0,0}\makebox(0,0)[lt]{\lineheight{0.25}\smash{\begin{tabular}[t]{l}$q_{2}$\end{tabular}}}}%
    \put(0.22834449,0.19192327){\color[rgb]{0,0,0}\makebox(0,0)[lt]{\lineheight{0.25}\smash{\begin{tabular}[t]{l}$q_{3}$\end{tabular}}}}%
    \put(0.25212268,0.31658783){\color[rgb]{0,0,0}\makebox(0,0)[lt]{\lineheight{0.25}\smash{\begin{tabular}[t]{l}$q_{4}$\end{tabular}}}}%
    \put(0.12467329,0.4676973){\color[rgb]{0,0,0}\makebox(0,0)[lt]{\begin{minipage}{0.13217569\unitlength}\raggedright $E^{[1]}$\end{minipage}}}%
    \put(0.12232097,0.35008869){\color[rgb]{0,0,0}\makebox(0,0)[lt]{\begin{minipage}{0.12747136\unitlength}\raggedright $E^{[2]}$\end{minipage}}}%
    \put(0.12349713,0.22895179){\color[rgb]{0,0,0}\makebox(0,0)[lt]{\begin{minipage}{0.11571049\unitlength}\raggedright $E^{[3]}$\end{minipage}}}%
    \put(0.12232106,0.10663898){\color[rgb]{0,0,0}\makebox(0,0)[lt]{\begin{minipage}{0.11806264\unitlength}\raggedright $E^{[4]}$\end{minipage}}}%
  \end{picture}%
\endgroup%

%% file: dcmtf_model_v3_pass_1.pdf_tex
\begingroup%
  \makeatletter%
  \providecommand\color[2][]{%
    \errmessage{(Inkscape) Color is used for the text in Inkscape, but the package 'color.sty' is not loaded}%
    \renewcommand\color[2][]{}%
  }%
  \providecommand\transparent[1]{%
    \errmessage{(Inkscape) Transparency is used (non-zero) for the text in Inkscape, but the package 'transparent.sty' is not loaded}%
    \renewcommand\transparent[1]{}%
  }%
  \providecommand\rotatebox[2]{#2}%
  \newcommand*\fsize{\dimexpr\f@size pt\relax}%
  \newcommand*\lineheight[1]{\fontsize{\fsize}{#1\fsize}\selectfont}%
  \ifx\svgwidth\undefined%
    \setlength{\unitlength}{496.97810135bp}%
    \ifx\svgscale\undefined%
      \relax%
    \else%
      \setlength{\unitlength}{\unitlength * \real{\svgscale}}%
    \fi%
  \else%
    \setlength{\unitlength}{\svgwidth}%
  \fi%
  \global\let\svgwidth\undefined%
  \global\let\svgscale\undefined%
  \makeatother%
  \begin{picture}(1,0.97276213)%
    \lineheight{1}%
    \setlength\tabcolsep{0pt}%
    \put(7.96591282,0.80559281){\color[rgb]{0,0,0}\makebox(0,0)[lt]{\begin{minipage}{1.09512264\unitlength}\raggedright \end{minipage}}}%
    \put(0,0){\includegraphics[width=\unitlength,page=1]{dcmtf_model_v3_pass_1.pdf}}%
    \put(0.05984332,0.79942962){\color[rgb]{0,0,0}\makebox(0,0)[lt]{\begin{minipage}{0.25052446\unitlength}\raggedright $Y^{(1)}_{[1]}$\end{minipage}}}%
    \put(0.1345632,0.70137017){\color[rgb]{0,0,0}\makebox(0,0)[lt]{\begin{minipage}{0.25052446\unitlength}\raggedright $\mu_{\epsilon}^{(1,1)}$\end{minipage}}}%
    \put(0,0){\includegraphics[width=\unitlength,page=2]{dcmtf_model_v3_pass_1.pdf}}%
    \put(0.07426232,0.95336572){\color[rgb]{0,0,0}\makebox(0,0)[lt]{\begin{minipage}{0.25052446\unitlength}\raggedright $(1,1)$\end{minipage}}}%
    \put(0,0){\includegraphics[width=\unitlength,page=3]{dcmtf_model_v3_pass_1.pdf}}%
    \put(0.22279438,0.95336572){\color[rgb]{0,0,0}\makebox(0,0)[lt]{\begin{minipage}{0.25052446\unitlength}\raggedright $(1,2)$\end{minipage}}}%
    \put(0,0){\includegraphics[width=\unitlength,page=4]{dcmtf_model_v3_pass_1.pdf}}%
    \put(0.3713266,0.95336572){\color[rgb]{0,0,0}\makebox(0,0)[lt]{\begin{minipage}{0.25052446\unitlength}\raggedright $(3,2)$\end{minipage}}}%
    \put(0,0){\includegraphics[width=\unitlength,page=5]{dcmtf_model_v3_pass_1.pdf}}%
    \put(0.28415222,0.70358853){\color[rgb]{0,0,0}\makebox(0,0)[lt]{\begin{minipage}{0.25052446\unitlength}\raggedright $\mu_{\epsilon}^{(1,2)}$\end{minipage}}}%
    \put(0,0){\includegraphics[width=\unitlength,page=6]{dcmtf_model_v3_pass_1.pdf}}%
    \put(0.35768354,0.80016195){\color[rgb]{0,0,0}\makebox(0,0)[lt]{\begin{minipage}{0.25052446\unitlength}\raggedright $Y^{(2)}_{[3]}$\end{minipage}}}%
    \put(0.43165902,0.70412208){\color[rgb]{0,0,0}\makebox(0,0)[lt]{\begin{minipage}{0.25052446\unitlength}\raggedright $\mu_{\epsilon}^{(3,2)}$\end{minipage}}}%
    \put(0,0){\includegraphics[width=\unitlength,page=7]{dcmtf_model_v3_pass_1.pdf}}%
    \put(0.0725472,0.44532873){\color[rgb]{0,0,0}\makebox(0,0)[lt]{\begin{minipage}{0.29801083\unitlength}\raggedright $\Gamma_{\ell \in \mathcal{N}[1]}[ \mu_{\epsilon}^{(1,\ell)}$ ]\end{minipage}}}%
    \put(0.22424598,0.49956994){\color[rgb]{0,0,0}\makebox(0,0)[lt]{\begin{minipage}{0.25052446\unitlength}\raggedright Matrix concatenation\end{minipage}}}%
    \put(0,0){\includegraphics[width=\unitlength,page=8]{dcmtf_model_v3_pass_1.pdf}}%
    \put(0.51985877,0.95336572){\color[rgb]{0,0,0}\makebox(0,0)[lt]{\begin{minipage}{0.25052446\unitlength}\raggedright $(2,1)$\end{minipage}}}%
    \put(0,0){\includegraphics[width=\unitlength,page=9]{dcmtf_model_v3_pass_1.pdf}}%
    \put(0.50707886,0.8006955){\color[rgb]{0,0,0}\makebox(0,0)[lt]{\begin{minipage}{0.25052446\unitlength}\raggedright $Y^{(1)}_{[2]}$\end{minipage}}}%
    \put(0,0){\includegraphics[width=\unitlength,page=10]{dcmtf_model_v3_pass_1.pdf}}%
    \put(0.58193751,0.70542274){\color[rgb]{0,0,0}\makebox(0,0)[lt]{\begin{minipage}{0.25052446\unitlength}\raggedright $\mu_{\epsilon}^{(2,1)}$\end{minipage}}}%
    \put(0,0){\includegraphics[width=\unitlength,page=11]{dcmtf_model_v3_pass_1.pdf}}%
    \put(0.67489927,0.95336572){\color[rgb]{0,0,0}\makebox(0,0)[lt]{\begin{minipage}{0.25052446\unitlength}\raggedright $(2,3)$\end{minipage}}}%
    \put(0,0){\includegraphics[width=\unitlength,page=12]{dcmtf_model_v3_pass_1.pdf}}%
    \put(0.66211936,0.8006955){\color[rgb]{0,0,0}\makebox(0,0)[lt]{\begin{minipage}{0.25052446\unitlength}\raggedright $Y^{(3)}_{[2]}$\end{minipage}}}%
    \put(0,0){\includegraphics[width=\unitlength,page=13]{dcmtf_model_v3_pass_1.pdf}}%
    \put(0.73697801,0.70542274){\color[rgb]{0,0,0}\makebox(0,0)[lt]{\begin{minipage}{0.25052446\unitlength}\raggedright $\mu_{\epsilon}^{(2,3)}$\end{minipage}}}%
    \put(0,0){\includegraphics[width=\unitlength,page=14]{dcmtf_model_v3_pass_1.pdf}}%
    \put(0.82692146,0.95336572){\color[rgb]{0,0,0}\makebox(0,0)[lt]{\begin{minipage}{0.25052446\unitlength}\raggedright $(4,3)$\end{minipage}}}%
    \put(0,0){\includegraphics[width=\unitlength,page=15]{dcmtf_model_v3_pass_1.pdf}}%
    \put(0.81414155,0.8006955){\color[rgb]{0,0,0}\makebox(0,0)[lt]{\begin{minipage}{0.25052446\unitlength}\raggedright $Y^{(3)}_{[4]}$\end{minipage}}}%
    \put(0,0){\includegraphics[width=\unitlength,page=16]{dcmtf_model_v3_pass_1.pdf}}%
    \put(0.8890002,0.70542274){\color[rgb]{0,0,0}\makebox(0,0)[lt]{\begin{minipage}{0.25052446\unitlength}\raggedright $\mu_{\epsilon}^{(4,3)}$\end{minipage}}}%
    \put(0,0){\includegraphics[width=\unitlength,page=17]{dcmtf_model_v3_pass_1.pdf}}%
    \put(0.20824122,0.32073843){\color[rgb]{0,0,0}\makebox(0,0)[lt]{\begin{minipage}{0.25052446\unitlength}\raggedright $U^{[1]}$\end{minipage}}}%
    \put(0,0){\includegraphics[width=\unitlength,page=18]{dcmtf_model_v3_pass_1.pdf}}%
    \put(0.65884789,0.32073843){\color[rgb]{0,0,0}\makebox(0,0)[lt]{\begin{minipage}{0.25052446\unitlength}\raggedright $U^{[2]}$\end{minipage}}}%
    \put(0,0){\includegraphics[width=\unitlength,page=19]{dcmtf_model_v3_pass_1.pdf}}%
    \put(0.01042437,0.55363774){\color[rgb]{0,0,0}\makebox(0,0)[lt]{\begin{minipage}{0.2644562\unitlength}\raggedright ${\ell_\mathcal{A}}(Y^{(1)}_{[1]},Y^{(1)^\prime}_{[1]})$\end{minipage}}}%
    \put(0.16589431,0.55395033){\color[rgb]{0,0,0}\makebox(0,0)[lt]{\begin{minipage}{0.30654772\unitlength}\raggedright ${\ell_\mathcal{A}}(Y^{(2)}_{[1]},Y^{(2)^\prime}_{[1]})$\end{minipage}}}%
    \put(0.32779006,0.55548627){\color[rgb]{0,0,0}\makebox(0,0)[lt]{\begin{minipage}{0.27560153\unitlength}\raggedright ${\ell_\mathcal{A}}(Y^{(2)}_{[3]},Y^{(2)^\prime}_{[3]})$\end{minipage}}}%
    \put(0.48351028,0.55636788){\color[rgb]{0,0,0}\makebox(0,0)[lt]{\begin{minipage}{0.28733977\unitlength}\raggedright ${\ell_\mathcal{A}}(Y^{(1)}_{[2]},Y^{(1)^\prime}_{[2]})$\end{minipage}}}%
    \put(0.62330166,0.55743499){\color[rgb]{0,0,0}\makebox(0,0)[lt]{\begin{minipage}{0.29000753\unitlength}\raggedright ${\ell_\mathcal{A}}(Y^{(3)}_{[2]},Y^{(3)^\prime}_{[2]})$\end{minipage}}}%
    \put(0.76922886,0.55903565){\color[rgb]{0,0,0}\makebox(0,0)[lt]{\begin{minipage}{0.27453444\unitlength}\raggedright ${\ell_\mathcal{A}}(Y^{(3)}_{[4]},Y^{(3)^\prime}_{[4]})$\end{minipage}}}%
    \put(0,0){\includegraphics[width=\unitlength,page=20]{dcmtf_model_v3_pass_1.pdf}}%
    \put(0.46306052,0.01265213){\color[rgb]{0,0,0}\makebox(0,0)[lt]{\begin{minipage}{0.24839029\unitlength}\raggedright ${\ell_\mathcal{M}}({X^{(1)}}, X^{(1)^{\prime}})$\end{minipage}}}%
    \put(0,0){\includegraphics[width=\unitlength,page=21]{dcmtf_model_v3_pass_1.pdf}}%
    \put(0.00279658,0.33330509){\color[rgb]{0,0,0}\rotatebox{90}{\makebox(0,0)[lt]{\begin{minipage}{0.30121219\unitlength}\raggedright \textsc{Data Fusion}\end{minipage}}}}%
    \put(0.00385231,0.12537779){\color[rgb]{0,0,0}\rotatebox{90}{\makebox(0,0)[lt]{\begin{minipage}{0.18095933\unitlength}\raggedright \textsc{Matrix reconstruction}\end{minipage}}}}%
    \put(0.00541234,0.66426985){\color[rgb]{0,0,0}\rotatebox{90}{\makebox(0,0)[lt]{\begin{minipage}{0.22963579\unitlength}\raggedright \textsc{Variational Autoencoder}\end{minipage}}}}%
    \put(0,0){\includegraphics[width=\unitlength,page=22]{dcmtf_model_v3_pass_1.pdf}}%
    \put(0.28527768,0.28483585){\color[rgb]{0,0,0}\makebox(0,0)[lt]{\begin{minipage}{0.25052446\unitlength}\raggedright Matrix multiplication\end{minipage}}}%
    \put(0,0){\includegraphics[width=\unitlength,page=23]{dcmtf_model_v3_pass_1.pdf}}%
    \put(0.29402708,0.20557615){\color[rgb]{0,0,0}\makebox(0,0)[lt]{\begin{minipage}{0.25052446\unitlength}\raggedright $X^{(2)^{\prime}}$\end{minipage}}}%
    \put(0,0){\includegraphics[width=\unitlength,page=24]{dcmtf_model_v3_pass_1.pdf}}%
    \put(0.48751888,0.06759729){\color[rgb]{0,0,0}\makebox(0,0)[lt]{\begin{minipage}{0.25052446\unitlength}\raggedright $X^{(1)^\prime}$\end{minipage}}}%
    \put(0,0){\includegraphics[width=\unitlength,page=25]{dcmtf_model_v3_pass_1.pdf}}%
    \put(0.74935136,0.06910641){\color[rgb]{0,0,0}\makebox(0,0)[lt]{\begin{minipage}{0.25052446\unitlength}\raggedright $X^{(3)^\prime}$\end{minipage}}}%
    \put(0,0){\includegraphics[width=\unitlength,page=26]{dcmtf_model_v3_pass_1.pdf}}%
    \put(0.72430265,0.01211773){\color[rgb]{0,0,0}\makebox(0,0)[lt]{\begin{minipage}{0.25052446\unitlength}\raggedright ${\ell_\mathcal{M}}({X^{(3)}}, X^{(3)^{\prime}})$\end{minipage}}}%
    \put(0,0){\includegraphics[width=\unitlength,page=27]{dcmtf_model_v3_pass_1.pdf}}%
    \put(0.29296094,0.15443172){\color[rgb]{0,0,0}\makebox(0,0)[lt]{\begin{minipage}{0.32148722\unitlength}\raggedright ${\ell_\mathcal{M}}({X^{(2)}}, X^{(2)^{\prime}})$\end{minipage}}}%
    \put(0,0){\includegraphics[width=\unitlength,page=28]{dcmtf_model_v3_pass_1.pdf}}%
    \put(0.07491928,0.96988696){\color[rgb]{0,0,0}\makebox(0,0)[lt]{\begin{minipage}{0.25052446\unitlength}\raggedright $q_{1}$\end{minipage}}}%
    \put(0.22345134,0.96988696){\color[rgb]{0,0,0}\makebox(0,0)[lt]{\begin{minipage}{0.25052446\unitlength}\raggedright $q_{2}$\end{minipage}}}%
    \put(0.37198356,0.96988696){\color[rgb]{0,0,0}\makebox(0,0)[lt]{\begin{minipage}{0.25052446\unitlength}\raggedright $q_{3}$\end{minipage}}}%
    \put(0.52051573,0.96988696){\color[rgb]{0,0,0}\makebox(0,0)[lt]{\begin{minipage}{0.25052446\unitlength}\raggedright $q_{4}$\end{minipage}}}%
    \put(0.67555623,0.96988696){\color[rgb]{0,0,0}\makebox(0,0)[lt]{\begin{minipage}{0.25052446\unitlength}\raggedright $q_{5}$\end{minipage}}}%
    \put(0.21663095,0.80011078){\color[rgb]{0,0,0}\makebox(0,0)[lt]{\begin{minipage}{0.25052446\unitlength}\raggedright $Y^{(2)}_{[1]}$\end{minipage}}}%
    \put(0.82646504,0.96913241){\color[rgb]{0,0,0}\makebox(0,0)[lt]{\begin{minipage}{0.25052446\unitlength}\raggedright $q_{6}$\end{minipage}}}%
    \put(0.06283152,0.61838333){\color[rgb]{0,0,0}\makebox(0,0)[lt]{\begin{minipage}{0.25052446\unitlength}\raggedright $Y^{(1)^\prime}_{[1]}$\end{minipage}}}%
    \put(0.36067174,0.61911569){\color[rgb]{0,0,0}\makebox(0,0)[lt]{\begin{minipage}{0.25052446\unitlength}\raggedright $Y^{(2)^\prime}_{[3]}$\end{minipage}}}%
    \put(0.51006712,0.61964922){\color[rgb]{0,0,0}\makebox(0,0)[lt]{\begin{minipage}{0.25052446\unitlength}\raggedright $Y^{(1)^\prime}_{[2]}$\end{minipage}}}%
    \put(0.66510756,0.61964922){\color[rgb]{0,0,0}\makebox(0,0)[lt]{\begin{minipage}{0.25052446\unitlength}\raggedright $Y^{(3)^\prime}_{[2]}$\end{minipage}}}%
    \put(0.81712975,0.61964922){\color[rgb]{0,0,0}\makebox(0,0)[lt]{\begin{minipage}{0.25052446\unitlength}\raggedright $Y^{(3)^\prime}_{[4]}$\end{minipage}}}%
    \put(0.21961915,0.61906453){\color[rgb]{0,0,0}\makebox(0,0)[lt]{\begin{minipage}{0.25052446\unitlength}\raggedright $Y^{(2)^\prime}_{[1]}$\end{minipage}}}%
    \put(0.52412781,0.44564981){\color[rgb]{0,0,0}\makebox(0,0)[lt]{\begin{minipage}{0.29801083\unitlength}\raggedright $\Gamma_{\ell \in \mathcal{N}[2]}[ \mu_{\epsilon}^{(2,\ell)}$ ]\end{minipage}}}%
    \put(0,0){\includegraphics[width=\unitlength,page=29]{dcmtf_model_v3_pass_1.pdf}}%
    \put(0.43349192,0.32073849){\color[rgb]{0,0,0}\makebox(0,0)[lt]{\begin{minipage}{0.25052446\unitlength}\raggedright $U^{[3]}$\end{minipage}}}%
    \put(0,0){\includegraphics[width=\unitlength,page=30]{dcmtf_model_v3_pass_1.pdf}}%
    \put(0.88711539,0.32073846){\color[rgb]{0,0,0}\makebox(0,0)[lt]{\begin{minipage}{0.25052446\unitlength}\raggedright $U^{[4]}$\end{minipage}}}%
  \end{picture}%
\endgroup%

%% file: dcmtf_model_v3_pass_2.pdf_tex
\begingroup%
  \makeatletter%
  \providecommand\color[2][]{%
    \errmessage{(Inkscape) Color is used for the text in Inkscape, but the package 'color.sty' is not loaded}%
    \renewcommand\color[2][]{}%
  }%
  \providecommand\transparent[1]{%
    \errmessage{(Inkscape) Transparency is used (non-zero) for the text in Inkscape, but the package 'transparent.sty' is not loaded}%
    \renewcommand\transparent[1]{}%
  }%
  \providecommand\rotatebox[2]{#2}%
  \newcommand*\fsize{\dimexpr\f@size pt\relax}%
  \newcommand*\lineheight[1]{\fontsize{\fsize}{#1\fsize}\selectfont}%
  \ifx\svgwidth\undefined%
    \setlength{\unitlength}{496.97810135bp}%
    \ifx\svgscale\undefined%
      \relax%
    \else%
      \setlength{\unitlength}{\unitlength * \real{\svgscale}}%
    \fi%
  \else%
    \setlength{\unitlength}{\svgwidth}%
  \fi%
  \global\let\svgwidth\undefined%
  \global\let\svgscale\undefined%
  \makeatother%
  \begin{picture}(1,0.90784648)%
    \lineheight{1}%
    \setlength\tabcolsep{0pt}%
    \put(7.96591282,0.75576847){\color[rgb]{0,0,0}\makebox(0,0)[lt]{\begin{minipage}{1.09512264\unitlength}\raggedright \end{minipage}}}%
    \put(0,0){\includegraphics[width=\unitlength,page=1]{dcmtf_model_v3_pass_2.pdf}}%
    \put(0.05984332,0.74960528){\color[rgb]{0,0,0}\makebox(0,0)[lt]{\begin{minipage}{0.25052446\unitlength}\raggedright $Y^{(1)}_{[1]}$\end{minipage}}}%
    \put(0.1345632,0.65154583){\color[rgb]{0,0,0}\makebox(0,0)[lt]{\begin{minipage}{0.25052446\unitlength}\raggedright $\mu_{\epsilon}^{(1,1)}$\end{minipage}}}%
    \put(0,0){\includegraphics[width=\unitlength,page=2]{dcmtf_model_v3_pass_2.pdf}}%
    \put(0.07426232,0.88845007){\color[rgb]{0,0,0}\makebox(0,0)[lt]{\begin{minipage}{0.25052446\unitlength}\raggedright $(1,1)$\end{minipage}}}%
    \put(0,0){\includegraphics[width=\unitlength,page=3]{dcmtf_model_v3_pass_2.pdf}}%
    \put(0.22279438,0.88845007){\color[rgb]{0,0,0}\makebox(0,0)[lt]{\begin{minipage}{0.25052446\unitlength}\raggedright $(1,2)$\end{minipage}}}%
    \put(0,0){\includegraphics[width=\unitlength,page=4]{dcmtf_model_v3_pass_2.pdf}}%
    \put(0.3713266,0.88845007){\color[rgb]{0,0,0}\makebox(0,0)[lt]{\begin{minipage}{0.25052446\unitlength}\raggedright $(3,2)$\end{minipage}}}%
    \put(0,0){\includegraphics[width=\unitlength,page=5]{dcmtf_model_v3_pass_2.pdf}}%
    \put(0.28415222,0.65376418){\color[rgb]{0,0,0}\makebox(0,0)[lt]{\begin{minipage}{0.25052446\unitlength}\raggedright $\mu_{\epsilon}^{(1,2)}$\end{minipage}}}%
    \put(0,0){\includegraphics[width=\unitlength,page=6]{dcmtf_model_v3_pass_2.pdf}}%
    \put(0.35768354,0.7503376){\color[rgb]{0,0,0}\makebox(0,0)[lt]{\begin{minipage}{0.25052446\unitlength}\raggedright $Y^{(2)}_{[3]}$\end{minipage}}}%
    \put(0.43165902,0.65429774){\color[rgb]{0,0,0}\makebox(0,0)[lt]{\begin{minipage}{0.25052446\unitlength}\raggedright $\mu_{\epsilon}^{(3,2)}$\end{minipage}}}%
    \put(0,0){\includegraphics[width=\unitlength,page=7]{dcmtf_model_v3_pass_2.pdf}}%
    \put(0.14800372,0.39550439){\color[rgb]{0,0,0}\makebox(0,0)[lt]{\begin{minipage}{0.29801083\unitlength}\raggedright $\Gamma_{\ell \in \mathcal{N}[1]}[ \mu_{\epsilon}^{(1,\ell)}$ ]\end{minipage}}}%
    \put(0.22424598,0.44974559){\color[rgb]{0,0,0}\makebox(0,0)[lt]{\begin{minipage}{0.25052446\unitlength}\raggedright Matrix concatenation\end{minipage}}}%
    \put(0,0){\includegraphics[width=\unitlength,page=8]{dcmtf_model_v3_pass_2.pdf}}%
    \put(0.51985877,0.88845007){\color[rgb]{0,0,0}\makebox(0,0)[lt]{\begin{minipage}{0.25052446\unitlength}\raggedright $(2,1)$\end{minipage}}}%
    \put(0,0){\includegraphics[width=\unitlength,page=9]{dcmtf_model_v3_pass_2.pdf}}%
    \put(0.50707886,0.75087116){\color[rgb]{0,0,0}\makebox(0,0)[lt]{\begin{minipage}{0.25052446\unitlength}\raggedright $Y^{(1)}_{[2]}$\end{minipage}}}%
    \put(0,0){\includegraphics[width=\unitlength,page=10]{dcmtf_model_v3_pass_2.pdf}}%
    \put(0.58193751,0.6555984){\color[rgb]{0,0,0}\makebox(0,0)[lt]{\begin{minipage}{0.25052446\unitlength}\raggedright $\mu_{\epsilon}^{(2,1)}$\end{minipage}}}%
    \put(0,0){\includegraphics[width=\unitlength,page=11]{dcmtf_model_v3_pass_2.pdf}}%
    \put(0.67489927,0.88845007){\color[rgb]{0,0,0}\makebox(0,0)[lt]{\begin{minipage}{0.25052446\unitlength}\raggedright $(2,3)$\end{minipage}}}%
    \put(0,0){\includegraphics[width=\unitlength,page=12]{dcmtf_model_v3_pass_2.pdf}}%
    \put(0.66211936,0.75087116){\color[rgb]{0,0,0}\makebox(0,0)[lt]{\begin{minipage}{0.25052446\unitlength}\raggedright $Y^{(3)}_{[2]}$\end{minipage}}}%
    \put(0,0){\includegraphics[width=\unitlength,page=13]{dcmtf_model_v3_pass_2.pdf}}%
    \put(0.73697801,0.6555984){\color[rgb]{0,0,0}\makebox(0,0)[lt]{\begin{minipage}{0.25052446\unitlength}\raggedright $\mu_{\epsilon}^{(2,3)}$\end{minipage}}}%
    \put(0,0){\includegraphics[width=\unitlength,page=14]{dcmtf_model_v3_pass_2.pdf}}%
    \put(0.82692146,0.88845007){\color[rgb]{0,0,0}\makebox(0,0)[lt]{\begin{minipage}{0.25052446\unitlength}\raggedright $(4,3)$\end{minipage}}}%
    \put(0,0){\includegraphics[width=\unitlength,page=15]{dcmtf_model_v3_pass_2.pdf}}%
    \put(0.81414155,0.75087116){\color[rgb]{0,0,0}\makebox(0,0)[lt]{\begin{minipage}{0.25052446\unitlength}\raggedright $Y^{(3)}_{[4]}$\end{minipage}}}%
    \put(0,0){\includegraphics[width=\unitlength,page=16]{dcmtf_model_v3_pass_2.pdf}}%
    \put(0.8890002,0.6555984){\color[rgb]{0,0,0}\makebox(0,0)[lt]{\begin{minipage}{0.25052446\unitlength}\raggedright $\mu_{\epsilon}^{(4,3)}$\end{minipage}}}%
    \put(0,0){\includegraphics[width=\unitlength,page=17]{dcmtf_model_v3_pass_2.pdf}}%
    \put(0.20824122,0.27091409){\color[rgb]{0,0,0}\makebox(0,0)[lt]{\begin{minipage}{0.25052446\unitlength}\raggedright $U^{[1]}$\end{minipage}}}%
    \put(0,0){\includegraphics[width=\unitlength,page=18]{dcmtf_model_v3_pass_2.pdf}}%
    \put(0.65884789,0.27091409){\color[rgb]{0,0,0}\makebox(0,0)[lt]{\begin{minipage}{0.25052446\unitlength}\raggedright $U^{[2]}$\end{minipage}}}%
    \put(0,0){\includegraphics[width=\unitlength,page=19]{dcmtf_model_v3_pass_2.pdf}}%
    \put(0.02471013,0.50581816){\color[rgb]{0,0,0}\makebox(0,0)[lt]{\begin{minipage}{0.26222015\unitlength}\raggedright ${\ell_\mathcal{A}}(Y^{(1)}_{[1]},Y^{(1)^\prime}_{[1]})$\end{minipage}}}%
    \put(0.1822717,0.50544092){\color[rgb]{0,0,0}\makebox(0,0)[lt]{\begin{minipage}{0.27542495\unitlength}\raggedright ${\ell_\mathcal{A}}(Y^{(2)}_{[1]},Y^{(2)^\prime}_{[1]})$\end{minipage}}}%
    \put(0.33356118,0.50619548){\color[rgb]{0,0,0}\makebox(0,0)[lt]{\begin{minipage}{0.2859888\unitlength}\raggedright ${\ell_\mathcal{A}}(Y^{(2)}_{[3]},Y^{(2)^\prime}_{[3]})$\end{minipage}}}%
    \put(0.48398375,0.5066998){\color[rgb]{0,0,0}\makebox(0,0)[lt]{\begin{minipage}{0.27504767\unitlength}\raggedright ${\ell_\mathcal{A}}(Y^{(1)}_{[2]},Y^{(1)^\prime}_{[2]})$\end{minipage}}}%
    \put(0.63449986,0.50723336){\color[rgb]{0,0,0}\makebox(0,0)[lt]{\begin{minipage}{0.29013888\unitlength}\raggedright ${\ell_\mathcal{A}}(Y^{(3)}_{[2]},Y^{(3)^\prime}_{[2]})$\end{minipage}}}%
    \put(0.77981761,0.50761064){\color[rgb]{0,0,0}\makebox(0,0)[lt]{\begin{minipage}{0.2765568\unitlength}\raggedright ${\ell_\mathcal{A}}(Y^{(3)}_{[4]},Y^{(3)^\prime}_{[4]})$\end{minipage}}}%
    \put(0,0){\includegraphics[width=\unitlength,page=20]{dcmtf_model_v3_pass_2.pdf}}%
    \put(0.00279658,0.28348075){\color[rgb]{0,0,0}\rotatebox{90}{\makebox(0,0)[lt]{\begin{minipage}{0.30121219\unitlength}\raggedright \textsc{Data Fusion}\end{minipage}}}}%
    \put(0.00385231,0.07555344){\color[rgb]{0,0,0}\rotatebox{90}{\makebox(0,0)[lt]{\begin{minipage}{0.18095933\unitlength}\raggedright \textsc{Clustering}\end{minipage}}}}%
    \put(0.00541234,0.6144455){\color[rgb]{0,0,0}\rotatebox{90}{\makebox(0,0)[lt]{\begin{minipage}{0.22963579\unitlength}\raggedright \textsc{Variational Autoencoder}\end{minipage}}}}%
    \put(0,0){\includegraphics[width=\unitlength,page=21]{dcmtf_model_v3_pass_2.pdf}}%
    \put(0.07491928,0.90497131){\color[rgb]{0,0,0}\makebox(0,0)[lt]{\begin{minipage}{0.25052446\unitlength}\raggedright $q_{1}$\end{minipage}}}%
    \put(0.22345134,0.90497131){\color[rgb]{0,0,0}\makebox(0,0)[lt]{\begin{minipage}{0.25052446\unitlength}\raggedright $q_{2}$\end{minipage}}}%
    \put(0.37198356,0.90497131){\color[rgb]{0,0,0}\makebox(0,0)[lt]{\begin{minipage}{0.25052446\unitlength}\raggedright $q_{3}$\end{minipage}}}%
    \put(0.52051573,0.90497131){\color[rgb]{0,0,0}\makebox(0,0)[lt]{\begin{minipage}{0.25052446\unitlength}\raggedright $q_{4}$\end{minipage}}}%
    \put(0.67555623,0.90497131){\color[rgb]{0,0,0}\makebox(0,0)[lt]{\begin{minipage}{0.25052446\unitlength}\raggedright $q_{5}$\end{minipage}}}%
    \put(0.21663095,0.75028644){\color[rgb]{0,0,0}\makebox(0,0)[lt]{\begin{minipage}{0.25052446\unitlength}\raggedright $Y^{(2)}_{[1]}$\end{minipage}}}%
    \put(0.82646504,0.90421676){\color[rgb]{0,0,0}\makebox(0,0)[lt]{\begin{minipage}{0.25052446\unitlength}\raggedright $q_{6}$\end{minipage}}}%
    \put(0.06283152,0.56855899){\color[rgb]{0,0,0}\makebox(0,0)[lt]{\begin{minipage}{0.25052446\unitlength}\raggedright $Y^{(1)^\prime}_{[1]}$\end{minipage}}}%
    \put(0.36067174,0.56929135){\color[rgb]{0,0,0}\makebox(0,0)[lt]{\begin{minipage}{0.25052446\unitlength}\raggedright $Y^{(2)^\prime}_{[3]}$\end{minipage}}}%
    \put(0.51006712,0.56982488){\color[rgb]{0,0,0}\makebox(0,0)[lt]{\begin{minipage}{0.25052446\unitlength}\raggedright $Y^{(1)^\prime}_{[2]}$\end{minipage}}}%
    \put(0.66510756,0.56982488){\color[rgb]{0,0,0}\makebox(0,0)[lt]{\begin{minipage}{0.25052446\unitlength}\raggedright $Y^{(3)^\prime}_{[2]}$\end{minipage}}}%
    \put(0.81712975,0.56982488){\color[rgb]{0,0,0}\makebox(0,0)[lt]{\begin{minipage}{0.25052446\unitlength}\raggedright $Y^{(3)^\prime}_{[4]}$\end{minipage}}}%
    \put(0.21961915,0.56924019){\color[rgb]{0,0,0}\makebox(0,0)[lt]{\begin{minipage}{0.25052446\unitlength}\raggedright $Y^{(2)^\prime}_{[1]}$\end{minipage}}}%
    \put(0.57543824,0.39582547){\color[rgb]{0,0,0}\makebox(0,0)[lt]{\begin{minipage}{0.29801083\unitlength}\raggedright $\Gamma_{\ell \in \mathcal{N}[2]}[ \mu_{\epsilon}^{(2,\ell)}$ ]\end{minipage}}}%
    \put(0,0){\includegraphics[width=\unitlength,page=22]{dcmtf_model_v3_pass_2.pdf}}%
    \put(0.43349192,0.27091414){\color[rgb]{0,0,0}\makebox(0,0)[lt]{\begin{minipage}{0.25052446\unitlength}\raggedright $U^{[3]}$\end{minipage}}}%
    \put(0,0){\includegraphics[width=\unitlength,page=23]{dcmtf_model_v3_pass_2.pdf}}%
    \put(0.88711539,0.27091412){\color[rgb]{0,0,0}\makebox(0,0)[lt]{\begin{minipage}{0.25052446\unitlength}\raggedright $U^{[4]}$\end{minipage}}}%
    \put(0,0){\includegraphics[width=\unitlength,page=24]{dcmtf_model_v3_pass_2.pdf}}%
    \put(0.43543862,0.08415504){\color[rgb]{0,0,0}\makebox(0,0)[lt]{\begin{minipage}{0.25052446\unitlength}\raggedright $C^{[3]}$\end{minipage}}}%
    \put(0,0){\includegraphics[width=\unitlength,page=25]{dcmtf_model_v3_pass_2.pdf}}%
    \put(0.4411016,0.14187514){\color[rgb]{0,0,0}\makebox(0,0)[lt]{\begin{minipage}{0.25052446\unitlength}\raggedright $o$\end{minipage}}}%
    \put(0.42448066,0.17055227){\color[rgb]{0,0,0}\makebox(0,0)[lt]{\begin{minipage}{0.25052446\unitlength}\raggedright $\tilde{C}^{[3]}$\end{minipage}}}%
    \put(0,0){\includegraphics[width=\unitlength,page=26]{dcmtf_model_v3_pass_2.pdf}}%
    \put(0.36759702,0.03040143){\color[rgb]{0,0,0}\makebox(0,0)[lt]{\begin{minipage}{0.25052446\unitlength}\raggedright $\ell_\mathcal{C} = Tr(C^{(3)^T}L^{(3)}C^{(3)})$\end{minipage}}}%
    \put(0,0){\includegraphics[width=\unitlength,page=27]{dcmtf_model_v3_pass_2.pdf}}%
    \put(0.20981957,0.08293493){\color[rgb]{0,0,0}\makebox(0,0)[lt]{\begin{minipage}{0.25052446\unitlength}\raggedright $C^{[1]}$\end{minipage}}}%
    \put(0,0){\includegraphics[width=\unitlength,page=28]{dcmtf_model_v3_pass_2.pdf}}%
    \put(0.21548256,0.14065503){\color[rgb]{0,0,0}\makebox(0,0)[lt]{\begin{minipage}{0.25052446\unitlength}\raggedright $o$\end{minipage}}}%
    \put(0.19886161,0.16933216){\color[rgb]{0,0,0}\makebox(0,0)[lt]{\begin{minipage}{0.25052446\unitlength}\raggedright $\tilde{C}^{[1]}$\end{minipage}}}%
    \put(0,0){\includegraphics[width=\unitlength,page=29]{dcmtf_model_v3_pass_2.pdf}}%
    \put(0.14037733,0.03024844){\color[rgb]{0,0,0}\makebox(0,0)[lt]{\begin{minipage}{0.25052446\unitlength}\raggedright $\ell_\mathcal{C} = Tr(C^{(1)^T}L^{(1)}C^{(1)})$\end{minipage}}}%
    \put(0,0){\includegraphics[width=\unitlength,page=30]{dcmtf_model_v3_pass_2.pdf}}%
    \put(0.66029963,0.08694207){\color[rgb]{0,0,0}\makebox(0,0)[lt]{\begin{minipage}{0.25052446\unitlength}\raggedright $C^{[2]}$\end{minipage}}}%
    \put(0,0){\includegraphics[width=\unitlength,page=31]{dcmtf_model_v3_pass_2.pdf}}%
    \put(0.66596264,0.14466217){\color[rgb]{0,0,0}\makebox(0,0)[lt]{\begin{minipage}{0.25052446\unitlength}\raggedright $o$\end{minipage}}}%
    \put(0.64934168,0.1733393){\color[rgb]{0,0,0}\makebox(0,0)[lt]{\begin{minipage}{0.25052446\unitlength}\raggedright $\tilde{C}^{[2]}$\end{minipage}}}%
    \put(0,0){\includegraphics[width=\unitlength,page=32]{dcmtf_model_v3_pass_2.pdf}}%
    \put(0.59512581,0.03078747){\color[rgb]{0,0,0}\makebox(0,0)[lt]{\begin{minipage}{0.25052446\unitlength}\raggedright $\ell_\mathcal{C} = Tr(C^{(2)^T}L^{(2)}C^{(2)})$\end{minipage}}}%
    \put(0,0){\includegraphics[width=\unitlength,page=33]{dcmtf_model_v3_pass_2.pdf}}%
    \put(0.88919464,0.08800918){\color[rgb]{0,0,0}\makebox(0,0)[lt]{\begin{minipage}{0.25052446\unitlength}\raggedright $C^{[4]}$\end{minipage}}}%
    \put(0,0){\includegraphics[width=\unitlength,page=34]{dcmtf_model_v3_pass_2.pdf}}%
    \put(0.89485768,0.14572929){\color[rgb]{0,0,0}\makebox(0,0)[lt]{\begin{minipage}{0.25052446\unitlength}\raggedright $o$\end{minipage}}}%
    \put(0.87823668,0.17440641){\color[rgb]{0,0,0}\makebox(0,0)[lt]{\begin{minipage}{0.25052446\unitlength}\raggedright $\tilde{C}^{[4]}$\end{minipage}}}%
    \put(0,0){\includegraphics[width=\unitlength,page=35]{dcmtf_model_v3_pass_2.pdf}}%
    \put(0.82235206,0.03132102){\color[rgb]{0,0,0}\makebox(0,0)[lt]{\begin{minipage}{0.25052446\unitlength}\raggedright $\ell_\mathcal{C} = Tr(C^{(4)^T}L^{(4)}C^{(4)})$\end{minipage}}}%
  \end{picture}%
\endgroup%

%% file: adr_setup_v2.pdf_tex
\begingroup%
  \makeatletter%
  \providecommand\color[2][]{%
    \errmessage{(Inkscape) Color is used for the text in Inkscape, but the package 'color.sty' is not loaded}%
    \renewcommand\color[2][]{}%
  }%
  \providecommand\transparent[1]{%
    \errmessage{(Inkscape) Transparency is used (non-zero) for the text in Inkscape, but the package 'transparent.sty' is not loaded}%
    \renewcommand\transparent[1]{}%
  }%
  \providecommand\rotatebox[2]{#2}%
  \newcommand*\fsize{\dimexpr\f@size pt\relax}%
  \newcommand*\lineheight[1]{\fontsize{\fsize}{#1\fsize}\selectfont}%
  \ifx\svgwidth\undefined%
    \setlength{\unitlength}{137.90080052bp}%
    \ifx\svgscale\undefined%
      \relax%
    \else%
      \setlength{\unitlength}{\unitlength * \real{\svgscale}}%
    \fi%
  \else%
    \setlength{\unitlength}{\svgwidth}%
  \fi%
  \global\let\svgwidth\undefined%
  \global\let\svgscale\undefined%
  \makeatother%
  \begin{picture}(1,0.91254439)%
    \lineheight{1}%
    \setlength\tabcolsep{0pt}%
    \put(0,0){\includegraphics[width=\unitlength,page=1]{adr_setup_v2.pdf}}%
    \put(0.0816084,0.33021936){\color[rgb]{0,0,0}\makebox(0,0)[lt]{\begin{minipage}{0.14951105\unitlength}\raggedright $p$\end{minipage}}}%
    \put(0.28268828,0.47162526){\color[rgb]{0,0,0}\makebox(0,0)[lt]{\begin{minipage}{0.14951105\unitlength}\raggedright $r$\end{minipage}}}%
    \put(0.67000878,0.47162526){\color[rgb]{0,0,0}\makebox(0,0)[lt]{\begin{minipage}{0.14951105\unitlength}\raggedright $t$\end{minipage}}}%
    \put(0.25207131,0.84145656){\color[rgb]{0,0,0}\makebox(0,0)[lt]{\begin{minipage}{0.14951105\unitlength}\raggedright $r$\end{minipage}}}%
    \put(0.27432127,0.33514781){\color[rgb]{0,0,0}\makebox(0,0)[lt]{\begin{minipage}{0.32214536\unitlength}\raggedright $X^{(p,r)}$\end{minipage}}}%
    \put(0.62820851,0.33514781){\color[rgb]{0,0,0}\makebox(0,0)[lt]{\begin{minipage}{0.32214536\unitlength}\raggedright $X^{(p,t)}$\end{minipage}}}%
    \put(0.20980143,0.7049791){\color[rgb]{0,0,0}\makebox(0,0)[lt]{\begin{minipage}{0.32214536\unitlength}\raggedright $X^{(s,r)}$\end{minipage}}}%
    \put(0.08433129,0.68208883){\color[rgb]{0,0,0}\makebox(0,0)[lt]{\begin{minipage}{0.14951105\unitlength}\raggedright $s$\end{minipage}}}%
  \end{picture}%
\endgroup%

%% file: wiki_setup_v1.pdf_tex
\begingroup%
  \makeatletter%
  \providecommand\color[2][]{%
    \errmessage{(Inkscape) Color is used for the text in Inkscape, but the package 'color.sty' is not loaded}%
    \renewcommand\color[2][]{}%
  }%
  \providecommand\transparent[1]{%
    \errmessage{(Inkscape) Transparency is used (non-zero) for the text in Inkscape, but the package 'transparent.sty' is not loaded}%
    \renewcommand\transparent[1]{}%
  }%
  \providecommand\rotatebox[2]{#2}%
  \newcommand*\fsize{\dimexpr\f@size pt\relax}%
  \newcommand*\lineheight[1]{\fontsize{\fsize}{#1\fsize}\selectfont}%
  \ifx\svgwidth\undefined%
    \setlength{\unitlength}{137.90080153bp}%
    \ifx\svgscale\undefined%
      \relax%
    \else%
      \setlength{\unitlength}{\unitlength * \real{\svgscale}}%
    \fi%
  \else%
    \setlength{\unitlength}{\svgwidth}%
  \fi%
  \global\let\svgwidth\undefined%
  \global\let\svgscale\undefined%
  \makeatother%
  \begin{picture}(1,0.91254439)%
    \lineheight{1}%
    \setlength\tabcolsep{0pt}%
    \put(0,0){\includegraphics[width=\unitlength,page=1]{wiki_setup_v1.pdf}}%
    \put(0.0816084,0.33021936){\color[rgb]{0,0,0}\makebox(0,0)[lt]{\begin{minipage}{0.14951104\unitlength}\raggedright $[b]$\end{minipage}}}%
    \put(0.67000877,0.47162526){\color[rgb]{0,0,0}\makebox(0,0)[lt]{\begin{minipage}{0.14951104\unitlength}\raggedright $[t]$\end{minipage}}}%
    \put(0.2520713,0.84145656){\color[rgb]{0,0,0}\makebox(0,0)[lt]{\begin{minipage}{0.14951104\unitlength}\raggedright $[z]$\end{minipage}}}%
    \put(0.27432127,0.33514781){\color[rgb]{0,0,0}\makebox(0,0)[lt]{\begin{minipage}{0.32214536\unitlength}\raggedright $X_{[b],[z]}^{(2)}$\end{minipage}}}%
    \put(0.6282085,0.33514781){\color[rgb]{0,0,0}\makebox(0,0)[lt]{\begin{minipage}{0.32214536\unitlength}\raggedright $X_{[b],[t]}^{(3)}$\end{minipage}}}%
    \put(0.27506573,0.70497911){\color[rgb]{0,0,0}\makebox(0,0)[lt]{\begin{minipage}{0.32214536\unitlength}\raggedright $X_{[t],[z]}^{(1)}$\end{minipage}}}%
    \put(0.08433129,0.68208883){\color[rgb]{0,0,0}\makebox(0,0)[lt]{\begin{minipage}{0.14951104\unitlength}\raggedright $[t]$\end{minipage}}}%
  \end{picture}%
\endgroup%

%% file: cancer_setup_v2.pdf_tex
\begingroup%
  \makeatletter%
  \providecommand\color[2][]{%
    \errmessage{(Inkscape) Color is used for the text in Inkscape, but the package 'color.sty' is not loaded}%
    \renewcommand\color[2][]{}%
  }%
  \providecommand\transparent[1]{%
    \errmessage{(Inkscape) Transparency is used (non-zero) for the text in Inkscape, but the package 'transparent.sty' is not loaded}%
    \renewcommand\transparent[1]{}%
  }%
  \providecommand\rotatebox[2]{#2}%
  \newcommand*\fsize{\dimexpr\f@size pt\relax}%
  \newcommand*\lineheight[1]{\fontsize{\fsize}{#1\fsize}\selectfont}%
  \ifx\svgwidth\undefined%
    \setlength{\unitlength}{179.97105276bp}%
    \ifx\svgscale\undefined%
      \relax%
    \else%
      \setlength{\unitlength}{\unitlength * \real{\svgscale}}%
    \fi%
  \else%
    \setlength{\unitlength}{\svgwidth}%
  \fi%
  \global\let\svgwidth\undefined%
  \global\let\svgscale\undefined%
  \makeatother%
  \begin{picture}(1,0.82947994)%
    \lineheight{1}%
    \setlength\tabcolsep{0pt}%
    \put(0,0){\includegraphics[width=\unitlength,page=1]{cancer_setup_v2.pdf}}%
    \put(0.16037564,0.44649137){\color[rgb]{0,0,0}\makebox(0,0)[lt]{\begin{minipage}{0.24684026\unitlength}\raggedright $X^{(3)}$\end{minipage}}}%
    \put(0.45654094,0.44649137){\color[rgb]{0,0,0}\makebox(0,0)[lt]{\begin{minipage}{0.24684026\unitlength}\raggedright $X^{(2)}$\end{minipage}}}%
    \put(0.72770207,0.44649137){\color[rgb]{0,0,0}\makebox(0,0)[lt]{\begin{minipage}{0.24684026\unitlength}\raggedright $X^{(1)}$\end{minipage}}}%
    \put(0.4541725,0.16950115){\color[rgb]{0,0,0}\makebox(0,0)[lt]{\begin{minipage}{0.24684026\unitlength}\raggedright $X^{(5)}$\end{minipage}}}%
    \put(0.74884746,0.6957069){\color[rgb]{0,0,0}\makebox(0,0)[lt]{\begin{minipage}{0.24684026\unitlength}\raggedright $X^{(4)}$\end{minipage}}}%
    \put(0.01338174,0.41771096){\color[rgb]{0,0,0}\makebox(0,0)[lt]{\begin{minipage}{0.20433724\unitlength}\raggedright $E^{[1]}$\end{minipage}}}%
    \put(0.19389348,0.56061507){\color[rgb]{0,0,0}\makebox(0,0)[lt]{\begin{minipage}{0.15697872\unitlength}\raggedright $E^{[4]}$\end{minipage}}}%
    \put(0.48449081,0.56210341){\color[rgb]{0,0,0}\makebox(0,0)[lt]{\begin{minipage}{0.15697872\unitlength}\raggedright $E^{[3]}$\end{minipage}}}%
    \put(0.31318348,0.1675459){\color[rgb]{0,0,0}\makebox(0,0)[lt]{\begin{minipage}{0.15697872\unitlength}\raggedright $E^{[5]}$\end{minipage}}}%
    \put(0.61144573,0.679533){\color[rgb]{0,0,0}\makebox(0,0)[lt]{\begin{minipage}{0.15697872\unitlength}\raggedright $E^{[6]}$\end{minipage}}}%
    \put(0.78364607,0.82256195){\color[rgb]{0,0,0}\makebox(0,0)[lt]{\begin{minipage}{0.15697872\unitlength}\raggedright $E^{[2]}$\end{minipage}}}%
  \end{picture}%
\endgroup%

%% file: freebase_setup_v2.pdf_tex
\begingroup%
  \makeatletter%
  \providecommand\color[2][]{%
    \errmessage{(Inkscape) Color is used for the text in Inkscape, but the package 'color.sty' is not loaded}%
    \renewcommand\color[2][]{}%
  }%
  \providecommand\transparent[1]{%
    \errmessage{(Inkscape) Transparency is used (non-zero) for the text in Inkscape, but the package 'transparent.sty' is not loaded}%
    \renewcommand\transparent[1]{}%
  }%
  \providecommand\rotatebox[2]{#2}%
  \newcommand*\fsize{\dimexpr\f@size pt\relax}%
  \newcommand*\lineheight[1]{\fontsize{\fsize}{#1\fsize}\selectfont}%
  \ifx\svgwidth\undefined%
    \setlength{\unitlength}{276.07427162bp}%
    \ifx\svgscale\undefined%
      \relax%
    \else%
      \setlength{\unitlength}{\unitlength * \real{\svgscale}}%
    \fi%
  \else%
    \setlength{\unitlength}{\svgwidth}%
  \fi%
  \global\let\svgwidth\undefined%
  \global\let\svgscale\undefined%
  \makeatother%
  \begin{picture}(1,0.8762033)%
    \lineheight{1}%
    \setlength\tabcolsep{0pt}%
    \put(0,0){\includegraphics[width=\unitlength,page=1]{freebase_setup_v2.pdf}}%
    \put(0.15947504,0.56383208){\color[rgb]{0,0,0}\makebox(0,0)[lt]{\begin{minipage}{0.14150891\unitlength}\raggedright $X^{(1)}$\end{minipage}}}%
    \put(0.02534043,0.56053121){\color[rgb]{0,0,0}\makebox(0,0)[lt]{\begin{minipage}{0.13148145\unitlength}\raggedright $E^{[1]}$\end{minipage}}}%
    \put(0.16078541,0.64901672){\color[rgb]{0,0,0}\makebox(0,0)[lt]{\begin{minipage}{0.13148145\unitlength}\raggedright $E^{[1]}$\end{minipage}}}%
    \put(0,0){\includegraphics[width=\unitlength,page=2]{freebase_setup_v2.pdf}}%
    \put(0.59416492,0.56480232){\color[rgb]{0,0,0}\makebox(0,0)[lt]{\begin{minipage}{0.14150891\unitlength}\raggedright $X^{(4)}$\end{minipage}}}%
    \put(0.46003035,0.56150144){\color[rgb]{0,0,0}\makebox(0,0)[lt]{\begin{minipage}{0.13148145\unitlength}\raggedright $E^{[1]}$\end{minipage}}}%
    \put(0.59547529,0.64998696){\color[rgb]{0,0,0}\makebox(0,0)[lt]{\begin{minipage}{0.13148145\unitlength}\raggedright $E^{[2]}$\end{minipage}}}%
    \put(0,0){\includegraphics[width=\unitlength,page=3]{freebase_setup_v2.pdf}}%
    \put(0.84545597,0.56577255){\color[rgb]{0,0,0}\makebox(0,0)[lt]{\begin{minipage}{0.14150891\unitlength}\raggedright $X^{(2)}$\end{minipage}}}%
    \put(0.71132143,0.56247168){\color[rgb]{0,0,0}\makebox(0,0)[lt]{\begin{minipage}{0.13148145\unitlength}\raggedright $E^{[1]}$\end{minipage}}}%
    \put(0.84676637,0.65095719){\color[rgb]{0,0,0}\makebox(0,0)[lt]{\begin{minipage}{0.13148145\unitlength}\raggedright $E^{[3]}$\end{minipage}}}%
    \put(0,0){\includegraphics[width=\unitlength,page=4]{freebase_setup_v2.pdf}}%
    \put(0.84545597,0.78572589){\color[rgb]{0,0,0}\makebox(0,0)[lt]{\begin{minipage}{0.14150891\unitlength}\raggedright $X^{(7)}$\end{minipage}}}%
    \put(0.71132143,0.78242501){\color[rgb]{0,0,0}\makebox(0,0)[lt]{\begin{minipage}{0.13148145\unitlength}\raggedright $E^{[4]}$\end{minipage}}}%
    \put(0.84676637,0.8654772){\color[rgb]{0,0,0}\makebox(0,0)[lt]{\begin{minipage}{0.13148145\unitlength}\raggedright $E^{[3]}$\end{minipage}}}%
    \put(0,0){\includegraphics[width=\unitlength,page=5]{freebase_setup_v2.pdf}}%
    \put(0.59707562,0.35348495){\color[rgb]{0,0,0}\makebox(0,0)[lt]{\begin{minipage}{0.14150891\unitlength}\raggedright $X^{(6)}$\end{minipage}}}%
    \put(0.46294105,0.35018408){\color[rgb]{0,0,0}\makebox(0,0)[lt]{\begin{minipage}{0.13148145\unitlength}\raggedright $E^{[5]}$\end{minipage}}}%
    \put(0.59838599,0.43866959){\color[rgb]{0,0,0}\makebox(0,0)[lt]{\begin{minipage}{0.13148145\unitlength}\raggedright $E^{[2]}$\end{minipage}}}%
    \put(0,0){\includegraphics[width=\unitlength,page=6]{freebase_setup_v2.pdf}}%
    \put(0.59707562,0.14983155){\color[rgb]{0,0,0}\makebox(0,0)[lt]{\begin{minipage}{0.14150891\unitlength}\raggedright $X^{(5)}$\end{minipage}}}%
    \put(0.46294105,0.14653068){\color[rgb]{0,0,0}\makebox(0,0)[lt]{\begin{minipage}{0.13148145\unitlength}\raggedright $E^{[6]}$\end{minipage}}}%
    \put(0.59838599,0.23501619){\color[rgb]{0,0,0}\makebox(0,0)[lt]{\begin{minipage}{0.13148145\unitlength}\raggedright $E^{[2]}$\end{minipage}}}%
    \put(0,0){\includegraphics[width=\unitlength,page=7]{freebase_setup_v2.pdf}}%
    \put(0.32735007,0.14585449){\color[rgb]{0,0,0}\makebox(0,0)[lt]{\begin{minipage}{0.14150891\unitlength}\raggedright $X^{(3)}$\end{minipage}}}%
    \put(0.1932155,0.14255362){\color[rgb]{0,0,0}\makebox(0,0)[lt]{\begin{minipage}{0.13148145\unitlength}\raggedright $E^{[6]}$\end{minipage}}}%
    \put(0.32866044,0.23103913){\color[rgb]{0,0,0}\makebox(0,0)[lt]{\begin{minipage}{0.13148145\unitlength}\raggedright $E^{[7]}$\end{minipage}}}%
  \end{picture}%
\endgroup%

%% file: pubmed_setup_v2.pdf_tex
\begingroup%
  \makeatletter%
  \providecommand\color[2][]{%
    \errmessage{(Inkscape) Color is used for the text in Inkscape, but the package 'color.sty' is not loaded}%
    \renewcommand\color[2][]{}%
  }%
  \providecommand\transparent[1]{%
    \errmessage{(Inkscape) Transparency is used (non-zero) for the text in Inkscape, but the package 'transparent.sty' is not loaded}%
    \renewcommand\transparent[1]{}%
  }%
  \providecommand\rotatebox[2]{#2}%
  \newcommand*\fsize{\dimexpr\f@size pt\relax}%
  \newcommand*\lineheight[1]{\fontsize{\fsize}{#1\fsize}\selectfont}%
  \ifx\svgwidth\undefined%
    \setlength{\unitlength}{310.3600065bp}%
    \ifx\svgscale\undefined%
      \relax%
    \else%
      \setlength{\unitlength}{\unitlength * \real{\svgscale}}%
    \fi%
  \else%
    \setlength{\unitlength}{\svgwidth}%
  \fi%
  \global\let\svgwidth\undefined%
  \global\let\svgscale\undefined%
  \makeatother%
  \begin{picture}(1,0.78458669)%
    \lineheight{1}%
    \setlength\tabcolsep{0pt}%
    \put(0,0){\includegraphics[width=\unitlength,page=1]{pubmed_setup_v2.pdf}}%
    \put(0.12275078,0.69139427){\color[rgb]{0,0,0}\makebox(0,0)[lt]{\begin{minipage}{0.1258763\unitlength}\raggedright $X^{(2)}$\end{minipage}}}%
    \put(0.00343415,0.68845804){\color[rgb]{0,0,0}\makebox(0,0)[lt]{\begin{minipage}{0.11695658\unitlength}\raggedright $E^{[1]}$\end{minipage}}}%
    \put(0.13358271,0.77200159){\color[rgb]{0,0,0}\makebox(0,0)[lt]{\begin{minipage}{0.11695658\unitlength}\raggedright $E^{[2]}$\end{minipage}}}%
    \put(0,0){\includegraphics[width=\unitlength,page=2]{pubmed_setup_v2.pdf}}%
    \put(0.12275078,0.50057243){\color[rgb]{0,0,0}\makebox(0,0)[lt]{\begin{minipage}{0.1258763\unitlength}\raggedright $X^{(5)}$\end{minipage}}}%
    \put(0.00343415,0.49763621){\color[rgb]{0,0,0}\makebox(0,0)[lt]{\begin{minipage}{0.11695658\unitlength}\raggedright $E^{[3]}$\end{minipage}}}%
    \put(0.13358271,0.58044656){\color[rgb]{0,0,0}\makebox(0,0)[lt]{\begin{minipage}{0.11695658\unitlength}\raggedright $E^{[2]}$\end{minipage}}}%
    \put(0,0){\includegraphics[width=\unitlength,page=3]{pubmed_setup_v2.pdf}}%
    \put(0.12275078,0.31458369){\color[rgb]{0,0,0}\makebox(0,0)[lt]{\begin{minipage}{0.1258763\unitlength}\raggedright $X^{(9)}$\end{minipage}}}%
    \put(0.00343415,0.31164746){\color[rgb]{0,0,0}\makebox(0,0)[lt]{\begin{minipage}{0.11695658\unitlength}\raggedright $E^{[4]}$\end{minipage}}}%
    \put(0.13358271,0.395191){\color[rgb]{0,0,0}\makebox(0,0)[lt]{\begin{minipage}{0.11695658\unitlength}\raggedright $E^{[2]}$\end{minipage}}}%
    \put(0,0){\includegraphics[width=\unitlength,page=4]{pubmed_setup_v2.pdf}}%
    \put(0.12275078,0.13342804){\color[rgb]{0,0,0}\makebox(0,0)[lt]{\begin{minipage}{0.1258763\unitlength}\raggedright $X^{(3)}$\end{minipage}}}%
    \put(0.00343415,0.13049181){\color[rgb]{0,0,0}\makebox(0,0)[lt]{\begin{minipage}{0.11695658\unitlength}\raggedright $E^{[2]}$\end{minipage}}}%
    \put(0.13358271,0.20920225){\color[rgb]{0,0,0}\makebox(0,0)[lt]{\begin{minipage}{0.11695658\unitlength}\raggedright $E^{[2]}$\end{minipage}}}%
    \put(0,0){\includegraphics[width=\unitlength,page=5]{pubmed_setup_v2.pdf}}%
    \put(0.3696532,0.69139427){\color[rgb]{0,0,0}\makebox(0,0)[lt]{\begin{minipage}{0.1258763\unitlength}\raggedright $X^{(1)}$\end{minipage}}}%
    \put(0.25033657,0.68845804){\color[rgb]{0,0,0}\makebox(0,0)[lt]{\begin{minipage}{0.11695658\unitlength}\raggedright $E^{[1]}$\end{minipage}}}%
    \put(0.38048504,0.77200159){\color[rgb]{0,0,0}\makebox(0,0)[lt]{\begin{minipage}{0.11695658\unitlength}\raggedright $E^{[1]}$\end{minipage}}}%
    \put(0,0){\includegraphics[width=\unitlength,page=6]{pubmed_setup_v2.pdf}}%
    \put(0.3696532,0.50057243){\color[rgb]{0,0,0}\makebox(0,0)[lt]{\begin{minipage}{0.1258763\unitlength}\raggedright $X^{(4)}$\end{minipage}}}%
    \put(0.25033657,0.49763621){\color[rgb]{0,0,0}\makebox(0,0)[lt]{\begin{minipage}{0.11695658\unitlength}\raggedright $E^{[3]}$\end{minipage}}}%
    \put(0.38048504,0.58044656){\color[rgb]{0,0,0}\makebox(0,0)[lt]{\begin{minipage}{0.11695658\unitlength}\raggedright $E^{[1]}$\end{minipage}}}%
    \put(0,0){\includegraphics[width=\unitlength,page=7]{pubmed_setup_v2.pdf}}%
    \put(0.36977824,0.3144127){\color[rgb]{0,0,0}\makebox(0,0)[lt]{\begin{minipage}{0.1258763\unitlength}\raggedright $X^{(8)}$\end{minipage}}}%
    \put(0.25046161,0.31147647){\color[rgb]{0,0,0}\makebox(0,0)[lt]{\begin{minipage}{0.11695658\unitlength}\raggedright $E^{[4]}$\end{minipage}}}%
    \put(0.38061017,0.39502002){\color[rgb]{0,0,0}\makebox(0,0)[lt]{\begin{minipage}{0.11695658\unitlength}\raggedright $E^{[1]}$\end{minipage}}}%
    \put(0,0){\includegraphics[width=\unitlength,page=8]{pubmed_setup_v2.pdf}}%
    \put(0.61655553,0.50057243){\color[rgb]{0,0,0}\makebox(0,0)[lt]{\begin{minipage}{0.1258763\unitlength}\raggedright $X^{(6)}$\end{minipage}}}%
    \put(0.4972389,0.49763621){\color[rgb]{0,0,0}\makebox(0,0)[lt]{\begin{minipage}{0.11695658\unitlength}\raggedright $E^{[3]}$\end{minipage}}}%
    \put(0.62738746,0.58044656){\color[rgb]{0,0,0}\makebox(0,0)[lt]{\begin{minipage}{0.11695658\unitlength}\raggedright $E^{[3]}$\end{minipage}}}%
    \put(0,0){\includegraphics[width=\unitlength,page=9]{pubmed_setup_v2.pdf}}%
    \put(0.85550439,0.49910604){\color[rgb]{0,0,0}\makebox(0,0)[lt]{\begin{minipage}{0.1258763\unitlength}\raggedright $X^{(7)}$\end{minipage}}}%
    \put(0.74102089,0.49616981){\color[rgb]{0,0,0}\makebox(0,0)[lt]{\begin{minipage}{0.11695658\unitlength}\raggedright $E^{[3]}$\end{minipage}}}%
    \put(0.87116944,0.58044656){\color[rgb]{0,0,0}\makebox(0,0)[lt]{\begin{minipage}{0.11695658\unitlength}\raggedright $E^{[4]}$\end{minipage}}}%
    \put(0,0){\includegraphics[width=\unitlength,page=10]{pubmed_setup_v2.pdf}}%
    \put(0.85550439,0.31795039){\color[rgb]{0,0,0}\makebox(0,0)[lt]{\begin{minipage}{0.1258763\unitlength}\raggedright $X^{(10)}$\end{minipage}}}%
    \put(0.74102089,0.31501416){\color[rgb]{0,0,0}\makebox(0,0)[lt]{\begin{minipage}{0.11695658\unitlength}\raggedright $E^{[4]}$\end{minipage}}}%
    \put(0.87116944,0.39855771){\color[rgb]{0,0,0}\makebox(0,0)[lt]{\begin{minipage}{0.11695658\unitlength}\raggedright $E^{[4]}$\end{minipage}}}%
  \end{picture}%
\endgroup%